\documentclass[11pt]{article}
\usepackage[utf8]{inputenc}
\usepackage{times}
\usepackage[backend=bibtex,giveninits=true, maxbibnames=5]{biblatex}
\usepackage{biblatex}
\usepackage{subfig}
\usepackage[normalem]{ulem}
\usepackage{float}
\usepackage{cancel}
\usepackage{algorithm, changepage}
\usepackage[noend]{algpseudocode}

\bibliography{cc-online}

\usepackage{setspace}

\usepackage{soul}

\usepackage{nicefrac}

\usepackage[margin=1in]{geometry}
\usepackage{comment}

\usepackage{mathtools}

\usepackage{amsmath,amsthm,amssymb,latexsym,amsfonts, dsfont}
\usepackage{thmtools}
\usepackage{graphicx}
\usepackage[export]{adjustbox}
\usepackage[font=footnotesize,labelfont=bf]{caption}
\usepackage{mathabx}
\usepackage{hyperref}
\usepackage{lineno}
\usepackage{amsthm}
\usepackage{bm}
\usepackage{listings}
\usepackage[nottoc,numbib]{tocbibind}
\usepackage{soul}
\usepackage{thm-restate}
\usepackage{xcolor}

\definecolor{mypink}{RGB}{220,38,127}
\definecolor{myblue}{RGB}{100,143,255}
\definecolor{mygreen}{RGB}{74,167,103}
\definecolor{myorange}{RGB}{254,97,0}
\definecolor{myyellow}{RGB}{255,176,0}
\definecolor{mypurple}{RGB}{120,94,240}

\newtheorem{theorem}{Theorem}
\newtheorem{case}{Case}
\newtheorem{claim}{Claim}

\newtheorem{corollary}{Corollary}

\newtheorem{definition}{Definition}

\newtheorem{lemma}{Lemma}

\newtheorem{proposition}{Proposition}

\newtheorem{observation}{Observation}
\newtheorem{fact}{Fact}

\newcommand{\eps}{\varepsilon}
\newcommand{\ball}{\text{Ball}}

\newcommand{\cratio}{\textsf{CR}}

\newcommand{\opt}{\textsf{OPT}}

\allowdisplaybreaks

\DeclareCiteCommand{\citeyearpar}
    {}
    {\mkbibparens{\bibhyperref{\printdate}}}
    {\multicitedelim}
    {}

\title{Online Correlation Clustering:\\ Simultaneously Optimizing All $\ell_p$-Norms}

\author{Sami Davies\thanks{Department of EECS at UC Berkeley and RelationalAI.} \and Benjamin Moseley\thanks{Tepper School of Business, Carnegie Mellon University. Benjamin Moseley is supported in part by  a Google Research Award, NSF grants CCF-2121744 and  CCF-1845146, and ONR Grant N000142212702.} \and Heather Newman\thanks{Department of Computer Science, Vassar College}}

\begin{document}

\date{}


\maketitle


\abstract{
The $\ell_p$-norm objectives for correlation clustering present a fundamental trade-off between minimizing total disagreements (the $\ell_1$-norm) and ensuring fairness to individual nodes (the $\ell_\infty$-norm).
Surprisingly, in the offline setting it is possible to simultaneously approximate all $\ell_p$-norms with a single clustering.
Can this powerful guarantee be achieved in an online setting?
This paper provides the first affirmative answer. We present a single algorithm for the online-with-a-sample (AOS) model that, given a small constant fraction of the input as a sample, produces one clustering that is \emph{simultaneously} $O(\log^4 n)$-competitive for all $\ell_p$-norms with high probability, $O(\log n)$-competitive for the $\ell_\infty$-norm with high probability, and $O(1)$-competitive for the $\ell_1$-norm in expectation. This work successfully translates the  offline ``all-norms" guarantee to the online world.

Our setting is motivated by a new hardness result that demonstrates a fundamental separation between these objectives in the standard random-order (RO) online model. Namely, while the $\ell_1$-norm is trivially $O(1)$-approximable in the RO model, we prove that any algorithm in the RO model for the fairness-promoting $\ell_\infty$-norm must have a competitive ratio of at least $\Omega(n^{ 
1/3})$. This highlights the necessity of a different beyond-worst-case model. We complement our algorithm with  lower bounds, showing our competitive ratios for the $\ell_1$- and $\ell_\infty$- norms are nearly tight in the AOS model. 
}

\section{Introduction}\label{sec: intro}

Clustering is a fundamental task in unsupervised learning. In this paper, we study \emph{correlation clustering}, where the goal is to partition a set of $n$ items based on pairwise similarity ($+$) and dissimilarity ($-$) labels. Given a \emph{complete} graph where each edge has a label of positive or negative, a clustering is a partition of the $n$ vertices. A positive edge is a \emph{disagreement} if its endpoints are in different clusters, and a negative edge is a disagreement if its endpoints are in the same cluster. The goal is to find a partition that minimizes some function of these disagreements.

The objective function controls the balance between total cost and individual equity.
The most well-studied among the objectives for correlation clustering is minimizing the $\ell_1$-norm of the \emph{disagreement vector}, which is the vector of length $n$ where the $i$th coordinate indicates the number of disagreeing edges incident to node $i$.
This objective is equivalent to minimizing the total number of disagreements. 
However, the $\ell_1$-norm can be highly ``unfair,'' leaving some nodes with a disproportionately large number of incident disagreements. 
On the other hand, the $\mathbf{\ell_\infty}$-norm is a fair objective, as it minimizes the maximum number of disagreements incident to any single node. 
The $\ell_2$-norm and other finite $\ell_p$-norms, for $p \in (1,\infty)$, interpolate between the two, allowing one to balance total error against local equity.  It is natural to wonder whether it is possible to study correlation clustering (with any of these objectives) in an online setting, when nodes arrive sequentially and must be irrevocably assigned to a cluster.

However, the lower bounds for correlation clustering are pessimistic in the standard online model, where nodes arrive one at a time in adversarial order -- a simple construction due to Mathieu, Sankur, and Schudy shows it is impossible to achieve any sublinear competitive ratio for any $\ell_p$-norm objective while maintaining cluster consistency \cite{MathieuSS10}. To circumvent this, prior works have explored relaxed, \emph{beyond-worst-case} models, but these have focused exclusively on the $\ell_1$-norm. These models include allowing limited recourse (changing a node's cluster assignment) \cite{Cohen-AddadLMP22} or, as we study here, providing the algorithm with a small, random sample of the input upfront \cite{LattanziMVWZ21}, where the \emph{remaining} part of the instance then arrives \emph{adversarially} online. This latter model, which we call the \emph{online-with-a-sample} model (AOS\footnote{We note the abbreviation AOS stands for adversarial-order model with a sample.}  for short), is motivated by applications where historical data can inform decisions on new arrivals. The model has been used to study, for example, the Secretary problem \cite{kaplan2020competitive}, online bipartite matching \cite{kaplan2022online}, and online set cover \cite{gupta2024set}. Lattanzi et. al. \cite{LattanziMVWZ21} showed that the AOS model can be used to overcome the strong lower bounds for online $\ell_1$-norm correlation clustering; they gave an $O(\nicefrac{1}{\varepsilon})$-competitive algorithm when a random sample of the nodes of size $\varepsilon \cdot n$ is revealed upfront.

Despite the importance and large body of literature on fair clustering, e.g., \cite{chierichetti2017fair, backurs2019scalable, bera2019fair, ahmadi2020fair, balkanski2025faironlineCC, ahmadian2020fair, schwartz2022fair, friggstad2021fair}, the online landscape for objectives other than the $\ell_1$-norm for correlation clustering has remained entirely unexplored. In particular, there has been no work on the fairness-promoting $\ell_\infty$-norm, or general $\ell_p$-norms, in an online setting. While the problem is natural and well-motivated, all previous approximation algorithms for the $\ell_\infty$-norm, and in fact for any $\ell_p$-norm with $p>1$, were, up until recently, based on convex program rounding, making them more difficult to adapt to the online setting than new combinatorial algorithms (see subsection \ref{sec: related-work} for more details).

As the aforementioned lower bounds show that a beyond-worst-case approach is needed, we choose to study $\ell_p$-norm correlation clustering online using the AOS model. One motivation for this choice is the following. 
While the $\ell_1$-norm admits an $O(1)$-competitive algorithm in the standard \emph{random-order} (RO) model\footnote{In the RO model, vertices arrive online in uniformly random order.} via the Pivot algorithm \cite{ACN-pivot},  we prove that for $\ell_\infty$-norm correlation clustering, every algorithm in the RO model is  $\Omega(n^{1/3})$-competitive. In other words, the $\ell_\infty$-norm is much more difficult than the $\ell_1$-norm in the RO model.  
This hardness highlights a crucial limitation of the RO model, and provides strong motivation for using the AOS model for $\ell_\infty$-norm (and more generally $\ell_p$-norm) correlation clustering. Moreover, from the perspective of comparing beyond-worst-case models, it is very interesting to understand the class of problems for which the RO and AOS models lead to roughly the same competitiveness, versus when one model is provably stronger than the other. These models are seemingly incomparable,\footnote{While the RO model is fully online, it does not contain any adversarial component. The AOS model is not fully online, but it does contain an adversarial component.} but for some problems (e.g., $\ell_1$-norm correlation clustering)
the RO and AOS models admit algorithms with roughly the same competitive ratio, while for other problems (e.g., Steiner tree \cite{argue2022learning}) the AOS model can do better than what is possible in the RO model.\footnote{The RO and AOS models (as shown in \cite{LattanziMVWZ21}) both admit $O(1)$-competitive algorithms for $\ell_1$-norm correlation clustering when $\eps$ is a constant. On the other hand, online Steiner tree has a lower bound of $\Omega(\log n)$ in the RO model, but in the AOS model there is an algorithm with competitive ratio $O(\log(\nicefrac{1}{\eps}))$.}
We ask the following:

\begin{center}
    \emph{ Can a small sample in the AOS model provide enough structural information to break the online hardness barrier for $\ell_\infty$-norm correlation clustering?}
\end{center}

We then take things a step further and ask whether we can achieve this goal without sacrificing global efficiency (the $\ell_1$-norm) and, for that matter, any intermediate $\ell_p$-norm. In particular, we ask whether we can replicate -- in the online setting -- a recent result of Davies, Moseley, and Newman \cite{davies2023one} showing that for any instance of correlation clustering, there exists a single clustering that is simultaneously $O(1)$-approximate for all $\ell_p$-norms, and moreover that it can be found efficiently in the offline setting. 
This is known as the \emph{all-norms} objective. Formally introduced by Azar et. al. \cite{azar2004all}, the all-norms objective has received a steady stream of interest for various problems, including load balancing, set cover, $k$-clustering, and facility location, and in various settings (offline, online, distributed, and parallel)\cite{kleinberg1999fairness, golovin2008all, bernstein2017simultaneously, assadi2020improved, chakrabarty2019approximation, fac-location-portfolios, cao2024simultaneously, gupta2025balancing}. The all-norms guarantee is particularly powerful because it is \emph{parameter-free}: the algorithm is oblivious to the specific fairness versus total cost trade-off desired by the user, yet it returns a clustering that is simultaneously near-optimal for all $p \in [1,\infty]$. 
An ambitious goal would be to achieve a similar guarantee in the online setting:
\begin{center}
    \emph{ 
    Is it possible to approximate the all-norms objective for correlation clustering in the AOS model?}
\end{center}

\subsection*{Our Contributions}

We answer the above questions affirmatively, showing that in the AOS model, there is an algorithm that is locally fair ($\ell_\infty$-norm) without sacrificing global efficiency ($\ell_1$-norm), and that our guarantees are nearly best possible for both norms. Moreover, we obtain a robustness guarantee for all norms in between, thus translating the offline all-norms guarantee into the online world.

Let $\textsf{OPT}_p$ denote the cost of an optimal clustering for the $\ell_p$-norm objective. 
We assume in the AOS model that an adversary fixes the online input. However,
we see upfront a random sample $S$ of $\eps n$ nodes and the induced subgraph $G[S]$, 
where each node is sampled uniformly and independently into $S$ with probability $\eps$. The parameter $0<\varepsilon<1$ can be anything in our upper bound results (Theorem \ref{thm:main-all}), and as small as $n^{-1/4}$ in our lower bound results (Theorem \ref{thm: lowerbound-inf}).

\begin{theorem} \label{thm:main-all}
Given $0 < \varepsilon < 1$, there is an algorithm in the AOS model that produces a \emph{single} clustering that is:
\begin{enumerate}
    \item \textbf{(Fairness)} \label{item: thm-infty} $O\left(\frac{
    1
    }{\varepsilon^6} \cdot \log n \right)$-competitive  for the $\ell_\infty$-norm with probability at least $1-\nicefrac{1}{n}$. 
    \item \textbf{(Simultaneous Robustness)} \label{item: thm-allp}  $O\left(\frac{1}{\eps^8} \cdot \log^4 n \right)$-competitive for the $\ell_p$-norms,  $1 \leq p < \infty $, with probability at least $1-\nicefrac{1}{n}$. 
    \item \textbf{(Global Efficiency)} \label{item: thm-1}$O\left(\frac{1}{\varepsilon^6}\right)$-competitive in expectation for the $\ell_1$-norm.
\end{enumerate}
\end{theorem}

Complementing our upper bounds, we show that the competitive ratios of our algorithm are nearly optimal for the $\ell_1$- and $\ell_\infty$- norms. In particular, the logarithmic factor for the $\ell_\infty$-norm and a dependence on $\nicefrac{1}{\varepsilon}$ for the $\ell_1$- and $\ell_\infty$- norms are necessary.
While the lower bound for the $\ell_1$-norm is known \cite{LattanziMVWZ21}, we contribute the new lower bound for the $\ell_\infty$-norm.
\begin{theorem}\label{thm: lowerbound-inf}
For any $\nicefrac{1}{n^{1/4} }\leq \varepsilon \leq \nicefrac{3}{4}$, any \emph{randomized} algorithm in the AOS model is $\Omega\left( \nicefrac{1}{\varepsilon}\cdot \log n \right)$-competitive for the $\ell_{\infty}$-norm and  $\Omega\left( \nicefrac{1}{\varepsilon}\right)$-competitive for the $\ell_1$-norm.
\end{theorem}

Finally, we give a hardness result for the $\ell_\infty$-norm in the RO model, justifying our focus on the AOS model.

\begin{theorem}
\label{thm: lowerbound-inf-ro}
Any \emph{randomized} algorithm for $\ell_\infty$-norm correlation clustering in the RO model is $\Omega(n^{1/3})$-competitive.
\end{theorem}

This result establishes a fundamental separation between the $\ell_1$- and $\ell_\infty$- norm objectives in the RO model, where an $O(1)$-competitive ratio for the $\ell_1$-norm trivially follows from the Pivot algorithm. Recent work \cite{gupta2024set} shows that for a general class of minimization problems (called \emph{augmentable integer programs}), there is a reduction from the AOS model to the RO model with loss of a factor at most $ \nicefrac{1}{\eps}$ in the competitive ratio, i.e., if there is an algorithm with competitive ratio $\Delta$ in the RO model, then there is an algorithm with competitive ratio $\Delta/\eps$ in the AOS model. In contrast, $\ell_\infty$-norm correlation clustering is an example of a minimization problem where the AOS model is actually \emph{much stronger} at breaking through worst-case instances than the RO model. Thus in this regard, $\ell_\infty$-norm correlation clustering is more similar to online Steiner tree, where the AOS model can achieve competitive ratio $O(\log(\nicefrac{1}{\eps}))$, even though the RO model has a lower bound of $\Omega(\log n)$.

\subsection{Related work} \label{sec: related-work}

\subparagraph*{Prior work offline.} 
 
Bansal, Blum, and Chawla \cite{BBC04} proposed correlation clustering for the goal of minimizing the $\ell_1$-norm of the disagreement vector.  
The problem is NP-hard, and numerous approximation algorithms have been developed \cite{ACN-pivot, chawla2015near, cohen2022correlation, cao2024understanding}. A $1.437$-approximation is known for the $\ell_1$-norm \cite{cao2024understanding}, which improves upon the work that beat the threshold of 2 \cite{cohen2022correlation}.  
Puleo and Milenkovic \cite{puleo2015correlation} proposed the $\ell_p$-norm objective for $p > 1$ and for each fixed $p$ they gave a 48-approximation. This factor has since been improved in a series of works, first to 7 \cite{CGS17},
and then to $5$ \cite{KMZ19}.  
Notably, this entire line of work on $\ell_p$-norm objectives relies on rounding solutions to convex programs.   

Davies, Moseley, and Newman \cite{DMN23} introduced the first \emph{combinatorial} $O(1)$-approximation algorithm for the $\ell_\infty$-norm.  
Heidrich, Irmai, and Andres \cite{minmax4approx} built off of the techniques in \cite{DMN23} to prove a combinatorial 4-approximation for the $\ell_{\infty}$-norm, and this was later improved by Cao, Roche, and Su \cite{cao2025min} to a combinatorial 3-approximation. Then, Davies, Moseley, and Newman \cite{davies2023one} offered a new combinatorial algorithm proving there exists a single clustering that is an $O(1)$-approximation for all $\ell_p$-norms simultaneously (also known as the all-norms objective).
Cao, Li, and Ye \cite{cao2024simultaneously} modified their algorithm in order to improve the factor for the all-norms objective, as well as to run in near-linear time in the MPC model in polylogarithmic rounds.

\subparagraph*{Prior work online.}  In the online setting, nodes arrive over time and reveal the signs of all of their edges to nodes that have previously arrived.   Upon arrival of a node, an algorithm must irrevocably assign the node to a cluster.  In the popular \emph{competitive analysis} framework, the algorithm's clustering cost is compared to that of the optimal offline algorithm that is aware of the entire instance in advance.
Recall that no constant-competitive algorithm exists in the purely online setting for any $\ell_p$-norm objective of correlation clustering \cite{MathieuSS10}.
Clustering in general has received much attention in the online and streaming settings. For the popular class of $k$-clustering problems (including $k$-median, $k$-means, and $k$-center), which likewise face pessimistic lower bounds in the online setting, a popular remedy is recourse \cite{lattanzi2017consistent, jaghargh2019consistent,   fichtenberger2021consistent, guo2021consistent}. Likewise, to the best of our knowledge, the only previous beyond-worst-case results on correlation clustering in the online setting, besides the AOS model \cite{LattanziMVWZ21}, allow recourse \cite{Cohen-AddadLMP22, balkanski2025faironlineCC}. We note that the work of Balkanski, Chatzitheodorou, and Maggiori
\cite{balkanski2025faironlineCC} is similar in spirit to ours, as they also study fair correlation clustering in the online setting, but they use the traditional $\ell_1$-norm for the objective while adding fairness constraints, and allow recourse as their beyond-worst-case approach.

\subparagraph*{Prior work in the AOS model.} 
The AOS model was initiated by Kaplan, Naori, and Raz \cite{kaplan2020competitive} in the context of the secretary problem.\footnote{The analysis, however, differs from that presented here, in that we define competitiveness on the whole instance, whereas they restrict to the online portion of the input.} Lattanzi et al. \cite{LattanziMVWZ21} then used the AOS model for $\ell_1$-norm correlation clustering. They showed that the classic Pivot algorithm \cite{ACN-pivot}, modified to be seeded with an offline sample of size $\varepsilon n$, gives an $O(\nicefrac{1}{\eps})$-competitive algorithm, and this guarantee matches the lower bound up to constant factors. Since the works of Kaplan, Naori, and Raz \cite{kaplan2020competitive} and Lattanzi et. al. \cite{LattanziMVWZ21} in the AOS model, the model has been applied to   Steiner tree, load balancing, and facility location \cite{argue2022learning}; bipartite matching \cite{kaplan2022online}; and set cover \cite{gupta2024set}. A similar ``semi-online'' setting also appears in \cite{schild2019semi} for bipartite matching; like the AOS model, the semi-online model described there also contains predicted and adversarial parts of the input, but the predicted part is not necessarily a random sample, and further, the parts may be interleaved in an arbitrary manner.

\subsection{Technical overview}\label{sec: tech-overview}

Our primary challenge is to adapt an offline algorithm that requires complete, global knowledge of the graph to an online setting with limited information. The offline algorithm for all-norms correlation clustering due to Davies, Moseley and Newman \cite{davies2023one} relies on two offline-only steps. Step (1) is to compute a semi-metric $d^*$ over all vertex pairs; $d^*$ is an (almost) feasible solution to the canonical convex relaxation for the problem, but can be computed using explicit combinatorial properties of the graph. Step (2) is to, in place of an optimal solution $x^*$ to the relaxation, feed $d^*$ into the convex program rounding algorithm by Kalhan, Makarychev, and Zhou \cite{KMZ19} (from now on, the \emph{KMZ algorithm}). This is a ball-cutting procedure that iteratively cuts out clusters based on ``suggestions'' from $d^*$ (i.e., if $d^*_{uv}$ is small, then nodes $u$ and $v$ ``want'' to be clustered together). As in Step (1), the whole graph is required to determine the order in which clusters are cut out. So, both steps require significant changes in order to be adapted to the online setting.

To replace step (1), we compute a semi-metric $\tilde{d}$, only using the sample $S$, as a proxy for $d^*$. This will imply that when a  node $v$ arrives online, its distances $\tilde{d}_{uv}$ can immediately be computed for all $u$ that have already arrived.
To replace step (2), we note that the offline all-norms result holds (up to constants) when $d^*$ is fed into \emph{any} constant-approximate convex program rounding algorithm. We adapt the  offline rounding algorithm of Charikar, Gupta, and Schwartz \cite{CGS17} (from now on, the \emph{CGS algorithm}) instead of the KMZ algorithm, as we find the former easier to adapt to our online setting. The CGS algorithm is as follows: choose the unclustered node $v$ that has the most unclustered vertices in the ball of radius $r$ around it, then let $v$ and all unclustered nodes in its ball of radius $3r$ form a new cluster (where the balls are w.r.t the semi-metric that is the solution to the convex program). Notably, the algorithm is dynamic in that the sizes of the balls changes at each iteration, because nodes are removed when they are clustered. 

We next discuss some of the main technical challenges to computing a semi-metric based on the offline sample $S$, developing an online algorithm for $\ell_p$-norm correlation clustering, and analyzing our algorithm.

\paragraph*{Estimating a semi-metric via sampling.}
One key insight is that the semi-metric $d^*$ in \cite{davies2023one} -- which, crucially, is defined using explicit combinatorial properties of the graph -- can be approximated by computing it only on $G[S]$, the graph we see upfront on the random sample $S$. We first show this estimate, $\tilde{d}$, is of high quality (Section \ref{sec: guar-est}). Analyzing $\tilde{d}$ is non-trivial, as $\tilde{d}$ is defined based on non-linear calculations as well as on thresholds, both of which are highly sensitive to error. 

More specifically, in the offline world, $d^*$ is computed in two steps. Let $N_u^+$ be the set of positive neighbors of $u$. First,   intermediate distances, $d_{uv} = 1 -|N_u^+ \cap N_v^+| / |N_u^+ \cup N_v^+|$, are computed, and then some of these are rounded up to 1 based on thresholds. We estimate these intermediate distances using $S$, so we must work with quotients of correlated random variables which are prone to high variance and bias, especially for nodes with small positive neighborhoods. We then do threshold rounding with respect to these erroneous intermediate distances. This is technically problematic because, unlike the $\ell_1$ objective where local estimation errors tend to average out across the instance, $\ell_p$ and $\ell_\infty$ objectives are brittle. The ideal scenario would be to obtain pointwise bounds of the form $\mathbb{E}[\tilde{d}_{uv}] \approx d^*_{uv}$, which would in turn enable us to black-box previous results from \cite{davies2023one} bounding the cost of $d^*$ against optimal. This does not seem possible, so instead we develop a novel charging scheme. 

We pause to note that our technique, of using the sample $S$ to estimate combinatorial quantities of interest, is quite different from that for $\ell_1$-norm correlation clustering, which recall was previously studied in the AOS model \cite{LattanziMVWZ21}. That work adapts the Pivot algorithm, a 3-competitive algorithm in the RO model. In some sense, the Pivot algorithm is more readily adaptable to the AOS setting, since it is already an online algorithm, and the distribution of the (first $\eps$-fraction of) input matches that of the random sample. Our present work provides an example of how the AOS sample can be useful in other situations. In particular, we believe this general strategy of estimating quantities for an algorithm using the sample, and then charging the cost of an algorithm's objective to these estimates, is of broader interest for other problems in this model.

\paragraph*{Preprocessing the sample.}
    As discussed above, the sample $S$ is used to estimate $d^*$ with a proxy metric $\tilde{d}$. It is also used to adapt the CGS algorithm. The CGS algorithm treats all of $V$ as a set of candidate centers $v$, which are used to cut out clusters in the ball-cutting procedure. These are cut out in decreasing order of $|\ball_{x^*}(v,r)|$. The challenge in adapting this is that $V$ arrives online and adversarially. So, we restrict our centers, and the count of the ball sizes (computed now w.r.t. $\tilde{d}$) around those centers, to $S$.  
    Importantly, to avoid correlational issues arising from the fact that several sets / quantities are computed on $S$, we show how to simulate four independent subsamples on $S$. Then, we estimate different random variables of interest using different subsamples, rather than on the whole common sample $S$. The subsamples are seemingly essential for making the analysis tractable. 
 
\paragraph*{Feeding the proxy metric into a static, online version of the CGS algorithm.}
The algorithm has two phases. The first is an online version of the offline ball-cutting CGS algorithm. Vertices clustered in this phase are said to be \emph{pre-clustered}. The second phase handles the remaining vertices.

\emph{Pre-clustering phase:} When a vertex $v$ arrives, we first determine if the sample $S$ is trustworthy for $v$ via two checks. We first check whether the distances $\tilde{d}$ are sufficiently accurate for edges incident to $v$. If so, we then check if $v$ is close to one of the pre-selected centers.

    \begin{itemize}
        \item If both checks pass,  $v$ is assigned to a center that it is close to, particularly the earliest in the ordering of centers (see above). We note that to take the CGS algorithm online, this ordering is static, meaning it is pre-computed before the vertices arrive, unlike in the offline CGS algorithm, which dynamically orders the centers using the unclustered vertices at each iteration.  
        
        \item If the test fails, the algorithm falls back to one of two versions of the Pivot algorithm.
    \end{itemize}

\emph{(Modified) Pivot phases:} 
We perform the classic Pivot algorithm of \cite{ACN-pivot} on the vertices that fail the first check ($\tilde{d}$ is a poor estimator),
and a modified version of the Pivot algorithm on the vertices that only fail the second check ($\tilde{d}$ is a good estimator, but the cluster centers are poor). While in the classic Pivot algorithm pivots cluster their positive neighbors, our modified version restricts to positive neighbors that are close (w.r.t. $\tilde{d}$) to the pivot.

In both versions of the Pivot algorithm, our analysis looks different from that of classic Pivot. This is for two reasons. First, Pivot has previously only been used for the $\ell_1$-norm. It is not hard to find instances in which Pivot is $\Omega(n)$-approximate for the $\ell_\infty$ norm offline (see, e.g., Appendix A in \cite{PM16}). This is because the analysis of Pivot for the $\ell_1$-norm hinges on aggregating disagreements over bad triangles (triangles with two positive edges and one negative edge), which are configurations on which any clustering (thus an optimal one) must make a disagreement. But because $\ell_p$-norm objectives in general are node-wise rather than edge-wise, our charging to bad triangles is necessarily nonlocal. 

Second, Pivot crucially uses random order to ensure that disagreements in the optimal are not overcharged as a result of several bad triangles sharing the same edge. With adversarial order instead in our case, we show that overcharging can be controlled for a different reason, namely, that vertices that fail the first check have bounded positive degree.

For vertices that fail the second check, the argument is a bit more involved. Now, we have a generalization of bad triangles based not only on signs of edges but on $\tilde{d}$, so we need to take a hybrid approach and charge to both vanilla bad triangles and to the distances $\tilde{d}$. The rule that pivots grab \emph{nearby} positive neighbors is crucial for preventing overcharging: now, in lieu of the positive degree of vertices being bounded as is the case above, we can use that the number of vertices near a pivot must be bounded. (Otherwise, a nearby vertex would have been sampled into $S$, and the pivot would have been preclustered!)

\paragraph*{Lower Bounds.} Mathieu, Sankur, and Schudy showed that any strictly online algorithm for the $\ell_1$-norm objective must be at least $\Omega(n)$-competitive \cite{MathieuSS10}. It is not hard to see that this lower bound carries over to all $\ell_p$-norms, including $\ell_\infty$.

To establish our lower bounds for the RO model (Theorem \ref{thm: lowerbound-inf-ro}) and the AOS model (Theorem \ref{thm: lowerbound-inf}), we consider $k$ gadgets of the lower bound instance above, each of size $\frac{n}{k}$ (with $k$ set differently for each result). Crucially, for the $\ell_\infty$ objective, the offline optimal cost does not grow with $k$, unlike for finite $\ell_p$-norms.  For the RO model, we show that with constant probability there exists a gadget where the analogous $u,v$ in that gadget arrive first; then, the argument above can be repeated. For the AOS model, we show that, with constant probability, there is some gadget that arrives completely online, i.e., that is not touched by the sample; again, we can then repeat the argument above.

\subsection{Organization} In Section \ref{sec:prelim}, we discuss the AOS model and define the correlation metric and adjusted correlation metric. In Section \ref{sec: online-algo}, we define Algorithm \ref{alg: main-alg}, whose output satisfies Theorem \ref{thm:main-all}. We prove item \ref{item: thm-allp} of Theorem \ref{thm:main-all} (the statement for all finite $p$) in Sections \ref{sec: guar-est} and \ref{sec: lp-norm-analysis}, though a few proofs of lemmas and constructions stated in these sections are deferred to Appendix \ref{app: all-omits}. Then we prove item \ref{item: thm-1} of Theorem \ref{thm:main-all} (the statement for $p=1$) in Appendix \ref{app: omit-proofs-l1}, and item \ref{item: thm-infty} of Theorem \ref{thm:main-all} (the statement for $p=\infty$) in Appendix \ref{sec: linf}.
Lastly,
the lower bound results, Theorems \ref{thm: lowerbound-inf} and \ref{thm: lowerbound-inf-ro}, are in Appendix \ref{sec: LB}.

\section{Preliminaries}
\label{sec:prelim}

Let $G=(V,E)$ be a complete graph, where $E$ is partitioned into positive edges ($E^+$) and negative edges ($E^-$). 
Let $N_u^+$ and $N_u^-$ denote the positive and negative neighborhoods, respectively, of vertex $u$. That is, 
$N_u^+ = \{v \in V: uv \in E^+\}$ and
$N_u^- = \{v \in V: uv \in E^-\}.$
For convenience, assume that each vertex has a positive self-loop, i.e., for all $u \in V$, $u \in N_u^+$. 

Recall that $\textsf{OPT}_p$ is the optimal objective value of an integral solution for the $\ell_p$-norm objective, where here $p \in [1, \infty]$. 

We set some parameters. Let $\delta = 10/7$, and define $c=c(\delta) := 2\delta^2 + \delta = 270/49 \textrm{ and } r=r(\delta) := \frac{1}{2c\delta^2} = \frac{2401}{54000}.$

\subsection{The AOS model}

In the AOS model, an adversary fixes the online input, then
we are given a sample $S$, where each element of the universe $V$ is in $S$ independently with probability $\eps>0$. We assume $\varepsilon$ is known.\footnote{One can immediately remove this assumption if we consider an ``online-with-samples" model, suggested in the next footnote.} So for our problem, we see the induced subgraph on $S$ a priori.

We will estimate various quantities using the sample $S$, and, for the analysis to be tractable, these estimations should not be correlated with each other. To this end, we show how to ``split" the sample $S$ into four independent subsamples $S_p, S_d, S_b, S_r$ (where vertices in $S$ may be in more than one subsample). We define this notion of independence more precisely in Appendix \ref{app:omit-prelim}, but it will suffice to think of these subsamples as having the same joint distribution as four random subsets of vertices, each obtained by taking a uniformly random sample of (expected) size $\Theta(\varepsilon^2 n)$, and repeating this procedure \emph{independently} four times.\footnote{Alternatively, one could consider a type of ``online-with-samples" model, where an algorithm is given access to a constant number $k$ of independent random samples of size $\eps_i \cdot n$ of $V$, where $\sum_{i \in [k]} \eps_i = \eps$. We find this to be a perfectly reasonable model, and the reader may find it simpler to assume this model.}

We defer the construction of the four independent subsamples $S_p, S_d, S_b, S_r$, and the precise notions of independence they satisfy, to Lemma \ref{lem: simulate-sample} and Corollary \ref{cor: subsample-independence} in Appendix \ref{app:omit-prelim}.

\begin{definition}
     We call $S_d$ the \emph{distance sample}, $S_p$ the \emph{pre-clustering sample}, $S_b$ the \emph{counting sample}, and $S_r$ the \emph{rounding sample}. We call the vertices in $S_p$ \emph{centers}.
\end{definition}

 Since the four samples are constructed from $S$, edges in the (complete) subgraph induced by $S_d \cup S_p \cup S_b \cup S_r$ are known to the online algorithm a priori. Then, the remaining vertices arrive one by one in adversarial order. When a vertex arrives, it reveals the signs of its incident edges to all vertices that have previously arrived (including all of $S$), and it must be irrevocably assigned a cluster.

  We define $q(\varepsilon) :=\mathbb{P}[v \in S_i] =  \varepsilon^2/2$ for any subsample $S_i \in \{S_d,S_p,S_b,S_r\}$, per the construction in Lemma \ref{lem: simulate-sample}.
 
\subsection{Correlation metric, adjusted correlation metric, and their estimates}

The \emph{correlation metric} is a near-optimal feasible solution to the canonical convex program\footnote{See \cite{davies2023one} for more details on this convex program which, while motivating, is not necessary for understanding the work herein.} for $\ell_p$-norm correlation clustering. 
The correlation metric $d$ is feasible for this convex program, meaning here that it  satisfies the triangle inequality. So we may think of $d_{uv}$ as specifying a distance between $u$ and $v$, where the smaller the distance, the more that $u$ and $v$ would like to be clustered together. Relevant theorems on the (adjusted) correlation metric from \cite{DMN23, davies2023one} are in Section \ref{sec: guar-est}. 

Below, we generalize the definition of the correlation metric by defining it on a subgraph of $G$. Taking $U = V$ below recovers the correlation metric in \cite{DMN23}.

\begin{definition}[Correlation metric] \label{def: corr}
    Let $G = (V,E)$ be a complete, signed graph, and let $U \subseteq V$. For every (unordered) pair $u,v \in V$, define the \emph{correlation metric on $U$}, denoted $d^U: V \times V \to [0,1]$, by 
\[d_{uv}^U := 1 - \frac{|N_u^+ \cap N_v^+ \cap U|}{|(N_u^+ \cup N_v^+) \cap U|}.\]
To make this well-defined, we take $d^U_{uv}=1$ if $(N_u^+ \cup N_v^+) \cap U = \emptyset$ for $u \neq v$, and $d^U_{uu} = 0$ always. When $U = V$, we simply write $d$ for $d^V$ and refer to $d$ as the \emph{correlation metric}. 
\end{definition}

The correlation metric distances are near-optimal for the convex program with the $\ell_\infty$-norm objective (see Theorem \ref{thm: corr-metric-cost}), but the metric must be adjusted in order to be simultaneously near-optimal for all $\ell_p$-norms. We likewise generalize the definition of the so-called \emph{adjusted correlation metric} in \cite{davies2023one}; taking $U = W = V$ recovers their definition.

\begin{definition}[Adjusted correlation metric] \label{def: adj-corr}
    Let $G = (V,E)$ be a complete, signed graph, and let $U, W \subseteq V$. Let $d^U$ be the correlation metric on $U$ as in Definition \ref{def: corr}. Compute the \emph{adjusted correlation metric on $U$ and $W$}, denoted $d^{U,W}: E \to [0,1]$, as follows:
\begin{itemize}
    \item If $uv \in E^-$ and $d_{uv}^U > \nicefrac{7}{10}$, set $d^{U,W}_{uv} = 1$. (We say $uv$ is \emph{rounded up}.) 
    \item For $u \in V$ such that $| \{v \in N_u^-: d_{uv}^U \leq \nicefrac{7}{10}\} \cap W| \geq \nicefrac{10}{3} \cdot |N_u^+ \cap U| $, set $d^{U,W}_{uv}=1$ for all $v \in V \setminus \{u\}$. (We say $u$ is \emph{isolated} by $d^{U,W}$.)
\end{itemize}
When $U = W = V$, we write $d^*$ for $d^{V, V}$ and refer to $d^*$ as the \emph{adjusted correlation metric}.
\end{definition}

A key takeaway is that the correlation metric and the adjusted correlation metric, while intended as surrogates for \emph{non-combinatorial} optimal solutions to a convex program,  are based solely on \emph{combinatorial} properties of the graph -- thus making them more amenable to the online setting. In the AOS model, we obviously cannot exactly compute the correlation metric or the adjusted correlation metric as we go, as the full positive neighborhood of a vertex is not necessarily known upon its arrival. But, for any $u,v \in V$, we \emph{can} compute, e.g., $d^{U}_{uv}$ in the case that $U$ is a subset of $S$, \emph{as soon as $u,v$ have arrived}, since $S$ is known to the algorithm upfront! The hope is that $d^{U}$ and $d^{U,W}$ should be good approximations of $d = d^V$ and $d^* = d^{V,V}$, respectively, since $S$ is a random sample of $V$ -- while also being usable online.

To estimate the adjusted correlation metric, we proceed in two steps. First, we estimate the correlation metric using the distance sample $S_d$; call this $\bar{d}$. Then, using the rounding sample $S_r$, we round $\bar{d}$ to estimate the adjusted correlation metric; call this $\tilde{d}$.

\begin{definition}[Estimated correlation metric] \label{def: est-corr}
    For every (unordered) pair $u,v \in V$, define the \emph{estimated correlation metric} $\bar{d}: V \times V \to [0,1]$ by $\bar{d}_{uv} := d^{S_d}_{uv}.$
\end{definition}

\begin{definition}[Estimated adjusted correlation metric] \label{def: est-adj-corr}
Let $\bar{d}$ be the estimated correlation metric as in Definition \ref{def: est-corr}. Define the \emph{estimated adjusted correlation metric} $\tilde{d}: E \to [0,1]$ by $\tilde{d}_{uv} := d_{uv}^{S_d, S_r}.$\\
\noindent
Moreover, we let $R_1$ be the (random) subset of $V$ for which bullet 2 of Definition \ref {def: adj-corr} applies (i.e., the set of vertices that are isolated by $\tilde{d}$), $R_2 = V \setminus R_1$, and  $R_1(u):= \{v \in N_u^- : \bar{d}_{uv} \leq \nicefrac{7}{10}\}$.
\end{definition}

The next observation ensures our algorithm for the AOS model is an online algorithm.

\begin{observation} \label{obs: corr-computability}
    As soon as both $u$ and $v$ have arrived (including if one or both is in $S$), $\bar{d}_{uv}$ and $\tilde{d}_{uv}$ can be computed. 
\end{observation}
We see that if $u$ has no positive neighbors in the sample $S_d$, then $\tilde{d}$ isolates $u$ from all other vertices. 

\begin{fact} \label{fct: frac-isolation}
Fix $u \in V$ such that $N_u^+ \cap S_d = \emptyset$. Then $\tilde{d}_{uv} = 1$ for all $v \neq u$. 
\end{fact}

Both $\bar{d}$ and $\tilde{d}$ enjoy similar properties to $d$ and $d^*$, respectively, in that they are a semi-metric and near semi-metric, respectively. Thus, we can still view $\bar{d}$ and $\tilde{d}$ as specifying distances between vertices.

\begin{definition}
We say a symmetric function $f: V \times V \to \mathbb{R}_+$ is a $\delta$-\emph{semi-metric} if $f_{uv} \leq \delta \cdot (f_{uw} + f_{wv})$ for all $u,v,w \in V$ (along with the usual requirement that $f_{uu} = 0$). We say in this case that $f$ satisfies an \emph{approximate triangle inequality}, or $f$ is a \emph{near semi-metric}.
\end{definition}

\begin{restatable}{lemma}{trineq} \label{lem: tri-inequality}
    Let $\bar{d}$ and $\tilde{d}$ be as in Definitions \ref{def: est-corr} and \ref{def: est-adj-corr}, respectively. Then 
    \begin{itemize}
        \item $\bar{d}: V \times V \to [0,1]$ is a $1$-semi-metric, that is, $\bar{d}$ satisfies the triangle inequality. 
        \item $\tilde{d}: V \times V \to [0,1]$ is a $\frac{10}{7}$-semi-metric. 
    \end{itemize}
\end{restatable}

\subsection{Ordering the centers}

 Given a map $f: V \times V \to [0,1]$ on the vertices of $G$, for $c \in V$, $U \subseteq V$, and $\rho \geq 0$ we define:
 $\ball_f(c, \rho) := \{v \in V : f_{cv} \leq \rho\}$ and $\ball^{U}_f(c, \rho) := \{v \in U: f_{cv} \leq \rho \}.$

\begin{definition}[Density] \label{def: true-density}
   Given a semi-metric $f: V \times V \to [0,1]$ and vertex $c \in V$, define 
    $|\ball_f(c, r)|$
    to be the \emph{density of $c$} w.r.t $f$ and $r$. 
\end{definition}

A subroutine of our algorithm will be an adaptation of the CGS algorithm (see Section \ref{sec: tech-overview}). This algorithm takes as input a metric $f$ on the vertices and orders the vertices in decreasing order of their densities with respect to $f$ and some radius $r$. Due to our online setting, we will only be able to estimate these densities, which we do using the counting sample $S_b$. 

We note that $|\ball_{\tilde{d}}^{S_b}(c, r) |$ depends on the randomness of $S_b$, $S_d$, and $S_r$. 
The following observation will ensure our algorithm is well-defined.
\begin{observation} \label{obs: density-computability}
    For any center $c \in S_p$, the  density $|\ball_{\tilde{d}}^{S_b}(c, r) |$ can be computed using only the information given \emph{a priori} in the AOS model, i.e., the sample $S$.
\end{observation}

Lastly, we order the centers $S_p$ based on their estimated densities. 

\begin{definition}[Ordered center sample] \label{def: pivot-ordering}
    Let $S_p = \{u_1, \dots, u_{|S_p|}\}$. We assume the $u_i$ are labeled so that  
$|\ball_{\tilde{d}}^{S_b}(u_1, r) | \geq \cdots \geq 
|\ball_{\tilde{d}}^{S_b}(u_{|S_p|}, r) |$
(with ties broken arbitrarily). 
\end{definition}

\section{Algorithm Description} \label{sec: online-algo}

Now we are ready to describe our main algorithm (Algorithm \ref{alg: main-alg}) for the AOS model. 
Note that the \textsf{for} loop of Algorithm \ref{alg: main-alg} considers the vertices in $S$ in arbitrary order. So, we assume the algorithm considers the vertices in $S$ first, and then the vertices in $V \setminus S$ as they arrive online. 

Algorithm \ref{alg: main-alg} is an online version of the offline CGS algorithm \cite{CGS17} (see the first two paragraphs in Section \ref{sec: tech-overview}). The CGS algorithm must be adapted in several important ways in order to be taken online. First, the CGS algorithm takes as input an optimal solution to the convex program for the $\ell_p$-norm objective of interest. Since we are not given the graph upfront, this is impossible to compute on the fly. Moreover, the convex program requires specifying an $\ell_p$-norm objective, whereas we would like to optimize for all $\ell_p$-norms simultaneously. In place of this optimal solution, we use $\tilde{d}$, which \emph{can} be computed on the fly (Observation \ref{obs: corr-computability}), and does \emph{not} depend on $p$.

\begin{figure}
  \makebox[\linewidth]{%
  \scalebox{0.9}{\begin{minipage}{\dimexpr\linewidth-7em}
\begin{algorithm}[H]
\setstretch{.7}
\caption{Main Algorithm}\label{alg: main-alg}
\begin{algorithmic}
    \State {\small \textbf{Input: }  $G=(V, E)$, $S_d$,  $\tilde{d}$,  \emph{ordered}  $S_p = \{u_1, \dots, u_{|S_p|}\}$, radius $r$, constant $c$}
    \State {\small \textbf{Initialize: }  empty clusters $\mathcal{C}_{\textsf{ALG}} = \{C_1, \dots, C_{|S_p|}\}$, sets  $V_0 = V$ and $V' = \emptyset$ }
    \For{ {\small each arriving $v \in V$}}
   \State \hspace{-15pt} 
    \textsf{\quad {\small// Pre-clustering phase}} \If{ {\small $|N_v^+ \cap S_d| \neq \emptyset$ and there exists $u_i \in S_p$ such that $\tilde{d}_{u_i v} \leq c \cdot r$}}
    \State {\small Let $i^*$ be the earliest in the ordering among all such $u_i$ }
    \State {\small Add $v$ to cluster $C_{i^*}$}
    \State \hspace{-19pt}\textsf{ {\small // Pivot phase}}
    \Else 
    \State {\small Add $v$ to $V'$}
    \If{ {\small $v$ has $|N_v^+ \cap S_d| = \emptyset$}}
    \State {\small Remove $v$ from $V_0$, order $\overline{V}_0= V \setminus V_0$ by arrival order}
    \State {\small Set $E_c := E^+(G[\overline{V}_0])$}
    \State  {\small Add $v$ to $\mathcal{C}_{\textsf{ALG}}$ by running $\textsf{ModifiedPivot}(G[\overline{V}_0], \mathcal{C}_{\textsf{ALG}}, \overline{V}_0 \setminus \{v\}, E_c)$ }
    \Else
    \State {\small Order $V' \setminus \overline{V}_0$ by arrival order }
    \State {\small Set $E_c := E^+(G[V'\setminus \overline{V}_0]) \cap \{uv \in E : \tilde{d}_{uv} < c \cdot r \}$}
    \State {\small Add $v$ to $\mathcal{C}_{\textsf{ALG}}$  by running $\textsf{ModifiedPivot}(G[V' \setminus \overline{V}_0], \mathcal{C}_{\textsf{ALG}}, V' \setminus (\overline{V}_0  \cup \{v\}), E_c)$}
  \EndIf  
  \EndIf
    \EndFor
    \State {\small  \textbf{Output: } Clustering $\mathcal{C}_{\textsf{ALG}}$}
\end{algorithmic}
 \end{algorithm}
    \end{minipage}}}
\end{figure}

The CGS algorithm requires an ordering on \emph{all} of $V$ based on the ball densities around the vertices. We are only able to estimate these densities for vertices in the offline sample, so we use the ordered sample of centers $S_p$ (Definition \ref{def: pivot-ordering}), and $V \setminus S_p$ remains unordered. In the offline CGS algorithm, each vertex in $V$ is clustered by the earliest vertex in the ordering that is nearby (thus by itself if necessary). In our setting, we cannot guarantee every vertex will be clustered because only $S_p$ is ordered. Further, $\tilde{d}_{uv}$ is meaningless as a distance when $u$ or $v$ does not have positive neighbors in $S_d$ (see Definition \ref{def: corr}). Thus, we will not cluster every vertex in this way, and instead throw the vertices that are \textit{not} ``pre-clustered'' (for either of these two reasons) to a subroutine called \textsf{ModifiedPivot} (Algorithm \ref{alg: pivot}).

\textsf{ModifiedPivot} is a generalization of the Pivot algorithm from \cite{ACN-pivot}. Taking $E_c = E^+$ in Algorithm \ref{alg: pivot} (as we do in the first call to \textsf{ModifiedPivot} in Algorithm \ref{alg: main-alg}) recovers the original Pivot algorithm. In the second call to \textsf{ModifiedPivot} in Algorithm \ref{alg: main-alg}, we further restrict $E_c$ to edges that are ``short'' according to $\tilde{d}$; this is not simply an optimization, but seemingly needed in the analysis.

\begin{figure}
  \makebox[\linewidth]{%
   \scalebox{0.9}{\begin{minipage}{\dimexpr\linewidth-7em}
\begin{algorithm}[H]
\setstretch{0.7}
\caption{$\textsf{ModifiedPivot}(H, \mathcal{C}_{\textsf{ALG}}, P, E_c)$}\label{alg: pivot}
\begin{algorithmic}
    \State {\small  \textbf{Input:}  $H=(V_H,E_H)$ where $V_H$ is \emph{ordered}, $\mathcal{C}_{\textsf{ALG}}$ is a clustering on $P \subseteq V_H$, 
    and $E_c \subseteq E_H$ is a set of \emph{clusterable} edges  }
    \For{ {\small each $v_i \in V_H \setminus P$ (in order)}}
    \If{ {\small there exists $v_j$ that is before $v_i$ in the ordering with $v_jv_i \in E_c$}}
    \State {\small Let $j^*$ be the earliest in the ordering among all such $v_j$}
    \State {\small Add $v_i$ to the same cluster as $v_{j^*}$}
    \Else{}
    \State {\small Open a new cluster and add $v_i$ to it}
    \EndIf
    \EndFor
    \State {\small  \textbf{Output: } Clustering $\mathcal{C}_{\textsf{ALG}}$}
\end{algorithmic}
 \end{algorithm}
   \end{minipage}}}
\end{figure}

\subsection*{Terminology} \label{sec: terminology}
If the \textsf{else} statement in Algorithm \ref{alg: pivot} holds, we say $v_i$ is a \textit{pivot}, and that $v_i$'s pivot is itself, $v_i$. If the \textsf{if} statement holds, we refer to $v_{j^*}$ as $v_i$'s \textit{pivot}.

We say $v$ is \emph{pre-clustered} by $u_i$ if $v$ is added to $C_i$.
Otherwise, $v \in V'$, and we say $v$ is \emph{unclustered} or \emph{not pre-clustered}. Note that a vertex $v$ may be unclustered for one of two reasons: either $v$ has no positive neighbors in $S_d$ ($v \not \in V_0$ or as written in Algorithm \ref{alg: main-alg}, $v \in \overline{V}_0$, where $\overline{V}_0 = V \setminus V_0$), or $v \in V_0$ but $v$ is not close to any vertex in $S_p$ with respect to $\tilde{d}$. 
We call the vertices in $V_0$ \textit{eligible for pre-clustering}, or simply \textit{eligible}, and otherwise \textit{ineligible}.

If $v$ is pre-clustered by $s$, we denote $s$ by $s^*(v)$. We may refer to $s^*(v)$ as $v$'s \emph{center}. 
We say \emph{$u$ is clustered after $v$} or \emph{$v$ is clustered before $u$} if either both $u$ and $v$ are pre-clustered, but $s^*(v)$ is before $ s^*(u)$ (w.r.t the ordering of $S_p$), or if $v$ is pre-clustered but $u$ is not.  Notationally, this will denoted as $v \succ u$. 

\paragraph{The ``good" event.} For the cost analysis of Algorithm \ref{alg: main-alg} when $p \neq 1$ (Sections \ref{sec: guar-est}, \ref{sec: lp-norm-analysis}, and \ref{sec: linf}), we condition on a certain \emph{good event}, denoted $B^c$, that occurs with high probability (Lemma \ref{lem:good-whp}). For the formal definition of the good event, see Definition \ref{def: good-event}. Informally, the event $B^c$ ensures via concentration that the size of
sufficiently large sets of vertices can be well-estimated based on observing their intersection with the subsamples.  It also ensures that for pairs of nodes $u,v$ with large combined positive neighborhood, 
$\bar{d}_{uv}$ is a good estimate of $d_{uv}$.

\section{Fractional cost of correlation metric}\label{sec: guar-est}

In the offline setting, the algorithm has access to the \textit{true} correlation metric $d$ and adjusted correlation metric $d^*$.
We overview some notation and previous results that will be useful to us.

 \subsection{Preliminaries}

\begin{definition}[Fractional cost] \label{def: frac-cost}
    For $u \in V$, the \emph{fractional cost } of $u$ w.r.t $d$ (Definition \ref{def: corr}) is
    $D(u) := \sum_{v \in N_u^+} d_{uv} + \sum_{v \in N_u^-} (1-d_{uv}),$ and $D = (D(u))_{u \in V}$ is the vector of fractional costs of vertices. $D^*(u)$ and $D^*$ are defined analogously using $d^*$ (Definition \ref{def: adj-corr}).

   \smallskip

\noindent  Similarly, $\bar{D}(u)$ and $\bar{D}$ are defined analogously using $\bar{d}$ (Definition \ref{def: est-corr}), and $\tilde{D}(u)$ and $\tilde{D}$ using  $\tilde{d}$ (Definition \ref{def: est-adj-corr}), though these are \emph{estimated fractional cost} w.r.t the semi-metrics.
\end{definition}

The key result of Davies, Moseley, and Newman \cite{DMN23} is that the fractional cost of $d$ in the $\ell_\infty$-norm is a constant approximation to the optimal solution for the $\ell_\infty$-norm.  We will black-box this result in part of our analysis.

\begin{theorem}[Lemma 4.2 in \cite{DMN23}] \label{thm: corr-metric-cost}
    Let $D$ be as in Definition \ref{def: frac-cost}. Then 
    $||D||_\infty \leq 8 \cdot \textsf{OPT}_\infty.$
\end{theorem}

While the correlation metric must be modified to give bounded fractional cost for other $\ell_p$-norms, the following claim states that the correlation metric actually has bounded frational cost for the $\ell_1$-norm objective \emph{when restricted to positive edges}. This will be useful in our later analysis. 

\begin{lemma}[Lemma 8, Claim 1 in \cite{daviesArXiv}] \label{lem: bdd-pos-frac}
    $\sum_{u \in V} \sum_{v \in N_u^+} d_{uv} \leq 3 \cdot \textsf{OPT}_1$.
\end{lemma}

In the follow-up work \cite{davies2023one}, the authors show that taking $x = d^*$ gives bounded fractional cost for \textit{all} $\ell_p$-norms simultaneously (and thus a \textit{simultaneous} approximation for all $\ell_p$-norms). While we are not able to black-box this result often, it does help in bounding the cost of disagreements in some settings.

\begin{theorem}[Lemma 4 in \cite{davies2023one}] 
\label{thm: all-norms-offline}
    For $D^* = (D^*(u))_{u \in V}$ the fractional cost vector of the adjusted correlation metric $d^*$, there exists a universal constant $M$ (independent of $p$) such that,  for all $p \in [1, \infty]$,
    \[||D^*||_p \leq M \cdot \textsf{OPT}_p.\]
\end{theorem}

\subsection{Fractional cost of correlation metric for finite $p$} \label{sec: frac-cost-finite-p}

Recall that $V_0$ is the random set consisting of all vertices $u \in V$ with $N_u^+ \cap S_d = \emptyset$. 

In this subsection, we show that the fractional cost of the estimated adjusted correlation metric $\tilde{d}$ in the $\ell_p$-norm, \emph{restricted to $V_0$}, is bounded in expectation against $\textsf{OPT}_1$ for $p=1$, and bounded with high probability against $\textsf{OPT}_p$ for finite $p$. This will allow us to charge the cost of some disagreements incurred by Algorithm \ref{alg: main-alg} to $\tilde{d}$; other disagreements will require different surrogates for optimal.

For $p > 1$, we condition on the good event $B^c$ (see Section \ref{sec: terminology}).

\begin{restatable}{lemma}{whpposfraccost} 
\label{lem: whp-pos-frac-cost}
Let $1 \leq p < \infty$. The estimated adjusted correlation metric $\tilde{d}$ satisfies the following:
\begin{itemize}
    \item $\mathbb{E}\left[\sum_{u \in V_0} \sum_{v \in N_u^+ \cap V_0} \tilde{d}_{uv} \right] \leq O (\nicefrac{1}{\eps^4} )\cdot \opt_1$, and 
    \item Conditioned on the event $B^c$, $\sum_{u \in V_0} \left(\sum_{v \in N_u^+ \cap V_0} \tilde{d}_{uv}\right)^p\leq O \left (\left ( \nicefrac{1}{\eps^6}   \cdot \log^3 n \right ) ^p \right )\cdot \opt_p^p.$
\end{itemize}

\end{restatable}

\begin{lemma} \label{lem: whp-neg-frac-cost}
Let $1 \leq p < \infty$. The estimated adjusted correlation metric $\tilde{d}$ satisfies the following:
\begin{itemize}
    \item $\mathbb{E}\left[\sum_{u \in V_0} \sum_{v \in N_u^- \cap V_0} (1-\tilde{d}_{uv}) \right] \leq O(\nicefrac{1}{\eps^2}) \cdot \opt_1$, and 
    \item Conditioned on the event $B^c$, $\sum_{u \in V_0} \left (\sum_{v \in N_u^- \cap V_0} (1-\tilde{d}_{uv}) \right )^p \leq O \left (\left ( \nicefrac{1}{\eps^2}\cdot \log n \right )^p\right ) \cdot \opt_p^p.$
\end{itemize}
\end{lemma}

For $u \in V_0$, let $\tilde{D}_0(u)$ be the fractional cost with respect to $\tilde{d}$ of the edges incident to $u$ in the subgraph induced by $V_0$, i.e.,
$\tilde{D}_0(u):= \sum_{v \in N_u^+ \cap V_0} \tilde{d}_{uv} + \sum_{v \in N_u^- \cap V_0} (1-\tilde{d}_{uv}). $ By Lemmas \ref{lem: whp-pos-frac-cost} and \ref{lem: whp-neg-frac-cost}, we obtain the following corollary. 

\begin{corollary} \label{cor: bounded-exp-frac-cost}
Let $1 \leq p < \infty$. The estimated adjusted correlation metric $\tilde{d}$ satisfies the following:
\begin{itemize}
    \item $\mathbb{E}[\sum_{u \in V_0} \tilde{D}_0(u)] \leq O(\nicefrac{1}{\eps^4}) \cdot \opt_1$, and
    \item Conditioned on the event $B^c$, we have $\sum_{u \in V_0} \left(\tilde{D}_0(u)\right)^p\leq O\left(\left ( \nicefrac{1 }{\eps^6} \cdot \log^3 n\right ) ^p  \right )\cdot \opt_p^p.$
    \end{itemize}
\end{corollary}

We prove Lemma  \ref{lem: whp-pos-frac-cost} in Appendix \ref{sec: pos-frac-cost} and  Lemma \ref{lem: whp-neg-frac-cost} in this section, as both proofs are similar in flavor, but the proof of Lemma \ref{lem: whp-neg-frac-cost} serves us again in Section \ref{subsec: lp-neg-edge}. We give a brief overview here of both proofs. We will not be able to prove a pointwise bound comparing $\mathbb{E}[\tilde{d}_{uv}]$ to, e.g., $d^*_{uv}$, but instead, we will have to unbox the offline proof from \cite{davies2023one} that the fractional cost of the adjusted correlation metric is bounded (see Theorem \ref{thm: all-norms-offline} above). While the casing will be similar to in the offline case, there are nontrivial technical details, e.g., independence issues, taking expectations of sums defined over random sets, and taking expectations over quotients, to name a few, that must be handled delicately.

Recall that
$q(\varepsilon) =\mathbb{P}[v \in S_i] =  \varepsilon^2/2$ for any subsample $S_i \in \{S_d,S_p,S_b,S_r\}$.
In the following three lemmas, we show that, with some constant factor loss, we can distribute an expectation over a quotient for certain random variables that will be of interest.

\begin{lemma} \label{lem: exp-reciprocal}
Fix a subsample $S_* \in \{S_d,S_b,S_r,S_p\}$ and fix $\mathcal{B} \subseteq \{S_d,S_b,S_r,S_p\} \setminus \{S_*\}$. Let $A \subseteq V$ be a subset of vertices that may depend on the randomness of the sets in $\mathcal{B}$, but it does not depend on the randomness of any other subsamples. Define the random variable $ Z = \mathbf{1}_{\{|A \cap S_*| \geq 1\}} \cdot \frac{1}{|A \cap S_*|},$ where $Z$ is defined to be 0 if $A \cap S_* = \emptyset$. Then, \[\mathbb{E} [Z \mid \mathcal{B}] \leq O(\nicefrac{1}{\varepsilon^2 })\cdot \frac{1}{ |A|}
\quad \text{and} \quad \mathbb{E} [Z^2 \mid \mathcal{B}] 
\leq O(\nicefrac{1}{\varepsilon^4 })\cdot\frac{1}{ |A|^2},\]
where the expectation is taken only over the randomness of $S_*$.
\end{lemma}

Note in the above that if $\mathcal{B}=\emptyset$, then $\mathbb{E}[Z \mid \mathcal{B}]$ is a deterministic value, and otherwise $\mathbb{E}[Z \mid \mathcal{B}]$ is a random variable.

\begin{proof}[Proof of Lemma \ref{lem: exp-reciprocal}]
As in the lemma statement,
$ Z = \mathbf{1}_{\{|A \cap S_*| \geq 1\}} \cdot \frac{1}{|A \cap S_*|},$ where $Z$ is defined to be 0 if $A \cap S_* = \emptyset$. 
Define $Y := |A \cap S_*| = \sum_{w \in A} X_w$, where $X_w$ is 1 if $w \in S_*$ and 0 otherwise. Let $\mu := \mathbb{E}[Y] = q(\varepsilon) \cdot |A|$, where the expectation is taken only over the randomness of $S_*$. Again taking the expectation only over the randomness of $S_*$, we have 
\begin{align}
   \mathbb{E}[Z \mid \mathcal{B}] &= \mathbb{E}\Big[\frac{1}{|A \cap S_*|} \cdot \mathbf{1}_{\{|A \cap S_*| \geq 1\}} \mid \mathcal{B}\Big] \notag \\
   &= \sum_{\ell=1}^{|A|} \mathbb{P}[Y=\ell \mid \mathcal{B}] \cdot \frac{1}{\ell} 
   = \sum_{\ell=1}^{|A|} \mathbb{P}[Y=\ell ] \cdot \frac{1}{\ell} \label{eq: explicit-sum}\\
    &\leq \mathbb{P}[Y \leq \mu/2] + \sum_{\ell= \lceil \mu/2 \rceil}^{|A|} \mathbb{P}[Y=\ell] \cdot \frac{1}{\ell} \notag \\
    &\leq e^{-\mu/12} + \frac{2}{\mu} \cdot \sum_{\ell= \lceil \mu/2 \rceil}^{|A|} \mathbb{P}[Y=\ell] 
    \leq \frac{5}{\mu} + \frac{2}{\mu} = \frac{7}{q(\varepsilon) \cdot |A|} = \frac{14}{\varepsilon^2 \cdot |A|}. \notag
\end{align}
In the second line,  if $|A| = 0$, we treat the sum as 0. Further, the first equality in (\ref{eq: explicit-sum}) follows from the fact that $A$ does not depend on the randomness of $S_*$, so we may move the sum outside of the probability and then we use Corollary \ref{cor: subsample-independence} for the next equality (specifically saying that $\mathbb{P}[|A \cap S_*| = \ell \mid \mathcal{B}] = \mathbb{P}[|A \cap S_*| = \ell ]$).
In the last line, we applied a Chernoff bound (Theorem \ref{thm: chernoff}).
The bound for $\mathbb{E}[Z^2 \mid \mathcal{B}]$ is derived similarly: starting in line (\ref{eq: explicit-sum}), replace $\nicefrac{1}{\ell}$ with $\nicefrac{1}{\ell^2}$, and continuing as before we obtain a bound of $e^{-\mu/12} + 4/\mu^2 \leq 83/\mu^2 = \nicefrac{332}{\eps^4 \cdot |A|^2}$. 
\end{proof}

\begin{proposition}\label{prop: reciprocal-corr}
Define $\mathbf{1}_{\{u \in V_0\}} \cdot\frac{|N_u^+|}{|N_u^+ \cap S_d|}$ to be 0 if $u \not \in V_0$, i.e., if $|N_u^+ \cap S_d|=0$. Likewise, define $\mathbf{1}_{\{|R_1(u) \cap S_r| \geq 1\}} \cdot\frac{|R_1(u)|}{|R_1(u) \cap S_r|}$ to be 0 if $|R_1(u) \cap S_r| = 0$. Then
\begin{itemize}
    \item $\mathbb{E}\left[\mathbf{1}_{\{u \in V_0\}} \cdot\frac{|N_u^+|}{|N_u^+ \cap S_d|} \right] 
     \leq O(\nicefrac{1}{\eps^2})$
    , and $\mathbb{E}\left[\mathbf{1}_{\{u \in V_0\}} \cdot\frac{|N_u^+|^2}{|N_u^+ \cap S_d|^2} \right]   \leq O(\nicefrac{1}{\eps^4})$
    \item $\mathbb{E}\left[\mathbf{1}_{\{|R_1(u) \cap S_r| \geq 1\}} \cdot\frac{|R_1(u)|}{|R_1(u) \cap S_r|} \right]\leq O(\nicefrac{1}{\eps^2})$
    \item Conditioned on the event $B^c$, we have $\mathbf{1}_{\{u \in V_0\}} \cdot\frac{|N_u^+|}{|N_u^+ \cap S_d|}  \leq O(\nicefrac{1}{\varepsilon^2})\cdot \log n$
    \item Conditioned on the event $B^c$, we have $\mathbf{1}_{\{|R_1(u) \cap S_r| \geq 1\}} \cdot\frac{|R_1(u)|}{|R_1(u) \cap S_r|} 
    \leq O(\nicefrac{1}{\varepsilon^2})\cdot \log n$.
\end{itemize}
\end{proposition}

\begin{proof}[Proof of Proposition \ref{prop: reciprocal-corr}]
    The bounds in the first bullet follow directly from Lemma \ref{lem: exp-reciprocal}, taking $\mathcal{B} = \emptyset$, $S_* = S_d$, and $A = N_u^+$. The bound in the second bullet likewise follows from Lemma \ref{lem: exp-reciprocal}, taking $\mathcal{B}= \{S_d\}$, $S_* = S_r$, and $A = R_1(u)$, and then applying Law of Total Expectation.
    
    For the third bullet, since $B^c$ holds, we know that if $|N_u^+| \geq C \cdot \log n / \varepsilon^2$, then  $|N_u^+ \cap S_d| \geq \frac{\varepsilon^2}{4} \cdot |N_u^+|$, so $\frac{|N_u^+|}{|N_u^+ \cap S_d|} \leq \frac{4}{\varepsilon^2}$. Otherwise, we use that $u \in V_0$ to bound $\frac{|N_u^+|}{|N_u^+ \cap S_d|} \leq \frac{C \cdot \log n}{\varepsilon^2}$. Either way, we have the desired bound. The proof for the fourth bullet is nearly the same, again using the conditioning on $B^c$ with $R_1(u)$ taking the place of $|N_u^+|$ and $S_r$ taking the place of $S_d$. 
\end{proof}

\begin{fact} \label{fct: overlapping-nbhds-Sd}
    Fix $u,v \in V$ and let $\mathcal{C}$ be as above. Suppose $\bar{d}_{uv} \leq 7/10$ and $|N_u^+ \cap S_d \cap C(u)| \geq (17/20) \cdot |N_u^+ \cap S_d|$. Then $|N_u^+ \cap N_v^+ \cap S_d  \cap C(u)|/|N_u^+ \cap S_d| \geq 3/20$. 
\end{fact}

\begin{proof}[Proof of Fact \ref{fct: overlapping-nbhds-Sd}]
    Since $\bar{d}_{uv} \leq 7/10$, we have that 
    \[
|N_u^+ \cap N_u^+ \cap S_d| \geq \frac{3}{10}\cdot |(N_u^+ \cup N_v^+) \cap S_d| \geq \frac{3}{10}\cdot |N_u^+ \cap S_d| .
    \]
    Further, since $|N_u^+ \cap S_d \cap C(u)| \geq (17/20) \cdot |N_u^+ \cap S_d|$, it follows that
    $|N_u^+ \cap S_d \cap \widebar{C(u)}| \leq (3/20) \cdot |N_u^+ \cap S_d|$.
    Then we see that \begin{align*}
        |N_u^+ \cap N_v^+ \cap S_d \cap C(u)| &= |N_u^+ \cap N_v^+ \cap S_d | -|N_u^+ \cap N_v^+ \cap S_d \cap \widebar{C(u)}|\\
        & \geq \frac{3}{10}\cdot |N_u^+ \cap S_d| -\frac{3}{20}|N_u^+ \cap S_d | = \frac{3}{20}|N_u^+ \cap S_d |.
        \end{align*}
\end{proof}

\subsubsection{Negative fractional cost of correlation metric for finite $p$} \label{sec: neg-frac-cost}

  We next bound the $\ell_p$-norm cost of the estimated adjusted correlation metric $\tilde{d}$ on negative edges. 

\begin{proof}[Proof of Lemma \ref{lem: whp-neg-frac-cost}]
Let $y$ be the vector of disagreements in a fixed clustering $\mathcal{C}$, let $C(u)$ denote the set of vertices in vertex $v$'s cluster, and let $\overline{C(u)} := V \setminus C(u)$.

    By Definition \ref{def: est-adj-corr}, the only negative edges $uv$ that contribute to the fractional cost are those with $\bar{d}_{uv} \leq \nicefrac{7}{10}$, i.e., with $1-\bar{d}_{uv} \geq \nicefrac{3}{10}$. Moreover, if $u \in R_1$, no negative edges incident to $u$ contribute to the fractional cost. Thus, to prove the lemma, it suffices to bound 
$\sum_{u \in R_2 \cap V_0}  |R_1(u)|^p ,$
 where recall $R_2 = V \setminus R_1$, and $R_1(u) = \{v \in N_u^- : \bar{d}_{uv} \leq \nicefrac{7}{10}\}$. While $R_1(u)$ is a random set, we note its randomness only depends on subsample $S_d$.

    We consider a few cases for $u \in R_2$. The sets below depend on the randomness of $S_d$. 
    \begin{align*}
        V^1 &:= V_0 \cap \{u \in V : |N_u^+ \cap S_d \cap \widebar{C(u)}| \geq \nicefrac{3}{20} \cdot |N_u^+ \cap S_d| \} \\
        V^2 &:= V_0 \cap \{u \in V : |N_u^+ \cap S_d \cap C(u)| \geq \nicefrac{17}{20} \cdot |N_u^+ \cap S_d| \} \\
        V^{2a} &:= V^2 \cap \{u \in V : |N_u^- \cap C(u)| \geq |N_u^+|\}, \qquad 
        V^{2b} := V^2 \cap \{u \in V : |C(u)| \leq 2 \cdot |N_u^+|\}
    \end{align*}
    Note that $V_0 = V^1 \cup V^{2a} \cup V^{2b}$, so 
    \[\sum_{u \in R_2 \cap V_0} |R_1(u)|^p  = \sum_{u \in R_2 \cap V^1} |R_1(u)|^p + \sum_{u \in R_2 \cap V^{2a}} |R_1(u)|^p  + \sum_{u \in R_2 \cap V^{2b}} |R_1(u)|^p .\]
Now we bound the three sums separately.

\begin{claim} \label{clm: r2-v1-whp}
Let $1 \leq p < \infty$. The following bounds hold:
\begin{itemize}
    \item $\mathbb{E} \left [\sum_{u \in R_2 \cap V^1} |R_1(u)|\right ] \leq O \left (\nicefrac{1}{\eps^2} \right )\cdot \opt_1 $,
    and 
    \item Conditioned on the event $B^c$, we have $\sum_{u \in R_2 \cap V^1} |R_1(u)|^p 
    \leq  O \left (\left ( \nicefrac{1}{\eps^2}\cdot \log n \right )^p \right )\cdot \opt_p^p$.
\end{itemize}
\end{claim}

\begin{proof}[Proof of Claim \ref{clm: r2-v1-whp}]
We have 
\begin{align} 
    \sum_{u \in R_2 \cap V^1} |R_1(u)|^p  = \hspace{-10pt}\sum_{u \in R_2 \cap V^1} |R_1(u)|^p \cdot \mathbf{1}_{\{|R_1(u) \cap S_r| = 0 \}} + \hspace{-10pt}\sum_{u \in R_2 \cap V^1} |R_1(u)|^p \cdot \mathbf{1}_{\{|R_1(u) \cap S_r| \geq 1 \}} \label{eq: splitVu-whp}. 
\end{align}

We bound the first term of the right-hand side as 
\begin{align*}
   \sum_{u \in V} |R_1(u)|^p &\cdot \mathbf{1}_{\{|R_1(u) \cap S_r| = 0 \}} \cdot \mathbf{1}_{\{u \in R_2\}} \cdot \mathbf{1}_{\{u \in V^1\}} \\
   &\leq \sum_{u \in V} |R_1(u)|^p \cdot \mathbf{1}_{\{|R_1(u) \cap S_r| = 0 \}} \cdot \mathbf{1}_{\{u \in V^1\}} \cdot |N_u^+ \cap S_d|   \\
   &  \leq \nicefrac{20}{3} \cdot \sum_{u \in V} |R_1(u)|^p \cdot \mathbf{1}_{\{|R_1(u) \cap S_r| = 0 \}} \cdot |N_u^+ \cap S_d \cap \widebar{C(u)}|^p\\
   &\leq \nicefrac{20}{3} \cdot \sum_{u \in V}   \left (y(u) \right )^p \cdot |R_1(u)|^p \cdot \mathbf{1}_{\{|R_1(u) \cap S_r| = 0 \}} 
\end{align*}
which for $p = 1$ is bounded in expectation by $\nicefrac{20}{3} \cdot \sum_{u \in V} y(u) \cdot \mathbb{E}[|R_1(u)| \cdot (1-q(\eps))^{|R_1(u)|}]$ (by Corollary \ref{cor: subsample-independence}), which in turn is bounded by $\nicefrac{40}{3\eps^2} \cdot \sum_{u \in V} y(u)$. For finite $p$, we obtain an upper bound of $\nicefrac{20}{3} \cdot (\nicefrac{2C\cdot \log n}{\eps^2})^p \cdot \sum_{u \in V} y(u)^p$, using the definition of $B^c$ (see item 5 in Definition \ref{def: good-event}).

Next, we bound the second term of the right-hand side of (\ref{eq: splitVu-whp}): 
\begin{align}
&\quad \sum_{u \in V} |R_1(u)|^p \cdot \mathbf{1}_{\{|R_1(u) \cap S_r| \geq 1\}} \cdot \mathbf{1}_{\{u \in R_2\}} \cdot \mathbf{1}_{\{u \in V^1\}} \notag \\
&\leq \left(\nicefrac{10}{3}\right)^p \cdot \sum_{u \in V} |N_u^+ \cap S_d|^p \cdot \mathbf{1}_{\{u \in V^1\}} \cdot \mathbf{1}_{\{|R_1(u) \cap S_r| \geq 1\}} \cdot \frac{|R_1(u)|^p }{|R_1(u) \cap S_r|^p} \notag \\
&\leq \left(\nicefrac{200}{9}\right)^p \cdot \sum_{u \in V} |N_u^+ \cap S_d \cap \overline{C(u)}|^p \cdot \mathbf{1}_{\{|R_1(u) \cap S_r| \geq 1\}} \cdot \frac{|R_1(u)|^p }{|R_1(u) \cap S_r|^p} \notag \\
&\leq \left(\nicefrac{200}{9}\right)^p \cdot \sum_{u \in V} y(u)^p \cdot \mathbf{1}_{\{|R_1(u) \cap S_r| \geq 1\}} \cdot \frac{|R_1(u)|^p }{|R_1(u) \cap S_r|^p} \notag 
\end{align}
Applying Proposition \ref{prop: reciprocal-corr} we obtain a bound of $\nicefrac{14}{\eps^2} \cdot \nicefrac{200}{9} \cdot \sum_{u \in V}y(u)$ in expectation for $p = 1$, and of $\left(\nicefrac{(200 \cdot C \cdot \log n)}{(9 \cdot\varepsilon^2)}\right)^{p}  \cdot \sum_{v \in V} y(v)^p$ conditioned on the event $B^c$ for finite $p \geq 1$.
\end{proof}

\begin{claim} \label{clm: r2-v2a-whp}
Let $1 \leq p < \infty$. The following bounds hold:
\begin{itemize}
    \item $\mathbb{E} \left [\sum_{u \in R_2 \cap V^{2a}} |R_1(u)| \right ]
    \leq O(\nicefrac{1}{\eps^2}) \cdot \opt_1$ 
    \item Conditioned on the event $B^c$, we have $\left [\sum_{u \in R_2 \cap V^{2a}} |R_1(u)|^p \right ] \leq O \left (\left ( \nicefrac{1}{\eps^2} \cdot \log n\right )^p \right )\cdot \opt_p^p.$
\end{itemize}
   
\end{claim}

\begin{proof}[Proof of Claim \ref{clm: r2-v2a-whp}]
    The proof is identical to that of Claim \ref{clm: r2-v1-whp} until the penultimate line, at which point we upper bound $|N_u^+ \cap S_d|$ by $|N_u^+|$, and then use that $u \in V^{2a}$ implies $y(u) \geq |N_u^+|$.
\end{proof}

    \begin{claim}
\label{clm: r2-v2b-whp}  
Let $1 \leq p < \infty$. The following bounds holds:
\begin{itemize}
    \item $\mathbb{E} \left [\sum_{u \in R_2 \cap V^{2b}} |R_1(u)|  \right ]
    \leq  O(\nicefrac{1}{\eps^2} )\cdot \opt_1$ 
    \item Conditioned on the event $B^c$, we have $\sum_{u \in R_2 \cap V^{2b}} |R_1(u)|^p  \leq   O \left ((\nicefrac{1}{\eps^2} \cdot \log n)^p\right ) \cdot \opt_p^p$.
\end{itemize} 
    \end{claim}
    
\begin{proof}[Proof of Claim \ref{clm: r2-v2b-whp}]
 Fix $u \in R_2 \cap V^{2b}$. For $v \in R_1(u)$, define 
\[N_{u,v} := N_u^+ \cap N_v^+ \cap C(u).\]
Since $\bar{d}_{uv} \leq 7/10$ and $|N_u^+ \cap S_d \cap C(u)| \geq (17/20) \cdot |N_u^+ \cap S_d|$, we have that 
$\frac{|N_{u,v} \cap S_d|}{|N_u^+ \cap S_d|} \geq \nicefrac{3}{20}$
by Fact \ref{fct: overlapping-nbhds-Sd}.
For $w \in N_u^+ \cap C(u)$, define 
\[\varphi(u,w) := |R_1(u) \cap N_w^+|.\]
Observe that $\varphi(u,w) \leq y(w) + y(u)$.

We have, by way of double counting, 
\[\sum_{w \in N_u^+ \cap S_d \cap C(u)} \varphi(u,w) = \hspace{-10pt}\sum_{w \in N_u^+ \cap S_d \cap C(u)} |R_1(u) \cap N_w^+| = \sum_{v \in R_1(u)} |N_{u,v} \cap S_d| \geq \nicefrac{3}{20} \cdot |N_u^+ \cap S_d| \cdot |R_1(u)|,\]

which implies 

\begin{align*}
    &\quad  \sum_{u \in V^{2b}} \frac{|N_u^+|^p}{|N_u^+ \cap S_d|^p} \cdot \frac{1}{|C(u)|} \cdot \sum_{w \in N_u^+ \cap C(u) \cap S_d} \varphi(u,w)^p \\
    &\geq \sum_{u \in V^{2b}} \frac{|N_u^+|^p}{|N_u^+ \cap S_d|^p} \cdot \frac{1}{|C(u)|} \cdot \frac{1}{|N_u^+ \cap C(u) \cap S_d|^{p-1}} \cdot \left(\sum_{w \in N_u^+ \cap C(u) \cap S_d} \varphi(u,w)\right)^p \\
    &\geq \sum_{u \in V^{2b}} \frac{|N_u^+|^p}{|N_u^+ \cap S_d|^p} \cdot \frac{1}{2|N_u^+|} \cdot \frac{1}{|N_u^+|^{p-1}} \cdot \left(\sum_{w \in N_u^+ \cap C(u) \cap S_d} \varphi(u,w)\right)^p \\
    &\geq \sum_{u \in V^{2b}} \frac{|N_u^+|^p}{|N_u^+ \cap S_d|^p} \cdot \frac{1}{2|N_u^+|} \cdot \frac{1}{|N_u^+|^{p-1}} \cdot (3/20)^p \cdot |N_u^+ \cap S_d|^p \cdot |R_1(u)|^p \\
    &= \sum_{u \in V^{2b}} \nicefrac{1}{2} \cdot (\nicefrac{3}{20})^p \cdot |R_1(u)|^p.
\end{align*}
In the second line we have used Jensen's inequality. So to bound $\sum_{u \in V^{2b}} |R_1(u)|^p$, it suffices to bound 
\begin{align*}
    &\quad \sum_{u \in V^{2b}} \frac{|N_u^+|^p}{|N_u^+ \cap S_d|^p} \cdot \frac{1}{|C(u)|} \cdot \sum_{w \in N_u^+ \cap C(u) \cap S_d} \varphi(u,w)^p\\
    &\leq  \sum_{u \in V^{2b}} \frac{|N_u^+|^p}{|N_u^+ \cap S_d|^p} \cdot \frac{1}{|C(u)|} \cdot \sum_{w \in N_u^+ \cap C(u) \cap S_d} (y(u) + y(w))^p \\
    &\leq 2^{p-1} \cdot \sum_{u \in V} \frac{|N_u^+|^p}{|N_u^+ \cap S_d|^p} \cdot \mathbf{1}_{\{u \in V_0\}} \cdot \frac{1}{|C(u)|} \cdot \sum_{w \in N_u^+ \cap C(u) \cap S_d} y(u)^p \\
    &\qquad + 2^{p-1} \cdot \sum_{u \in V} \frac{|N_u^+|^p}{|N_u^+ \cap S_d|^p} \cdot \mathbf{1}_{\{u \in V_0\}} \cdot \frac{1}{|C(u)|} \cdot \sum_{w \in N_u^+ \cap C(u) \cap S_d} y(w)^p
\end{align*}
by Jensen's inequality again. By Proposition \ref{prop: reciprocal-corr}, the first term  is easily seen to be bounded by $\nicefrac{14}{\eps^2} \cdot \opt_1$ in expectation for $p=1$, and by $2^{p-1} \cdot (\nicefrac{C \cdot \log n}{\eps^2})^p \cdot \opt_p^p$ conditioned on the event $B^c$ for finite $p \geq 1$. We bound the second sum: 
\begin{align*}
& \quad 2^{p-1} \cdot \sum_{u \in V} \frac{|N_u^+|^p}{|N_u^+ \cap S_d|^p} \cdot \mathbf{1}_{\{u \in V_0\}} \cdot  \frac{1}{|C(u)|} \cdot \sum_{w \in N_u^+ \cap C(u) \cap S_d} y(w)^p \\
&\leq 2^{p-1} \cdot \sum_{w \in V} y(w)^p \cdot \sum_{u \in N_w^+ \cap C(w)} \frac{|N_u^+|^p}{|N_u^+ \cap S_d|^p} \cdot \mathbf{1}_{\{u \in V_0\}} \cdot \frac{1}{|C(u)|}\\
&\leq 2^{p-1} \cdot \sum_{w \in V} y(w)^p \cdot \sum_{u \in N_w^+ \cap C(w)} \frac{|N_u^+|^p}{|N_u^+ \cap S_d|^p} \cdot \mathbf{1}_{\{u \in V_0\}}  \cdot \frac{1}{|C(w)|}.
\end{align*}
Again applying Proposition \ref{prop: reciprocal-corr}, we obtain the same upper bounds as above for the first term.
\end{proof}
Combining Claims \ref{clm: r2-v1-whp}, \ref{clm: r2-v2a-whp}, and \ref{clm: r2-v2b-whp} completes the proof of Lemma \ref{lem: whp-neg-frac-cost}.
\end{proof}

\section{Cost of Algorithm \ref{alg: main-alg} for Finite $p$} \label{sec: lp-norm-analysis}

We bound the cost of Algorithm \ref{alg: main-alg} for $p \in [1,\infty)$. We still use $\opt_p$ to refer to the value of an optimal solution for the $\ell_p$-norm objective, and additionally we often fix such an optimal clustering $\mathcal{C}_{\textsf{OPT}}$.

\begin{figure}[h]
\begin{minipage}{0.65\textwidth}
\hfill
\includegraphics[height=3.5cm]{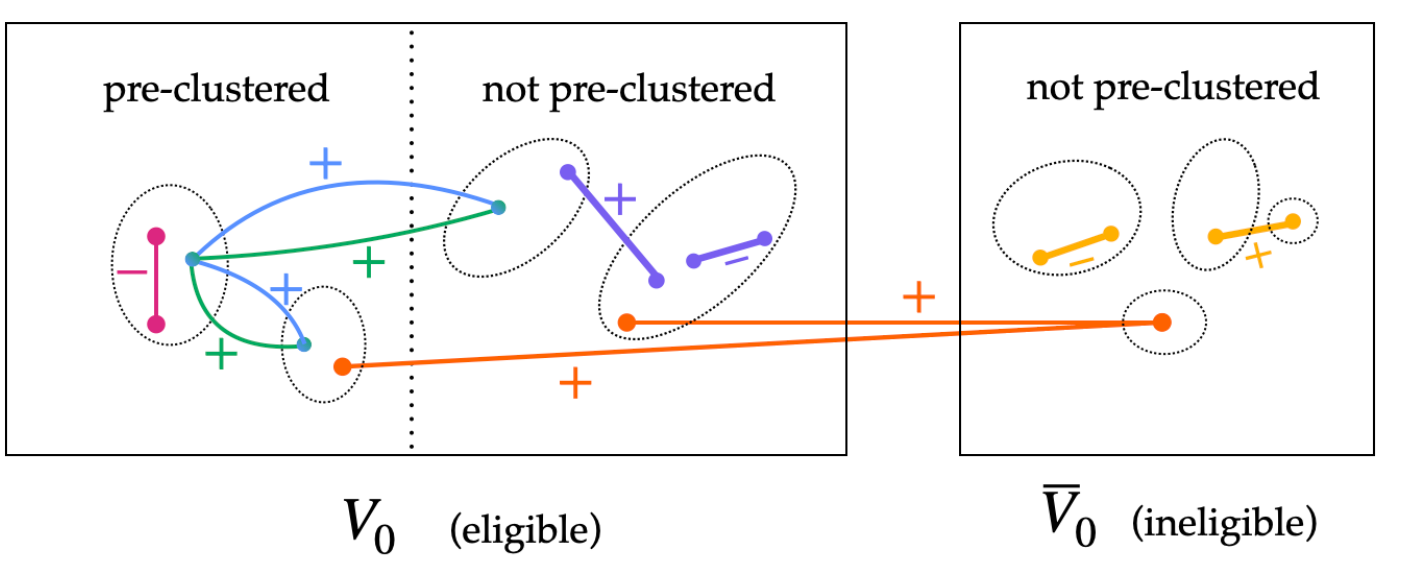}
\end{minipage}
\hfill 
\begin{minipage}{0.28\textwidth}
\footnotesize{
\textcolor{myblue}{Lemma \ref{lem: type1b-lp}}  \\ 
\textcolor{mygreen}{Lemma \ref{lem: type1b-lp-backwards}}  \\ 
\textcolor{mypink}{Lemma \ref{lem: negative-lp}}  \\
\textcolor{myyellow}{Lemma \ref{lem: pivot-cost-lp-v0}} \\
\textcolor{mypurple}{Lemma \ref{lem: pivot-cost-lp-v''}}\\
\textcolor{myorange}{Lemmas \ref{lem: type-t-lp} and \ref{lem: type-t-lp-backward}} }
\end{minipage}
\captionsetup{width=.9\textwidth}
\caption{Overview of lemmas for the analysis of Algorithm \ref{alg: main-alg} for finite $p$. Solid edges are disagreements, and dashed ovals are clusters. Vertices are partitioned into three sets based on whether or not they are eligible and pre-clustered, eligible and not pre-clustered, or ineligible.  Charging the cost of a disagreement then depends on which set its endpoints belong to, its sign, and potentially which endpoint was higher with respect to the partial ordering $\succ$. 
Edges are partitioned by color, with edge types of the same color bounded by the correspondingly colored lemma. 
Note Lemmas \ref{lem: type1b-lp} and \ref{lem: type1b-lp-backwards} correspond to the same $uv$ pair, but which lemma is relevant depends on (from the perspective of $u$'s disagreements) whether $u \succ v$ or $v \succ u$.}
\label{fig:summary-ellp}
\end{figure}

At a high-level, we charge disagreements made by Algorithm \ref{alg: main-alg} to the disagreements in $\mathcal{C}_{\textsf{OPT}}$. Recall Algorithm \ref{alg: main-alg} has several subroutines: 
\begin{itemize}
    \item a Pre-clustering phase that clusters nodes $v$ that have (i) some of their positive neighborhood sampled into $S_d$, i.e. $v \in V_0$, and (ii) are close (with respect to $\tilde{d}$) to a cluster center $S_p$;
    \item a subroutine that runs the standard Pivot algorithm on nodes $v \in \widebar{V_0}$, which are the nodes that did not have any positive neighbor sampled into $S_d$;
    \item and a subroutine that runs a modified version of the Pivot algorithm on nodes $v \in V_0'$, which are the nodes that have a positive neighbor sampled into $S_d$, but were far away from all cluster centers. 
\end{itemize} 
We note the last two subroutines are part of the Pivot phase of Algorithm \ref{alg: main-alg}. Further, the clusters output by each subroutine are totally disjoint.
The disagreements incurred by Algorithm \ref{alg: main-alg} can be partitioned into the disagreements made \emph{within} each subroutine (e.g., $u$ and $v$ are both clustered in the Pre-clustering phase, but $uv$ forms a disagreement in the solution output by Algorithm \ref{alg: main-alg}), and the disagreements \emph{between} each subroutine (e.g., $u$ is clustered in the Pre-clustering phase and $v$ is clustered in the Pivot phase, but $uv \in E^+$). Therefore, we partition our analysis into the cost of the disagreements incurred during the Pre-clustering phase (Section \ref{sec: pre-clustering-ellp}), the cost of the disagreements incurred during the  Pivot phase (see Section \ref{sec: pivot-phase-lp}), and those cost incurred between these two phases (Section \ref{sec: between-phases}). Note the cost of the disagreements incurred during the Pivot phase includes the cost from running the standard Pivot algorithm on nodes $G[\widebar{V_0}]$, the cost from running the modified Pivot algorithm on nodes $G[V_0']$, and the cost of disagreements between the two subroutines of the Pivot phase. See Figure \ref{fig:summary-ellp} for references to the lemmas for each type of disagreement.

Sometimes we are able to directly charge disagreements Algorithm \ref{alg: main-alg} makes to $\mathcal{C}_{\textsf{OPT}}$. More often, we use the estimated adjusted correlation metric, $\tilde{d},$ as an intermediary---specifically, we charge disagreements made by  Algorithm \ref{alg: main-alg} to  $\tilde{d}$, then charge the cost of $\tilde{d}$ to $\opt_p$. The latter charging arguments can be found in Section \ref{sec: guar-est}.

As in Section \ref{sec: guar-est}, we condition on the good event $B^c$ occurring with high probability (see the end of Section \ref{sec: terminology}).

\subsection{Cost of Pre-clustering phase} \label{sec: pre-clustering-ellp}

Throughout, we use the choices of $\delta,c,r$ in Algorithm \ref{alg: main-alg}, and let $t := r/(2\delta)$ be a threshold parameter, which will be used in our analysis.

Recall that for nodes $u$ assigned to clusters during the Pre-clustering phase, there is some node in $S_p$ that has close $\tilde{d}$ distance to $u$. The highest ordered, with respect to the ordering of $S_p$, is said to pre-cluster $u$ and is denoted by $s^*(u)$. We may refer to $s^*(u)$ as $u$'s \emph{center}. 

Recall we say \emph{$v$ is clustered before $u$} (or \emph{$u$ is clustered after $v$}) if either both $u$ and $v$ are pre-clustered, but $s^*(v)$ is before $ s^*(u)$ (with respect to the ordering of $S_p$), or if $v$ is pre-clustered but $u$ is not. For shorthand, we write $v \succ u$ when $v$ is clustered before $u$. 

Fix a node $u$ that is assigned a cluster during the Pre-clustering phase of Algorithm \ref{alg: main-alg}. Consider all nodes $v \in V_0$, so that $uv$ is a disagreement in the output of Algorithm \ref{alg: main-alg}. 

\subsubsection{Cost of positive edges}

We begin by bounding the cost of positive edges where at least one endpoint is pre-clustered, and both endpoints are eligible.

Fix a vertex $u \in V_0$. We partition the cost of positive disagreements incurred within the Pre-clustering phase based on whether $u \prec v$ or  $v \prec u$. Lemma \ref{lem: type1b-lp} handles the cost of disagreements $uv$ incident to $u$ when $v \succ u$, while Lemma \ref{lem: type1b-lp-backwards} handles the cost of disagreements $uv$ incident to $u$ when $u \succ v$ and $v \in V_0$. In the proofs of both lemmas, we will see that it is easy to charge a disagreement $uv$ to $\tilde{d}$ when $\tilde{d}_{uv}$ is sufficiently large, as we can then charge the $\ell_p$-norm cost of $\tilde{d}$ to $\opt_p$ using the lemmas in Section \ref{sec: guar-est}. On the other hand, the difficult settings for both lemmas are for edges $uv$ where $\tilde{d}_{uv}$ is small and both $u,v \in V_0$, so this distance is actually a reliable indicator that $u$ and $v$ do have many positive neighbors in common. The key is that even though $\tilde{d}_{uv}$ is small, the fact that $u$ and $v$ are not clustered together indicates there must be some other vertices we can charge to that do have large distance from $u$. 

Some of the future claims will use that for the choices of $\delta,c,r$ as in the algorithm, 
\begin{equation}\label{eq: choose-consts}
     \max \Big \{\frac{1}{cr/\delta-r}, \frac{1}{1-(\delta \cdot r + \delta^2 \cdot c \cdot r + \delta \cdot r/2)}\Big \} \leq 8 \quad \text{and} \quad  \max \Big \{ \frac{1}{r/2+c \cdot \delta\cdot r}, \frac{1}{c \cdot r}\Big \}\leq 5.\end{equation}

\begin{lemma} \label{lem: type1b-lp}
    Condition on the good event $B^c$ and fix $1 \leq p < \infty$. The $\ell_p$-cost for $u \in V_0$
    of the edges $uv \in E^+$ 
    in disagreement with respect to $\mathcal{C}_{\textsf{ALG}}$, where  $v \succ u$, is bounded by $O \left( \frac{1}{\eps^8} \cdot \log^4 n  \right) \cdot \opt_p.$
\end{lemma}

\begin{proof}[Proof of Lemma \ref{lem: type1b-lp}]
Note by definition of $\succ$ that each $v$ in the statement of the lemma is necessarily pre-clustered, thus also $v \in V_0$. We partition the set $\{v \in N_u^+ : v \succ u\}$
depending on whether $\tilde{d}_{uv}>t$ or 
$\tilde{d}_{uv}\leq t$. Define $E_1(u)$ to be the random set of $u$'s close, positive neighbors that are clustered before $u$:

\[E_1(u) := \{v \in N_u^+ : v \succ u\} \cap \ball_{\tilde{d}}(u,t). \]

Define $E_2(u)$ to be the remaining positive neighbors of $u$ that are clustered before $u$:

\[E_2(u):=  \{v \in N_u^+ : v \succ u\}\setminus E_1(u) .\]

We partition the sum we wish to bound using Jensen's inequality to see
\begin{align}
    \sum_{u \in V_0} \big | \{v \in N_u^+ : v \succ u\} \big |^p   
    &\leq 2^{p-1} \cdot \sum_{u \in V_0} | E_1(u)|^p + 2^{p-1}\cdot \sum_{u \in V_0}  |E_2(u)|^p. \label{eqn: partition-lem-v-first}
\end{align}

As we alluded to before the beginning of the proof, it is straightforward to bound the cost of disagreeing edges $uv$ when $\tilde{d}_{uv}$ is large. In particular, we can bound the latter sum: 
\begin{align}
    \sum_{u \in V_0}|E_2(u)|^p  = \sum_{u \in V_0} |\{v \in N_u^+ \cap V_0 : \tilde{d}_{uv} \geq t\}|^p &\leq \frac{1}{t^p} \cdot \sum_{u \in V_0}  \Bigg ( \sum_{v \in N_u^+ \cap V_0}  \tilde{d}_{uv}  \Bigg )^p \notag\\
    &\leq     \left ( \nicefrac{(7215 \cdot C^3 \cdot \log^3 n )}{\eps^6}\right ) ^p   \cdot  \opt_p^p. 
    \label{eqn:bound-e1}
\end{align}
where the third inequality is from  Lemma \ref{lem: whp-pos-frac-cost} and subbing in the values of $\delta$ and $r$. 

\medskip

Bounding $\sum_{u \in V_0}  |E_1(u)|^p$ in line (\ref{eqn: partition-lem-v-first})  is the more involved piece. We begin by partitioning $u \in V_0$ based on whether $u$ has a close neighbor sampled by the center sample $S_p$;
overall, we need to bound $E_{1a}$ and $E_{1b}$ where 
\begin{align*}
 \sum_{u \in V_0}  |E_1(u)|^p & =  \underbrace{\sum_{u \in V_0: \ball_{\tilde{d}}^{S_p}(u,t) = \emptyset}  |E_1(u)|^p}_{E_{1a}} +    \underbrace{\sum_{u \in V_0: \ball_{\tilde{d}}^{S_p}(u,t) \neq \emptyset}  |E_1(u)|^p}_{E_{1b}}
\end{align*}

\paragraph*{Bounding $E_{1a}$.} 
Intuitively, the term $E_{1a}$ will be easier to bound than $E_{1b}$, because, conditioned on $B^c$, the fact that $\ball_{\tilde{d}}^{S_p}(u, t)=\emptyset$ implies that $|\ball_{\tilde{d}}(u, t)|$ is small. So even though there are some nodes $v \in N_u^+$ that are close to $u$ but assigned a different cluster than $u$, there cannot be that many of them. We use this insight together with the following claim, Claim \ref{clm: sufficient-charge}, which proves that there is sufficient fractional cost incident to $u \in V_0$. In turn, the small number of disagreements incident to $u$ can be charged to this fractional cost, via Lemma \ref{lem: whp-pos-frac-cost}. 

Recall that 
$\tilde{D}_0(u):= \sum_{v \in N_u^+ \cap V_0} \tilde{d}_{uv} + \sum_{v \in N_u^- \cap V_0} (1-\tilde{d}_{uv}). $

\begin{restatable}{claim}{etwotwoprop}\label{clm: sufficient-charge}
     If $u \in V_0$ and $E_1(u) \neq \emptyset$, then $\tilde{D}_0(u) \geq  \nicefrac{1}{5}$. 
\end{restatable}

\begin{proof}[Proof of Claim \ref{clm: sufficient-charge}]
Let $v$ be any vertex in $E_1(u)$. Then $s^*(v)$ exists. We can upper and lower bound $\tilde{d}_{u \hspace{1pt}s^*(v)}$ by a constant. Namely, by the approximate triangle inequality (Lemma \ref{lem: tri-inequality}),
$\tilde{d}_{u \hspace{1pt}s^*(v)} \leq \delta(\tilde{d}_{uv} + \tilde{d}_{v \hspace{1pt}s^*(v)}) \leq \delta(t + c \cdot r)$. On the other hand,  since $u$ is eligible but $v \succ u$, we know that
$\tilde{d}_{u \hspace{1pt}s^*(v)} > c \cdot r$.

Since $\tilde{d}_{u \hspace{1pt}s^*(v)} < 1$, we have that $s^*(v) \in V_0$ (Fact \ref{fct: frac-isolation}). This means that $D_0(u)$ is lower bounded by $1-d_{u \hspace{1pt}s^*(v)} \geq 1- \delta(t + c \cdot r) \geq c \cdot r  \geq \nicefrac{1}{5}$ in the case that $u \hspace{1pt}s^*(v) \in E^-$ and by $d_{u \hspace{1pt} s^*(v)} \geq c \cdot r \geq \nicefrac{1}{5}$ in the case that $u \hspace{1pt}s^*(v) \in E^+$, using the lower bounds from line (\ref{eq: choose-consts}).
\end{proof}

We now bound $E_{1a}$ in the following claim.
\begin{claim}
\label{claim: E1a-lp}
Condition on the good event $B^c$. For $t = \frac{r}{2 \delta}$, it is the case that 
 \[E_{1a} := \sum_{u \in V_0: \ball_{\tilde{d}}^{S_p}(u,t) = \emptyset}  |E_1(u)|^p \leq  O \left( \left (\nicefrac{1}{\eps^8} \cdot \log^4 n  \right)^p \right ) \cdot \opt_p^p.\]
\end{claim}

\begin{proof}[Proof of Claim \ref{claim: E1a-lp}]
It suffices here to upper bound $E_1(u)$ by $|\ball_{\tilde{d}}(u,t)|$. Applying Claim \ref{clm: sufficient-charge}, we obtain
\begin{align*}
    E_{1a}  \leq 
    \sum_{u \in V_0 : \ball_{\tilde{d}}^{S_p}(u, t) = \emptyset} |\ball_{\tilde{d}}(u, t)|^p &\leq 5^p \cdot \sum_{u \in V_0 : \ball_{\tilde{d}}^{S_p}(u, t) = \emptyset} |\ball_{\tilde{d}}(u, t)|^p \cdot (\tilde{D}_0(u) )^p
    \\
   &\leq \Big ( 5C' \cdot \log_{1/(1-q(\varepsilon))} n  \Big )^p
    \sum_{u \in V_0} (\tilde{D}_0(u) )^p \\
    &
    \leq  
    \left ( \nicefrac{(4440 \cdot C' \cdot C^3 \cdot \log^4 n )}{ \eps^{8}}
    \right )^p
    \cdot \opt_p^p.
\end{align*}
In the penultimate line, we use that conditioning on $B^c$, if $\ball_{\tilde{d}}^{S_p}(u, t)= \emptyset$, it must be that $|\ball_{\tilde{d}}(u, t)| \leq C' \cdot \log_{1/(1-q(\varepsilon))} n \leq \nicefrac{8 C'}{\eps^2} \cdot \log n.$
In the last line we use Corollary \ref{cor: bounded-exp-frac-cost}.

\end{proof}

\paragraph*{Bounding $E_{1b}$.}  Recall 
\[E_{1b} := \sum_{u \in V_0 : \ball_{\tilde{d}}^{S_p}(u,t) \neq \emptyset} |E_1(u)|^p =  \sum_{u \in V_0 : \ball_{\tilde{d}}^{S_p}(u,t) \neq \emptyset} |\{v \in N_u^+ : v \succ u, \tilde{d}_{uv} \leq t\}|^p.\] 
Intuitively, because $u$ is both pre-clustered and has a close neighbor sampled in $S_p$, we are now closer to the offline setting. In particular, since $v \succ u$, we have $|\ball^{S_b}_{\tilde{d}}(s^*(u), r)| \leq |\ball^{S_b}_{\tilde{d}}(s^*(v), r)|$. The idea is that $u$ lies in an annulus around $\ball^{S_b}_{\tilde{d}}(s^*(v), r)$, and so the fractional cost of $u$ can be lower bounded by (a constant factor times) the $|\ball^{S_b}_{\tilde{d}}(s^*(v), r)|$. The subtlety is that the inequality above lower bounding $|\ball^{S_b}_{\tilde{d}}(s^*(v), r)|$ in turn  only holds when the balls are restricted to $S_b$, unlike in the offline case, where $S_b = V$. So it is not a priori clear that there will be enough fractional cost to which to charge $|E_1(u)|^p$.  See Figure \ref{fig: type1b-lp} for an illustration.

\begin{figure}
    \centering
  \captionsetup{width=.9\textwidth}
    \includegraphics[width=5cm]{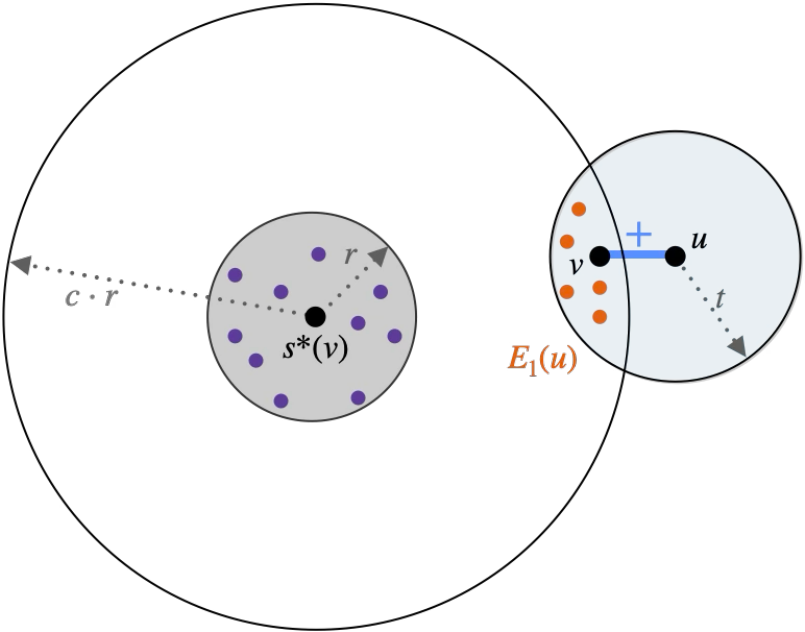}    \caption{Bounding $E_{1b}$ in the proof of Lemma \ref{lem: type1b-lp}, where $v \in N_u^+$ with $v \succ u$ and $\tilde{d}_{uv} \leq t$. We charge the disagreements between $v \in E_1(u)$ and $u$ to
    the purple nodes  in $\ball^{S_b}_{\tilde{d}}(s^*(v),r)$.
    }
    \label{fig: type1b-lp}
\end{figure}

For each $u \in V_0$ with $\ball_{\tilde{d}}^{S_p}(u,t) \neq \emptyset$, choose a fixed but arbitrary $z(u) \in \ball_{\tilde{d}}^{S_p}(u,t)$. Note that because $z(u) \in S_p$ and $\tilde{d}_{u \hspace{1pt} z(u)} \leq t \leq c \cdot r$, we know that $z(u)$ is a \emph{candidate} for clustering $u$, so in particular $s^*(u)$ exists. 
Note $z(u)$ is a random variable depending on $S_p, S_d,$ and $ S_r$.

Recall that $E_1(u) := \{v \in N_u^+ : v \succ u, \tilde{d}_{uv} \leq t\}$. Define

\[B(u) := \bigcup_{v \in E_1(u)} \ball^{S_b}_{\tilde{d}}(s^*(v), r),\] 
that is, $B(u)$ is the union of balls in $S_b$, cut out around the vertices that cluster the vertices in $E_1(u)$. We will show that we can charge $|E_1(u)|^p$ to $|B(u)|^p$. Further, the set $B(u)$ is constructed so that every node $b \in B(u)$ lies in an annulus around $u$, so we can in turn charge $|B(u)|^p$ to the $\ell_p$-cost of $\tilde{d}$.

The interesting case is when $|\ball_{\tilde{d}}(z(u),r)|$ is large. Here, we use Claims \ref{clm: upper-bd-Cu} and \ref{clm: lower-bd-Au} to bound $|E_1(u)|$ in terms of $|B(u)|$. Then we use Claim \ref{clm: upper-bd-Au} to relate $|B(u)|$ to the $\ell_p$-cost of $\tilde{d}$.

\begin{restatable}{claim}{upperbdCu} \label{clm: upper-bd-Cu}
    If $\ball_{\tilde{d}}^{S_p}(u, t) \neq \emptyset$, then $|E_1(u)| \leq |\ball_{\tilde{d}}(z(u), r)|$.
\end{restatable}

\begin{proof}[Proof of Claim \ref{clm: upper-bd-Cu}]
The claim follows from an application of the approximate triangle inequality (Lemma \ref{lem: tri-inequality}). If $\ball_{\tilde{d}}^{S_p}(u, t) \neq \emptyset$, $z(u)$ exists and  $\tilde{d}_{u \hspace{1pt}z(u)} \leq t$. Further, by definition, if $v \in E_1(u)$, then $\tilde{d}_{uv} \leq t$. So  for any $v \in E_1(u)$, we have $\tilde{d}_{v \hspace{1pt} z(u)} \leq 2\delta t = r$, so $E_1(u) \subseteq \ball_{\tilde{d}}(z(u), r)$. 
\end{proof}

\begin{restatable}{claim}{lowerbdAu} \label{clm: lower-bd-Au}
    If $u \in V_0$ and $\ball_{\tilde{d}}^{S_p}(u, t) \neq \emptyset$, then $|\ball^{S_b}_{\tilde{d}}(z(u),r)|  \leq |B(u)|$.
\end{restatable}

\begin{proof}[Proof of Claim \ref{clm: lower-bd-Au}]
    If $\ball_{\tilde{d}}^{S_p}(u, t)\neq \emptyset$, then $z(u)$ exists and is a center, and in particular is a candidate for pre-clustering $u$ (since $\tilde{d}_{u, z(u)} \leq t < c \cdot r$ and $u \in V_0$). Thus $s^*(u)$ exists and $|\ball^{S_b}_{\tilde{d}}(z(u),r)| \leq |\ball^{S_b}_{\tilde{d}}(s^*(u),r)|.$ Further, for any $v \in E_1(u)$, $|\ball^{S_b}_{\tilde{d}}(s^*(u),r) |\leq |\ball^{S_b}_{\tilde{d}}(s^*(v),r)|$, since $v$ by definition is clustered before $u$. The claim then follows from the definition of $B(u)$.
\end{proof}

\begin{restatable}{claim}{upperbdAu} \label{clm: upper-bd-Au}
    If $u \in V_0$, then $|B(u)|  \leq 8 \cdot \tilde{D}_0(u).$ 
\end{restatable}

\begin{proof}[Proof of Claim \ref{clm: upper-bd-Au}]
    If $E_1(u) = \emptyset
    $, then $B(u) = \emptyset$, so the claim holds. Thus we may assume that $E_1(u) \neq \emptyset$. It then suffices to show that $\tilde{d}_{ub} \geq \nicefrac{1}{8}$ and $1-\tilde{d}_{ub} \geq \nicefrac{1}{8}$ for every $b \in B(u)$. Then we will have (by Fact \ref{fct: frac-isolation}) that $b \in V_0$ (because $\tilde{d}_{ub} < 1$) for every $b \in B(u)$, and thus the claim follows. 
    
     To see that $\tilde{d}_{ub} \geq \nicefrac{1}{8}$ for any $b \in B(u)$, let $v \in E_1(u)$ be such that $b \in \ball_{\tilde{d}}(s^*(v), r)$ (such $v$ exists by the definition of $B(u)$). Since $v \in E_1(u)$, $u$ is clustered after $v$, so, using also that $u \in V_0$, we have that $\tilde{d}_{u \hspace{1pt}s^*(v)} > c \cdot r$. Also, by choice of $v$, $\tilde{d}_{b \hspace{1pt}s^*(v)} \leq r$. So by the approximate triangle inequality (Lemma \ref{lem: tri-inequality}), 
            $\tilde{d}_{ub} \geq cr/\delta -r$, which is lower bounded by $\nicefrac{1}{8}$ by line (\ref{eq: choose-consts}).

      To see that $1-\tilde{d}_{ub} \geq \nicefrac{1}{8}$ for any $b \in B(u)$, observe that $\tilde{d}_{b \hspace{1pt} s^*(v)} \leq r$ (by choice of $v$), $\tilde{d}_{v \hspace{1pt}s^*(v)} \leq c \cdot r$ (by definition of the algorithm and of $s^*(v)$), and $\tilde{d}_{vu} \leq t$ (since $v \in E_1(u)$). So by the approximate triangle inequality (Lemma \ref{lem: tri-inequality}), we have
    $\tilde{d}_{ub} \leq \delta \cdot [\tilde{d}_{b\hspace{1pt} s^*(v)} + \delta(\tilde{d}_{v \hspace{1pt}s^*(v) } + \tilde{d}_{vu})] \leq \delta \cdot r + \delta^2 \cdot c \cdot r + \delta^2 \cdot t,$
    so $1-\tilde{d}_{ub} \geq 1-(\delta \cdot r + \delta^2 \cdot c \cdot r + \delta^2 \cdot t)$, which is lower bounded by $\nicefrac{1}{8}$ by line (\ref{eq: choose-consts}).
\end{proof}

\begin{claim}
    \label{claim: E1b-lp}
Condition on the good event $B^c$. For $t = \frac{r}{2 \delta}$, it is the case that  
\[E_{1b} := \sum_{u \in V_0: \ball_{\tilde{d}}^{S_p}(u,t) \neq  \emptyset}  |E_1(u)|^p \leq  O \left( \left (\nicefrac{1}{\eps^8} \cdot \log^4 n  \right)^p \right ) \cdot \opt^p_p.\]
\end{claim}

\begin{proof}[Proof of Claim \ref{claim: E1b-lp}]
We use Claim \ref{clm: upper-bd-Cu}, and then partition the sum 
\begin{align}
    E_{1b}  &\leq \sum_{u \in V_0: \ball_{\tilde{d}}^{S_p}(u,t) \neq  \emptyset}
    \big |\ball_{\tilde{d}}(z(u),r) \big |^p \notag \\
    &\leq  \sum_{\overset{u \in V_0 :  \ball_{\tilde{d}}^{S_p}(u, t) \neq \emptyset,} {|\ball_{\tilde{d}}(z(u),r)| \geq 2 C \log n / \eps^2}}  \big |\ball_{\tilde{d}}(z(u),r)   \big |^p + 
    \sum_{\overset{u \in V_0 :  \ball_{\tilde{d}}^{S_p}(u, t) \neq \emptyset,}{|\ball_{\tilde{d}}(z(u),r)| <2 C \log n / \eps^2}}  \big |\ball_{\tilde{d}}(z(u),r) \big |^p.\label{eq: E01-finish}
\end{align}
 First, we upper bound the first sum in line (\ref{eq: E01-finish}): 
\begin{align*}
   \sum_{\overset{u \in V_0 :  \ball_{\tilde{d}}^{S_p}(u, t)  \neq \emptyset}{|\ball_{\tilde{d}}(z(u),r)| \geq 2 C \log n /\eps^2}} \hspace{-5mm} \big |\ball_{\tilde{d}}(z(u),r) \big |^p
    &\leq  \hspace{-15pt}\sum_{\overset{u \in V_0 :  }{|\ball_{\tilde{d}}(z(u),r)| \geq 2 C \log n /\eps^2}} \hspace{-15pt} \Bigg ( |\ball_{\tilde{d}}(z(u),r) |\cdot  8   \cdot \frac{\tilde{D}_0(u)}{|\ball^{S_b}_{\tilde{d}}(z(u),r) |}  \Bigg )^p\\
    &\leq  \left (\nicefrac{32}{ \eps^2} \right )^p\cdot \sum_{u \in V_0}    \big (\tilde{D}_0(u) \big )^p \leq \left ( \nicefrac{(3552 \cdot C^3 \cdot \log^3 n )}{\eps^8}\right ) ^p \cdot \opt_p^p,
\end{align*}
where in the first inequality, we have applied Claims \ref{clm: lower-bd-Au} and \ref{clm: upper-bd-Au}, and in the second inequality, we use the conditioning on $B^c$, which implies that $|\ball_{\tilde{d}}(z(u),r)| / |\ball^{S_b}_{\tilde{d}}(z(u),r)|  \leq 4 /\eps^2$, since $|\ball_{\tilde{d}}(z(u),r)| \geq 2 C \log n / \eps^2$.
In the last inequality we have used Corollary \ref{cor: bounded-exp-frac-cost}.

Finally, we bound the second sum in line (\ref{eq: E01-finish}): 
\begin{align*}
 \sum_{\overset{u \in V_0 : \ball_{\tilde{d}}^{S_p}(u,t)  \neq \emptyset,}{|\ball_{\tilde{d}}(z(u),r)| < 2 C \log n / \eps^2}}   \big |\ball_{\tilde{d}}(z(u),r) \big |^p
      &= 
    \sum_{\overset{u \in V_0 : }{|\ball_{\tilde{d}}(z(u),r)| < 2 C \log n / \eps^2}}  \big (5 \cdot  |\ball_{\tilde{d}}(z(u),r)| \big )^p \cdot \big (\tilde{D}_0(u) \big )^p\\
     &\leq      \left ( \nicefrac{(1110 \cdot C^4 \cdot \log^4 n )}{\eps^8}\right ) ^p \cdot \opt_p^p.
\end{align*}

The first line follows from Claim \ref{clm: sufficient-charge}.
In the second line,
we use the bound $|\ball_{\tilde{d}}(z(u),r)| < 2 C \log n / \eps^2$ and Corollary \ref{cor: bounded-exp-frac-cost}. 
\end{proof}
 
Combining the bounds on the sums and using that $C$ is sufficiently large (which is required by the good event, anyway), we conclude 
$
\sum_{u \in V_0} \big | \{v \in N_u^+ : v \succ u\} \big | ^p  \leq \left ( \nicefrac{(2900 \cdot C' \cdot C^4 \cdot \log^4 n )}{\eps^{8}}\right ) ^p \cdot \opt^p_p.
$
\end{proof}

\bigskip

The proof of Lemma \ref{lem: type1b-lp-backwards} is similar in spirit to that of Lemma \ref{lem: type1b-lp}, but must nonetheless be handled separately (except in the case of $p=1$, where we can sum over disagreements edge-wise rather than node-wise).

\begin{lemma} \label{lem: type1b-lp-backwards}
Condition on the good event $B^c$ and fix $1 \leq p < \infty$. The $\ell_p$-cost for $u$ that are pre-clustered (thus are necessarily in $V_0$) of the edges $uv \in E^+$
in disagreement with respect to $\mathcal{C}_{\textsf{ALG}}$, where $u \succ v$, is bounded by   
\[\sum_{u \in V_0}\big |\{v \in V_0 \cap N_u^+ :   u \succ v\} \big |^p  \leq O\left ( \left ( \nicefrac{1}{\eps^8} \cdot \log^4 n \right )^p\right )\cdot \opt^p_p.\]
\end{lemma}

\begin{proof}[Proof of Lemma \ref{lem: type1b-lp-backwards}]
This proof will follow the same structure as that of Lemma \ref{lem: type1b-lp}, but the fact that $v$ is clustered \emph{after} $u$ (which is the differentiating factor from Lemma \ref{lem: type1b-lp})  plays a key role in the analysis. Note that any $u$ for which $\{v \in V_0 \cap N_u^+ : u \succ v\} \neq \emptyset$ is necessarily pre-clustered, by definition of $\succ$.
We partition the set $\{v \in V_0 \cap N_u^+ \mid  u \succ v\}$
depending on whether $\tilde{d}_{uv}>t$ or 
$\tilde{d}_{uv}\leq t$. The random set of $u$'s close, positive neighbors that are clustered \emph{after} $u$ is

\[E_1(u) := \{v \in N_u^+ \cap V_0 :  u \succ v \} \cap \ball_{\tilde{d}}(u,t).\]

Define $E_2(u)$ to be the remaining positive neighbors of $u$ that are clustered \emph{after} $u$:

\textbf{\[E_2(u) := \{v \in N_u^+ \cap V_0 :  u \succ v \} \setminus E_1(u).\]}

Then as in Lemma
\ref{lem: type1b-lp}, we partition the sum and apply Jensen's inequality to see
\begin{align}
    \sum_{u \in V_0}\big | \{v \in V_0 \cap N_u^+ : u \succ v\} \big |^p   
    &\leq   2^{p-1} \cdot \sum_{u \in V_0} | E_1(u)|^p+ 2^{p-1} \cdot \sum_{u \in V_0}  |E_2(u)|^p. \label{eqn: partition-lem-v-first-backward}
\end{align}

Bounding $\sum_{u \in V_0}|E_2(u)|^p$ is the same as in the proof of Lemma \ref{lem: type1b-lp}, as that bound does not rely on whether $u \succ v$ or $v \succ u$.  Specifically, line (\ref{eqn:bound-e1}) is also an upper bound on $\sum_{u \in V_0}|E_2(u)|^p$, so
\begin{align}\label{eqn: e1(u)-bound}
    \sum_{u \in V_0}|E_2(u)|^p \leq \left ( \nicefrac{(7215 \cdot C^3 \cdot \log^3 n )}{\eps^6}\right ) ^p   \cdot  \opt_p^p. 
\end{align}

We now case on $|E_1(u)|$:
\begin{align*}
 \sum_{u \in V_0}  |E_1(u)|^p & =  \underbrace{\sum_{\overset{u \in V_0 :}{ |E_1(u)| < 2C \log n /\eps^2}}   |E_1(u)|^p}_{E_{1a}} +    \underbrace{\sum_{\overset{u \in V_0 : }{ |E_1(u)| \geq 2C \log n /\eps^2}} |E_1(u)|^p}_{E_{1b}}
\end{align*}

Note that we may assume $u$ is pre-clustered for all $u$ in the summations above, since if $E_1(u) \neq \emptyset$, then $u$ is necessarily pre-clustered, by definition of $E_1(u)$.

\paragraph*{Bounding $E_{1a}$.} 
 As in the proof of Lemma \ref{lem: type1b-lp}, $E_{1a}$ is easier to bound, because the number of disagreements incident to each $u$ in the sum is small, but again we need to show there is sufficient fractional cost to which to charge these disagreements so that we can then invoke Lemma \ref{lem: whp-pos-frac-cost}. We use the following claim, Claim \ref{clm: sufficient-charge-backwards}, which is the analogue of Claim \ref{clm: sufficient-charge} in Lemma \ref{lem: type1b-lp}; the difference here is that the edge $(u, s^*(u))$ contributes to the fractional cost of $\tilde{d}$, rather than the edges $(u, s^*(v))$ for $v \in E_1(u)$. 

\begin{claim}\label{clm: sufficient-charge-backwards}
    If $E_1(u) \neq \emptyset$, then $\tilde{D}_0(u) \geq \nicefrac{3}{20}$.
\end{claim}

\begin{proof}[Proof of Claim \ref{clm: sufficient-charge-backwards}]
We will lower bound both $\tilde{d}_{u \hspace{1pt}s^*(u)}$ and $1-\tilde{d}_{u\hspace{1pt}s^*(u)}$ by $\nicefrac{3}{20}$, and argue that $s^*(u) \in V_0$. Moreover, since $E_1(u) \neq \emptyset$, $u$ is pre-clustered, so $u \in V_0$. Thus the claim follows. 

First, since $\tilde{d}_{u\hspace{1pt}s^*(u)} \leq c \cdot r$, we have that $1-\tilde{d}_{u\hspace{1pt}s^*(u)} \geq 1-c \cdot r$. Then, fix some $v \in E_1(u)$, and apply the approximate triangle inequality (Lemma \ref{lem: tri-inequality}) to see that 
$\tilde{d}_{v\hspace{1pt}s^*(u)} \leq \delta (\tilde{d}_{uv} + \tilde{d}_{u\hspace{1pt}s^*(u)})$. Substituting in $\tilde{d}_{uv} \leq t$ and $\tilde{d}_{v\hspace{1pt}s^*(u)} > c \cdot r$, we see $c \cdot r \leq \delta (t+ \tilde{d}_{u\hspace{1pt}s^*(u)})$, which rearranging shows 
$\frac{c \cdot r}{\delta}-t \leq  \tilde{d}_{u\hspace{1pt}s^*(u)}$. 

It just remains to show that $s^*(u) \in V_0$, which follows from the contrapositive of Fact \ref{fct: frac-isolation}, because $\tilde{d}_{u\hspace{1pt}s^*(u)}<1$ and $u \neq s^*(u)$ since $d_{u \hspace{1pt}s^*(u)}>0$.

In total, since we take $t = \frac{r}{2 \delta}$, we obtain $\tilde{D}_0(u) \geq \min\{1-c \cdot r, \frac{c \cdot r}{\delta}-t\} \geq \nicefrac{3}{20}.$
\end{proof}

We can now bound $E_{1a}$ in the next claim.
\begin{claim}
    \label{claim: E1a-backwards}
Condition on the good event $B^c$. For $t = \frac{r}{2 \delta}$, it is the case that 
\[E_{1a} := \sum_{\overset{u \in V_0 :}{ |E_1(u)| < 2C \log n /\eps^2}} |E_1(u)|^p \leq  
  O\left (\left (\nicefrac{1}{\eps^8} \cdot \log^4 n\right )^p \right ) \cdot \opt_p^p. \]
\end{claim}

\begin{proof}[Proof of Claim \ref{claim: E1a-backwards}]

If $E_1(u) = \emptyset$, it contributes nothing to the sum. For $E_1(u) \neq \emptyset$, we apply Claim \ref{clm: sufficient-charge-backwards} to charge to $\tilde{D}_0(u)$ and use the upper bound on $|E_1(u)|$.
Then we apply Corollary \ref{cor: bounded-exp-frac-cost}. Together, this leads to the following string of inequalities:
\begin{align}
\sum_{\overset{u \in V_0:  }{ |E_1(u)| \leq 2C\log n/\eps^2}} |E_1(u)|^p 
&\leq 
 \Bigg (\nicefrac{(40 \cdot C\log n)}{3\eps^2} \Bigg )^p \cdot \sum_{u \in V_0 }  \big (\tilde{D}_0(u)\big )^p \notag \leq  
\left (\nicefrac{(1480 \cdot C^4 \cdot \log^4 n )}{\eps^8}\right )^p \cdot \opt_p^p. \notag
 \label{eqn: e22-bound-small}
\end{align}
\end{proof}

\paragraph*{Bounding $E_{1b}$.} 
Recall
\[E_{1b} := \sum_{\overset{u \in V_0 : }{ |E_1(u)| \geq 2C \log n /\eps^2}} |E_1(u)|^p = \sum_{\overset{u \in V_0 : }{ |E_1(u)| \geq 2C \log n /\eps^2}}|\{v \in N_u^+ : u \succ v, \tilde{d}_{uv} \leq t\}|^p\]

We first show $|E_1(u)| = O(|\ball_{\tilde{d}}(s^*(u),r)|)$. 
This bound is useful because, although $s^*(u)$ was chosen as the cluster center for $u$, both $s^*(u)$ and the nodes that are close to it with respect to $\tilde{d}$ must actually be relatively far from $u$. Intuitively, this is because the nodes $v \in E_1(u)$ are all close to $u$, so $u$ cannot be very close to $s^*(u)$---otherwise, those nearby nodes $v$ would have been assigned to the same cluster as $s^*(u)$, which they were not. See Figure \ref{fig: e1b-lp}.

\begin{figure}
    \centering
  \captionsetup{width=.9\textwidth}
    \includegraphics[width=5cm]{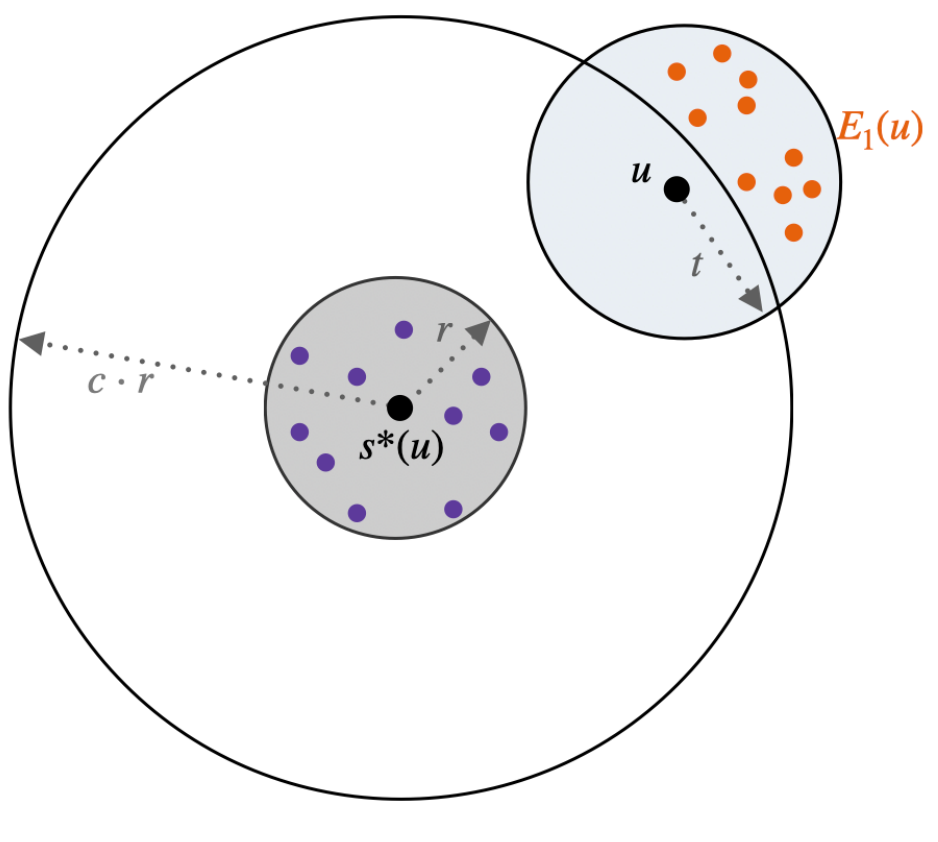}    \caption{Bounding $E_{1b}$ in the proof of Lemma \ref{lem: type1b-lp-backwards}. We show all the purple nodes are sufficiently far from $u$.
    Then, we will show that the number of orange nodes, $|E_1(u)|$, is roughly equal to the number of purple nodes, $\ball_{\tilde{d}}(s^*(u),r)$, so we can charge the disagreements $uv$ for $v \in E_1(u)$ to the cost of $\tilde{d}$ on edges $uw$ for $w \in \ball_{\tilde{d}}(s^*(u),r)$. 
    }
    \label{fig: e1b-lp}
\end{figure}

In the follow claim, we show $|E_1(u)| = O(|\ball_{\tilde{d}}(s^*(u), r)|)$. Then, we see all $a \in \text{Ball}_{\tilde{d}}(s^*(u),r)$ have sufficiently far distance from $u$, and so we can charge to those distances. 

\begin{claim}\label{claim: ku-ball-bound}
    Condition on the good event $B^c$. For $u$  with $|E_1(u)| \geq 2C \log n/\eps^2$, it is the case that $|E_1(u)| \leq \nicefrac{8}{\eps^2} \cdot |\ball_{\tilde{d}}(s^*(u),r)|$. 
\end{claim}

 \begin{proof}[Proof of Claim \ref{claim: ku-ball-bound}]
By the good event, since $|\ball_{\tilde{d}}(u,t)| \geq |E_1(u)| \geq 2C \log n/\eps^2 \geq C'\log_{1/(1-\eps^2/2)} n$ is sufficiently large, there is some sampled center in $\ball_{\tilde{d}}(u,t)$, i.e. there exists some $z \in \ball_{\tilde{d}}^{S_p}(u,t)$.
Since all $v \in E_1(u)$ have $\tilde{d}_{vz} \leq \delta (\tilde{d}_{vu} +\tilde{d}_{uz} ) \leq 2t \delta =r$, we have $E_1(u) \subseteq \ball_{\tilde{d}}(z,r)$ and in particular,
all $v \in E_1(u) \cap S_b$ are close enough to $z$ to be counted in $\ball_{\tilde{d}}^{S_b}(z,r)$. But since $v \in E_1(u)$ are clustered after $u$ it must be that 
\begin{equation}\label{eq: z-s*u}
    \big | \ball^{S_b}_{\tilde{d}}(z,r)\big | \leq \big |\ball^{S_b}_{\tilde{d}}(s^*(u),r)\big |\leq \big |\ball_{\tilde{d}}(s^*(u),r)\big |.
\end{equation}

Since $|E_1(u)| > 2C \log n / \eps^2$ and $\text{Ball}_{\tilde{d}}(z,r) \supseteq E_1(u)$, we have $ |\text{Ball}_{\tilde{d}}(z,r)| > 2C \log n / \eps^2$. Then we use the good event to see $\big |\ball^{S_b}_{\tilde{d}}(z,r) \big |\geq \frac{q(\eps)}{4} \big |\ball_{\tilde{d}}(z,r)\big |$. Combining these inequalities, we see
\begin{equation}\label{eq: z-s*u-2}
|E_1(u)|  \leq  \big |\ball_{\tilde{d}}(z,r)\big |   \leq \nicefrac{4}{q (\eps)} \cdot \big |\ball^{S_b}_{\tilde{d}}(z,r)\big |.
\end{equation}
Then combining the inequalities in (\ref{eq: z-s*u}) and (\ref{eq: z-s*u-2}) gives the claim.
 \end{proof}

We note the proof of the following claim is similar in structure to that of Claim \ref{clm: upper-bd-Au}, but we charge to different nodes. 

\begin{claim} \label{clm: upper-bd-Au-backward}
    If $E_1(u) \neq \emptyset$, then $|\text{Ball}_{\tilde{d}}(s^*(u),r)|  \leq 17 \cdot \tilde{D}_0(u).$
\end{claim}

\begin{proof}[Proof of Claim \ref{clm: upper-bd-Au-backward}]
    It suffices to show that $\tilde{d}_{ua} \geq \nicefrac{1}{17}$ and $1-\tilde{d}_{ua} \geq \nicefrac{1}{17}$ for every $a \in \text{Ball}_{\tilde{d}}(s^*(u),r)$. Then we will have (by Fact \ref{fct: frac-isolation}) that $a \in V_0$ (because $\tilde{d}_{ua} < 1$) for every $a \in \text{Ball}_{\tilde{d}}(s^*(u),r)$, and thus the claim follows. 

    Since $\tilde{d}_{u\hspace{1pt}s^*(u)} \geq \frac{c \cdot r}{\delta}-t$ (by the proof of Claim \ref{clm: sufficient-charge-backwards}), by Lemma \ref{lem: tri-inequality} we have that 
    for $a \in \text{Ball}_{\tilde{d}}(s^*(u),r)$, 
    \[ \frac{c \cdot r}{\delta}-t\leq \tilde{d}_{u\hspace{1pt}s^*(u)}\leq \delta ( \tilde{d}_{a\hspace{1pt}s^*(u)} +\tilde{d}_{ua}) \leq \delta (r+\tilde{d}_{ua}),\]
    which rearranging gives us that
    $ \frac{c \cdot r}{\delta^2}-\frac{t}{\delta}-r\leq  \tilde{d}_{ua}$. For our choice of constants, $  \tilde{d}_{ua} \geq \nicefrac{1}{17}$. 

    To lower bound $1-\tilde{d}_{ua}$, we again use  Lemma \ref{lem: tri-inequality} to see
    \[\tilde{d}_{ua} \leq \delta (\tilde{d}_{u\hspace{1pt}s^*(u)} + \tilde{d}_{a\hspace{1pt}s^*(u)}) \leq \delta (c+1)r,\]
    which rearranged is $1-\tilde{d}_{ua} \geq 1-\delta r (c+1) \geq \nicefrac{1}{17}$.
\end{proof}

We can now bound $E_{1b}$ in the next claim.
\begin{claim}
    \label{claim: E1b-backwards}
Condition on the good event $B^c$. For $t = \frac{r}{2 \delta}$, it is the case that 
\[E_{1b} := \sum_{\overset{u \in V_0 :}{ |E_1(u)| \geq 2C \log n /\eps^2}} |E_1(u)|^p \leq  
O \left (\left ( \nicefrac{1}{\eps^8} \cdot \log^3 n\right)^p\right) \cdot \opt_p^p . \]
\end{claim}

\begin{proof}[Proof of Claim \ref{claim: E1b-backwards}]
Now we combine the above claims with the bound on the cost of $\tilde{d}$ on $V_0$ to see
\begin{align}
    E_{1b} &\leq 
    \left ( \nicefrac{8}{\eps^2} \right )^p\cdot \sum_{u \in V_0}  \big | \ball_{\tilde{d}}(s^*(u),r)\big |^p \leq   
     \left (\nicefrac{136}{\eps^2} \right )^p \cdot \sum_{u \in V_0} \Big (\tilde{D}_0(u) \Big )^p \leq  \left ( \nicefrac{(15096 \cdot C^3 \cdot \log^3 n)}{\eps^8}\right)^p \cdot \opt_p^p. \label{eqn: e22-big}
\end{align}
The first inequality is from Claim \ref{claim: ku-ball-bound}, the second is from  Claim \ref{clm: upper-bd-Au-backward}, and the last is from Corollary \ref{cor: bounded-exp-frac-cost}.
\end{proof}

\medskip

Continuing the bound from line (\ref{eqn: partition-lem-v-first-backward}), we substitute in the inequalities from line
(\ref{eqn: e1(u)-bound}) and Claims \ref{claim: E1a-backwards} and \ref{claim: E1b-backwards}, and use the fact that $C$ is sufficiently large by the good event, to see that
\[\sum_{u \in V_0}\big |\{v \in V_0 \cap N_u^+ \mid   u \succ v\} \big |^p  \leq
  \left ( \nicefrac{(3414 \cdot C^4 \cdot \log^4 n )}{\eps^8}\right )^p\cdot \opt^p_p.
\]
\end{proof}

\subsubsection{Cost of negative edges} \label{subsec: lp-neg-edge}
The edges $uv \in E^-$, where at least one endpoint is pre-clustered, that are in disagreement with respect to $\mathcal{C}_{\textsf{ALG}}$ are those where $u$ is clustered with $v$. The proof actually follows easily.

\begin{lemma}\label{lem: negative-lp}
Condition on the good event $B^c$ and  fix $1 \leq p < \infty$. The $\ell_p$-cost for $u$ in $V_0$ of the negative edges in the Pre-clustering phase is bounded by
\[ \sum_{u \in V_0}\Big |  \big \{v \in N_u^- : v \text{ clustered with } u \big \}\Big |^p \leq    O \left (\left ( \nicefrac{1}{\eps^2} \cdot \log n \right )^p \right ) \cdot \opt_p^p.\]
\end{lemma}

\begin{proof}[Proof of Lemma \ref{lem: negative-lp}]
    If $u,v \in V_0$ are clustered together, there exists $s^* \in S_p$ such that $s^* = s^*(u) = s^*(v)$ (possibly with $s^* = u$ or $s^* = v$). By the approximate triangle inequality (Lemma \ref{lem: tri-inequality}),
    \[\tilde{d}_{u,v} \leq \delta \cdot (\tilde{d}_{u \hspace{1pt}s^*} + \tilde{d}_{v\hspace{1pt}s^* }) \leq 2\delta c r \leq 7/10.\] 
    Since $\tilde{d}_{uv} < 1$, we know that $\tilde{d}_{uv} = \bar{d}_{uv}$ and also that $u \in R_2$, where we recall that $R_2 = V \setminus R_1$, for $R_1$  the set of vertices that are isolated by $\tilde{d}$. 
    Thus, to prove the lemma, it suffices to bound $\sum_{u \in V_0 \cap R_2}|R_1(u)|^p$, where
    $R_1(u) := \{v \in N_u^- :  \bar{d}_{uv} \leq 7/10\}.$
    Since we bound this exact quantity in Lemma 
    \ref{lem: whp-neg-frac-cost}, we find that 
    \[\sum_{u \in V_0 \cap R_2}|R_1(u)|^p\ \leq  \left ( \nicefrac{(74 \cdot C \cdot \log n)}{\eps^2} \right )^p \cdot \opt_p^p.\]
\end{proof}

\subsection{Cost of Pivot phase}\label{sec: pivot-phase-lp}

Let $G' = (V', E')$ be the subgraph induced by the vertices that are \emph{not} pre-clustered in Algorithm \ref{alg: main-alg}. In this section, we bound the cost of disagreements in $G'$. 
Recall that $V_0$ is the set of eligible vertices, i.e., those vertices $v \in V$ such that $|N_v^+ \cap S_d| \neq \emptyset$. So $V'$ contains the vertices that are \textit{not} eligible (those in $\widebar{V_0} = V \setminus V_0$), as well as vertices that are eligible but that are far from all vertices in $S_p$:
\[V' := \widebar{V_0} \cup V_0' \text{, where } V_0' = V_0 \cap \{v \in V: \tilde{d}_{v u_i} > c \cdot r  \text{ for all } u_i \in S_p\} .\]

Algorithm \ref{alg: main-alg} runs the standard Pivot algorithm on $G[\widebar{V_0}]$, and runs Modified Pivot on $G[V_0']$. In Lemma \ref{lem: pivot-cost-lp-v0}, we bound the disagreements incurred by Pivot on $G[\widebar{V_0}]$, and in Lemma \ref{lem: pivot-cost-lp-v''} we bound the disagreements incurred by Modified Pivot on $G[V_0']$.

We note the arguments in this section may look rather different than those in the Pre-clustering phase. This is a consequence of the fact that we are using totally different clustering subroutines in each phase. In this Pivot phase, we use two versions of the (modified) Pivot algorithm. Therefore, the analysis is more combinatorial; often the charging arguments use ``bad triangles'' as intermediaries:

\begin{definition}\label{def: bad-tri}
    A \emph{bad triangle} is a triple $uvw$ such that $uv, uw \in E^+$ and $vw \in E^-$.
\end{definition}

Note that every clustering must incur a disagreement on at least one edge in a bad triangle. 

\begin{definition}
    In Algorithm \ref{alg: pivot}, for $v_i \in \widebar{V_0}$ (analogously, $v_i \in V_0'$), we define \emph{$v_i$'s pivot} to be $v_j^*$ if the \emph{\textbf{if}} statement holds, and to be $v_i$ otherwise. 
\end{definition}

\subsubsection{Disagreements in $G[\widebar{V_0}]$}

To bound disagreements incident to $ u \in \widebar{V_0}$, ideally we would  relate the set of bad triangles that contain $u$ to the disagreements incident to $u$ in $\mathcal{C}_{\textsf{ALG}}$.
However, while the optimal must make a disagreement on each bad triangle it does not necessarily have any disagreements incident $u$. So in effect, we have to charge some of our disagreements incident to $u$ to the optimal solution's disagreements on other vertices. 

\medskip

As before, fix an optimal clustering $\mathcal{C}_{\textsf{OPT}}$ (for the entire graph $G$) for any fixed $\ell_p$-norm. 
Let $\textsf{OPT}(u)$ be the (positive or negative) neighbors inducing disagreements with $u$ in $\mathcal{C}_{\textsf{OPT}}$:
\[\textsf{OPT}(u) := \{v \in V \mid uv \text{ a disagreement in }\mathcal{C} _{\textsf{OPT}}\}.\]

Analogously, for $u \in \widebar{V_0}$, define $\textsf{Pivot}(u)$ to be the (positive or negative) neighbors inducing disagreements with $u$, \emph{restricted to the clusters formed by the Pivot phase of Algorithm \ref{alg: main-alg} on $G[\widebar{V_0}]$}: 
\[\textsf{Pivot}(u) := \{v \in \widebar{V_0} \mid uv \text{ a disagreement in }\mathcal{C}_{\textsf{ALG}} \}.\]

\begin{lemma} \label{lem: pivot-cost-lp-v0}
    Condition on the good event $B^c$ and fix $1 \leq p < \infty$. 
    The $\ell_p$-cost for $u \in \widebar{V_0}$ of the edges $uv$ in disagreement with respect to $\mathcal{C}_{\textsf{ALG}}$ for $v \in \widebar{V_0}$,
is bounded by 
\[\sum_{u \in \widebar{V_0}}\big |\textsf{Pivot}(u) | ^p  \leq O \left (\left(\nicefrac{1}{\eps^2} \cdot \log n \right )^{p} \right ) \cdot \opt_p^p. \]
\end{lemma}

\begin{proof}[Proof of Lemma \ref{lem: pivot-cost-lp-v0}]

Let $\mathcal{T}$ denote the set of bad triangles in $G[\widebar{V_0}]$, and let $\mathcal{T}(u)$ denote the set of bad triangles in $G[\widebar{V_0}]$ that contain vertex $u$.
As mentioned before the start of the proof, to bound disagreements $uv$ where both $u$ and $v$ are in $\widebar{V_0}$, we would like to relate $|\mathcal{T}(u)|$  to $|\textsf{Pivot}(u)| $. However, this is not possible, so instead we charge some of the disagreements  incident to $u$  incurred by our Pivot phase to disagreements the optimal solution incurs on other vertices.

Let $\mathcal{T}_\textsf{OPT}(u) \subseteq \mathcal{T}(u)$ be the bad triangles $T$ for which $\mathcal{C}_{\textsf{OPT}}$ has a disagreement incident to $u$ in $T$, and let $\overline{\mathcal{T}_{\textsf{OPT}}(u)} = \mathcal{T}(u)  \setminus \mathcal{T}_\textsf{OPT}(u) $ be the remaining bad triangles in $G[\widebar{V_0}]$ containing $u$. 
We observe that for
every $u \in \widebar{V_0}$, every disagreement in $G[\widebar{V_0}]$ incident to $u$ can be mapped to some $T \in \mathcal{T}(u)$ --  namely, the unique bad triangle containing the disagreement and $u$'s pivot. 
Moreover, this mapping is injective because Pivot  incurs \emph{exactly} one disagreement on each bad triangle.  Therefore we can bound the disagreements that our algorithm makes in $G[\widebar{V_0}]$ that are incident to $u$ by applying Jensen's inequality, 
\begin{align}
    \sum_{u \in \widebar{V_0}}|\textsf{Pivot}(u)|^p  &\leq \sum_{u \in \widebar{V_0}} |\mathcal{T}(u)|^p  
    \leq 2^{p-1} \cdot \underbrace{\sum_{u \in \widebar{V_0}}|\mathcal{T}_\textsf{OPT}(u)|^p }_{S_1} + 2^{p-1} \cdot \underbrace{\sum_{u \in \widebar{V_0}} \Big|\overline{\mathcal{T}_\textsf{OPT}(u)}\Big|^p }_{S_2}. \label{eqn: bad-tri-combine}
\end{align}

Recall that $u \in \widebar{V_0}$ because $N_u^+ \cap S_d = \emptyset$. So by the good event, it must be that $|N_u^+| < C \log n / \varepsilon^2$. This bound on $|N_u^+|$ will repeatedly be used. 

First we bound $S_1$, which will be simpler to bound since these triangles directly correspond to a disagreement that $\mathcal{C}_{\textsf{OPT}}$ has on $u$.
\begin{claim} \label{clm: lp-T-opt}
    $\sum_{u \in \widebar{V_0}}|\mathcal{T}_\textsf{OPT}(u)|^p \leq 
    O \left (\left(\nicefrac{1}{\eps^2} \cdot \log n \right )^{p} \right ) \cdot \opt_p^p.$
\end{claim}

\begin{proof}[Proof of Claim \ref{clm: lp-T-opt}]
Fix $u \in \widebar{V_0}$. We use the definition of $\mathcal{T}_\textsf{OPT}(u)$ to see that
\[|\mathcal{T}_\textsf{OPT}(u)| = \sum_{T \in \mathcal{T}_\textsf{OPT}(u)} 1 \leq \sum_{v \in \textsf{OPT}(u) \cap \widebar{V_0}} \quad \sum_{\overset{T \in \mathcal{T}_\textsf{OPT}(u):}{uv \in T}} 1 \leq \sum_{v \in \opt(u) \cap \widebar{V_0}} (|N_u^+ \cap \widebar{V_0}| + |N_v^+ \cap \widebar{V_0}|). \]
 Then we use the bound $|N_u^+ \cap \widebar{V_0}| \leq |N_u^+| \leq C \log n / \varepsilon^2$ for $u \in V_0$ and the fact that $|\textsf{OPT}(u) \cap \widebar{V_0}| \leq y(u)$ to upper bound $S_1$:
\begin{align*}
S_1 =\sum_{u \in \widebar{V_0}} |\mathcal{T}_{\textsf{OPT}}(u)|^p &\leq  \sum_{u \in \widebar{V_0}} \left(|\textsf{OPT}(u) \cap \widebar{V_0}| \cdot \nicefrac{(2C \cdot \log n)}{\varepsilon^2}  \right)^p
\leq \left (\nicefrac{(2C \cdot \log n)}{\varepsilon^2} \right)^p \cdot \opt_p^p.
\end{align*}
\end{proof}

It remains to bound $S_2$, and this sum contains the bad triangles which we will charge to disagreements not incident to $u$.

\begin{claim} \label{clm: T-opt-comp}
    $\sum_{u \in \widebar{V_0}} \Big|\overline{\mathcal{T}_\textsf{OPT}(u)}\Big|^p  \leq  
    O \left (\left(\nicefrac{1}{\eps^2} \cdot \log n \right )^{p} \right ) \cdot \opt_p^p.$
\end{claim}

\begin{proof}[Proof of Claim \ref{clm: T-opt-comp}]
Fix $u \in \widebar{V_0}$. By definition, for every $T \in \overline{\mathcal{T}_\textsf{OPT}(u)}$, $\mathcal{C}_{\textsf{OPT}}$ has an edge in disagreement on the \emph{unique} edge of $T$ not incident to $u$. So, no other triangle in $\overline{\mathcal{T}_\textsf{OPT}(u)}$ contains this edge as a disagreement. Moreover, by the definition of a bad triangle, one of the endpoints of this disagreeing edge is a positive neighbor of $u$ in $G[\widebar{V_0}]$. 
So we have by the discussion above that 

\begin{align}
    S_2  = \sum_{u \in \widebar{V_0}} \Big|\overline{\mathcal{T}_\textsf{OPT}(u)}\Big|^p &\leq \sum_{u \in \widebar{V_0}} \left (\sum_{\widebar{V_0} \cap N_u^+}|\opt(v) \cap \widebar{V_0}| \right )^p \notag \\
    &\leq  \sum_{u \in \widebar{V_0}} |N_u^+|^{p-1}\sum_{v\in \widebar{V_0} \cap N_u^+}|\opt(v) \cap \widebar{V_0}|^p  \label{eq: jensen}\\
    &\leq  \sum_{v \in \widebar{V_0}}|\opt(v) \cap \widebar{V_0}|^p \sum_{ u \in \widebar{V_0}\cap N_v^+ } |N_u^+|^{p-1}\notag\\
    &\leq   \left(\nicefrac{C}{\eps^2} \cdot \log n \right )^{p} \cdot \opt_p^p. \label{eq: small-nbhds}
\end{align}

Line (\ref{eq: jensen}) follows from Jensen's inequality.  Line (\ref{eq: small-nbhds}) follows from the fact that because we conditioned on the good event $B^c$, the maximum positive degree of any $u \in \widebar{V_0}$ is $\frac{C}{\eps^2} \cdot \log n $ and for $v \in \widebar{V_0}$, there are at least $|\opt(v) \cap \widebar{V_0}|$ disagreements incident to $v$ in $\mathcal{C}_{\textsf{OPT}}$ by definition of $\opt(v)$.  
\end{proof}
The lemma statement follows from the claims, since continuing from line (\ref{eqn: bad-tri-combine}), we see that
\begin{align*}
    \sum_{u \in \widebar{V_0}}\big|\textsf{Pivot}(u)\big|^p   
    &\leq 2^{p-1} \cdot \left (\nicefrac{2C }{\varepsilon^2} \cdot \log n \right)^p \cdot \opt_p^p+ 2^{p-1} \cdot \left(\nicefrac{C}{\eps^2} \cdot \log n \right )^{p} \cdot \opt_p^p \\
    & \leq 
\left(\nicefrac{4C}{\eps^2} \cdot \log n \right )^{p} \cdot \opt_p^p. 
\end{align*}

\end{proof}

\subsubsection{Disagreements in $G[V_0']$}

Next we will bound the disagreements in $G[V_0'].$ Recall that the vertices in $V_0'$ are those that have sufficiently large positive neighborhood sampled in $S_d$, but were far from all cluster centers.
As has been the case for other disagreement types, the disagreements whose cost is most difficult to bound are on the positive edges $uv$, where $\tilde{d}_{uv}$ is quite small. Here, we are able to charge $uv$ to some other edges $uw$, for $uvw$ a bad triangle. 

As before, fix an optimal clustering $\mathcal{C}_{\textsf{OPT}}$ (for the entire graph $G$) for any fixed $\ell_p$-norm. 
Let $\textsf{OPT}(u)$ to be the (positive or negative)  disagreements incident to $u$ in $\mathcal{C}_{\textsf{OPT}}$:
\[\textsf{OPT}(u) := \{uv \mid uv \text{ a disagreement in }\mathcal{C} _{\textsf{OPT}}\}.\]

For $u \in V_0'$, define $\textsf{Pivot}(u)$ to be the (positive or negative) disagreements incident to $u$, \emph{restricted to the clusters formed by the (Modified) Pivot phase of Algorithm \ref{alg: main-alg} on $G[V_0']$}: 
\[\textsf{Pivot}(u) := \{uv \mid v \in V_0',  uv \text{ a disagreement in }\mathcal{C}_{\textsf{ALG}} \}.\] 

Note that both sets are sets of \emph{edges}, unlike in Lemma \ref{lem: pivot-cost-lp-v0} where the analogous sets are sets of vertices. 

We write $\textsf{Pivot}$ for brevity, but recall that the algorithm on $G[V_0']$ is actually a modified version of the classic Pivot algorithm, as the pivots in our algorithm grab positive neighbors that are additionally required to be nearby with respect to $\tilde{d}$ (see the definition of $E_c$ in the \textbf{else} statement for $V_0' = V' \setminus \widebar{V_0}$ in Algorithm \ref{alg: main-alg}).

\begin{lemma} \label{lem: pivot-cost-lp-v''}
    Condition on the good event $B^c$ and fix $1 \leq p < \infty$. The $\ell_p$-cost for $u \in V_0'$ of the edges $uv$ in disagreement with respect to $\mathcal{C}_{\textsf{ALG}}$ for $v \in V_0'$
is bounded by 
  \[\sum_{u \in V_0'}\big |\textsf{Pivot}(u)\big | ^p  \leq  O \left (\left(\nicefrac{1}{\eps^8} \cdot \log^4 n\right)^p \right )  \cdot \opt_p^p.\] 
\end{lemma}

\begin{proof}[Proof of Lemma \ref{lem: pivot-cost-lp-v''}]

    All disagreements in $\textsf{Pivot}(u) \cap \opt(u)$ can be charged directly to $\opt(u)$, so we focus on $\textsf{Pivot}(u) \cap \overline{\opt(u)}$.
    We partition the remaining disagreements into several sets:
    \begin{align*}
\textsf{Pivot}(u) \cap \overline{\opt(u)} &= \underbrace{\big \{uv \in \textsf{Pivot}(u)\cap  \overline{\opt(u)} \cap E^+ \mid \tilde{d}_{uv} \geq cr \big\}}_{S_1(u)}  \\
&\dot\cup \underbrace{\big  \{uv \in \textsf{Pivot}(u) \cap \overline{\opt(u)}\cap E^+ \mid \tilde{d}_{uv} < cr \big\}}_{S_2(u)}  \dot \cup \underbrace{\big (\textsf{Pivot}(u)\cap \overline{\opt(u)} \cap E^- \big)}_{S_3(u)} .
\end{align*}
So all together, we can apply Jensen's inequality to see that
\begin{align}
    \sum_{u \in V_0'} 
    \big |\textsf{Pivot}(u)  \big |^p &\leq 4^p \cdot \sum_{u \in V_0'} 
    \big |\textsf{Pivot}(u)   \cap \opt(u) \big |^p  \notag \\
    &+4^p \cdot \underbrace{\sum_{u \in V_0'} 
    \big |S_1(u)\big |^p}_{S_1}+ 4^p \cdot \underbrace{\sum_{u \in V_0'} 
    \big |S_2(u)\big |^p}_{S_2}+ 4^p \cdot \underbrace{\sum_{u \in V_0'} 
    \big |S_3(u)\big |^p}_{S_3}.\label{eqn: combine-pivot-V_0'}
\end{align}

\noindent \textbf{Bounding $S_1$.}
As has been the case for other types of disagreements, bounding the cost of edges that have large $\tilde{d}$ is relatively straightforward.

We charge the cost of all $uv \in \textsf{Pivot}(u)\cap  \overline{\opt(u)} \cap E^+$ with $\tilde{d}_{uv} \geq c \cdot r$ to $\tilde{d}$. Specifically, we 
use that $V_0' \subseteq V_0$ and $\textsf{Pivot}(u) \subseteq V_0 \cap N_u^+$, and then apply Lemma \ref{lem: whp-pos-frac-cost}.
\begin{align}
    S_1 = \sum_{u \in V_0'} |S_1(u)|^p  &\leq\frac{1}{(cr)^p}  \cdot 
    \sum_{u \in V_0'}  \Bigg (\sum_{uv \in \textsf{Pivot}(u) \cap \overline{\opt(u)} \cap E^+}  \tilde{d}_{uv} \Bigg)^p \notag \\
    & \leq \frac{1}{(cr)^p}  \cdot \sum_{u \in V_0} \bigg ( \sum_{v \in V_0 \cap N_u^+} \tilde{d}_{uv} \bigg )^p \leq   \left ( \nicefrac{(550 \cdot C^3 \cdot \log^3 n)}{\eps^6}
    \right )^p \cdot  \opt_p^p. \label{eqn: s1-pivot}
\end{align}

    \medskip

\noindent\textbf{Bounding $S_2$.}
For positive disagreements $uv$ with small $\tilde{d}_{uv}$, we go through bad triangles to find a suitable edge to charge.

    For $uv \in S_2(u)$, we have that $\tilde{d}_{uv} < c \cdot r$. Since $uv \in E^+$,  there are only two ways for $uv$ to be in $\textsf{Pivot}(u)$. 

    \begin{itemize}
    \item The first is if $uv$ is involved in a bad triangle with some $w$ which is $u$'s pivot but not $v$'s. In this case,  $uw \in E^+$ and $vw \in E^-$. In the second case,  $uv$ is involved in a bad triangle with some $w$ which is $v$'s pivot but not $u$'s. In this case,  $vw \in E^+$ and $uw \in E^-$. Since $uvw$ is a bad triangle, and $uv \not \in \opt(u)$, we have that either $wu$ or $wv$ is a disagreement in $\mathcal{C}_{\textsf{OPT}}$. We will charge the cost of the disagreement $uv$ made by $\mathcal{C}_{\textsf{ALG}}$ to whichever one is a disagreement in $\mathcal{C}_\opt$ (choosing one arbitrarily if both are disagreements).

    \item The second is if $uv$ is on a triangle $uvw$ of \textit{all} positive edges with some $w$ which is $u$'s pivot but not $v$'s in the case that  $\tilde{d}_{wv} \geq c \cdot r$ but $\tilde{d}_{wu} < c \cdot r$; or, $w$ which is $v$'s pivot but not $u$'s in the case that $\tilde{d}_{wu} \geq c \cdot r$ but $\tilde{d}_{wv} < c \cdot r$. In the former case, we can charge the cost of the disagreement $uv$ made by $\mathcal{C}_{\textsf{ALG}}$ to $wv$ as $\tilde{d}_{wv} \geq c \cdot r$, and in the latter case, to $wu$ as $\tilde{d}_{wu} \geq c \cdot r$.     \end{itemize}

    Call the subset of disagreements in $S_2(u)$ satisfying the former case $S_{2a}(u)$ and those satisfying the latter case $S_{2b}(u)$. 

    We start by bounding the contribution of all $S_{2a}(u)$.

    \begin{claim} \label{clm: pivot-lp-S2a}
        $\sum_{u \in V_0'}|S_{2a}(u)|^p \leq  O \left (\left (\nicefrac{1}{\eps^2} \cdot \log n\right )^p  \right )\cdot \opt_p^p.$
    \end{claim}

    \begin{proof}[Proof of Claim \ref{clm: pivot-lp-S2a}]
     For each disagreement in $\mathcal{C}_\opt$, we have to bound how many times it can be charged by an edge $uv$ in $S_{2a}(u)$. By the discussion above, a disagreement in  $\mathcal{C}_\opt$ can only be charged by $uv$ if it is of the form $uw$ or $wv$. If it is of the form $uw$, then it can only be charged by $uv$ with $v \in N_u^+ \cap \ball_{\tilde{d}}(u, c \cdot r)$. If it is of the form $wv$, then it can only be charged by once, namely, by $uv$ where again $v \in N_u^+ \cap \ball_{\tilde{d}}(u, c \cdot r)$. So 

    \[|S_{2a}(u)| \leq |\opt(u)| \cdot |N_u^+ \cap \ball_{\tilde{d}}(u, c\cdot r)| + \sum_{v \in N_u^+ \cap \ball_{\tilde{d}}(u, c \cdot r) \cap V_0'} |\opt(v)| \]
    and so by Jensen's inequality,
    \begin{align*}
        \sum_{u \in V_0'} |S_{2a}(u)|^p &\leq 2^{p-1} \cdot \sum_{u \in V_0'} |\opt(u)|^p \cdot |N_u^+ \cap \ball_{\tilde{d}}(u, c\cdot r)|^p \\
        &+ 2^{p-1} \cdot \sum_{u \in V_0'} \Bigg(\sum_{v \in N_u^+ \cap \ball_{\tilde{d}}(u, c \cdot r) \cap V_0'} |\opt(v)| \Bigg)^p.
    \end{align*}

    The first sum is bounded by $(C' \cdot \log_{1/(1-\varepsilon^2/2)} n)^p \cdot \opt_p^p$. This is because $u \in V_0'$, which implies that $\ball_{\tilde{d}}^{S_p}(u, c\cdot r) = \emptyset$, which in turn implies that $|\ball_{\tilde{d}}(u, c\cdot r)| \leq C' \cdot \log_{1/(1-\varepsilon^2/2)} n$, since we have conditioned on the good event $B^c$. To bound the second sum, we apply Jensen's inequality and then flip it:

    \begin{align*}
        \sum_{u \in V_0'} \Bigg(\sum_{v \in N_u^+ \cap \ball_{\tilde{d}}(u, c \cdot r)} |\opt(v)| \Bigg)^p &\leq \sum_{u \in V_0'} |N_u^+ \cap \ball_{\tilde{d}}(u, c \cdot r)|^{p-1} \sum_{v \in N_u^+ \cap \ball_{\tilde{d}}(u, \cdot r)} |\opt(v)|^p \\
        &\leq \sum_{v \in V_0'} |\opt(v)|^p \sum_{u \in V_0' \cap \ball_{\tilde{d}}(v, \cdot r)} |\ball_{\tilde{d}}(u, c \cdot r)|^{p-1} \\
        &\leq (C' \cdot \log_{1/(1-\varepsilon^2/2)} n)^p \cdot \opt_p^p
    \end{align*}
    where we have used the same reasoning as above to bound the inner sum by $(C' \cdot \log_{1/(1-\varepsilon^2/2)} n)^p$, due to the good event $B^c$. 

    Combining the bounds on the first and second sums, and doing a change of base finishes the claim with 
    $\sum_{u \in V_0'}|S_{2a}(u)|^p \leq   \left (\nicefrac{(4 \cdot C' \cdot \log n)}{\eps^2} \right )^p \cdot \opt_p^p.$

\end{proof}

We now bound the contribution of all $S_{2b}(u)$, so that we may then finish off the bound of $S_2$.

\begin{claim} \label{clm: pivot-lp-S2b}
$\sum_{u \in V_0'} |S_{2b}(u)|^p \leq  O \left (\left(\nicefrac{1}{\eps^8} \cdot \log^4 n\right)^p \right ) \cdot \opt_p^p$.
\end{claim}

\begin{proof}[Proof of Claim \ref{clm: pivot-lp-S2b}]
For each disagreement in $\mathcal{C}_\opt$, we have to bound how many times it can be charged by an edge $uv$ in $S_{2b}(u)$. By the discussion above, a disagreement in  $\mathcal{C}_\opt$ can only be charged by $uv$ if it is of the form $uw$ or $wv$. If it is of the form $uw$, then it can only be charged by $uv$ with $v \in N_u^+ \cap \ball_{\tilde{d}}(u, c \cdot r)$. If it is of the form $wv$, then it can only be charged by once, namely, by $uv$ where again $v \in N_u^+ \cap \ball_{\tilde{d}}(u, c \cdot r)$. So 

    \[|S_{2b}(u)| \leq \sum_{\overset{w \in N_u^+ \cap V_0':}{\tilde{d}_{uw} \geq c \cdot r}} |N_u^+ \cap \ball_{\tilde{d}}(u,c \cdot r)| + \sum_{v \in N_u^+ \cap \ball_{\tilde{d}}(u, c \cdot r) \cap V_0'} |\{w \in N_v^+ \cap V_0' : \tilde{d}_{wv} \geq c \cdot r \} | \]

and so by Jensen's inequality, 
\begin{align*}
    \sum_{u \in V_0'} |S_{2b}(u)|^p &\leq 2^{p-1} \cdot \sum_{u \in V_0'} \Bigg(\sum_{\overset{w \in N_u^+ \cap V_0':}{\tilde{d}_{uw} \geq c \cdot r}} |N_u^+ \cap \ball_{\tilde{d}}(u,c \cdot r)|\Bigg)^p \\
    &+ 2^{p-1} \cdot \sum_{u \in V_0'} \Bigg(\sum_{v \in N_u^+ \cap \ball_{\tilde{d}}(u, c \cdot r) \cap V_0'} |\{w \in N_v^+ \cap V_0' : \tilde{d}_{wv} \geq c \cdot r \} |\Bigg)^p
\end{align*}

The first sum is bounded by 
\begin{align*}
 \sum_{u \in V_0'}|N_u^+ \cap \ball_{\tilde{d}}(u,c \cdot r)|^p \cdot \left(\sum_{w \in N_u^+ \cap V_0'} \nicefrac{1}{c \cdot r} \cdot \tilde{d}_{uw}\right)^p
&\leq \left(\nicefrac{(2C' \cdot \log n)}{(c \cdot r \cdot \eps^2)}\right)^p \cdot \sum_{u \in V_0} \Bigg(\sum_{w \in N_u^+ \cap V_0} \tilde{d}_{uw}\Bigg)^p 
\end{align*}
where again we have used the conditioning on $B^c$ and the fact that $u \in V_0'$ to bound $|\ball_{\tilde{d}}(u, c \cdot r)|$. The second sum is bounded by 

\begin{align*}
    &\quad \sum_{u \in V_0'} \Bigg(\sum_{v \in N_u^+ \cap \ball_{\tilde{d}}(u, c \cdot r) \cap V_0'} |\{w \in N_v^+ \cap V_0' : \tilde{d}_{wv} \geq c \cdot r \} |\Bigg)^p \\
    &\leq\sum_{u \in V_0'} |\ball_{\tilde{d}}(u, c \cdot r)|^{p-1} \sum_{v \in N_u^+ \cap \ball_{\tilde{d}}(u, c \cdot r) \cap V_0'} \Bigg(\sum_{w \in N_v^+ \cap V_0'} \frac{1}{c \cdot r} \cdot \tilde{d}_{wv} \Bigg)^p \\
    &\leq \sum_{v \in V_0'} \Bigg(\sum_{w \in N_v^+ \cap V_0'} \frac{1}{c \cdot r} \cdot \tilde{d}_{wv} \Bigg)^p \sum_{u \in V_0' \cap \ball_{\tilde{d}}(v, c \cdot r)} |\ball_{\tilde{d}}(u, c \cdot r)|^{p-1} \\
        &\leq \left(\nicefrac{(C' \cdot \log_{1/(1-\varepsilon^2/2)} n)}{c \cdot r}\right)^p  \cdot \sum_{u \in V_0} \left(\sum_{w \in N_u^+ \cap V_0} \tilde{d}_{uw}\right)^p.
\end{align*}
    Combining the bounds  on the first and second sums, and applying Lemma \ref{lem: whp-pos-frac-cost}, finishes the claim with  $\sum_{u \in V_0'} |S_{2b}(u)|^p \leq \left(\nicefrac{(440 \cdot C'\cdot C^3 \cdot \log^4 n)}{\eps^8}\right)^p  \cdot \opt_p^p$.
\end{proof}

Tying it all together using Claims \ref{clm: pivot-lp-S2a} and \ref{clm: pivot-lp-S2b},
and using that $C$ is sufficiently large to ensure the good event, we have
\begin{equation}
S_2 
\leq 2^{p-1} \cdot \sum_{u \in V_0'}|S_{2a}(u)|^p + 2^{p-1} \cdot \sum_{u \in V_0'} |S_{2b}(u)|^p \leq \left(\nicefrac{(882 \cdot C'\cdot C^3 \cdot \log^4 n)}{\eps^8}\right)^p  \cdot \opt_p^p. \label{eqn: s2-pivot}
\end{equation}

\paragraph*{Bounding $S_3$.}
It remains to bound the cost of the negative edges.
 Fix $uv \in S_3(u)$. Then the only way $u$ and $v$ can be clustered together is if $u$ and $v$ are clustered by the same pivot $w$, where $w \in N_u^+ \cap N_v^+$ and $\tilde{d}_{uw}, \tilde{d}_{vw} \leq c \cdot r$. But then $1-\tilde{d}_{uv} \geq 1- 2\delta c r$, so we have 
    \begin{align}
        S_3 = \sum_{u \in V_0'}|S_3(u)|^p \leq \Big (\frac{1}{1-2\delta cr} \Big )^p\cdot \sum_{u \in V_0} \Bigg (\sum_{v \in N_u^- \cap V_0} (1-\tilde{d}_{uv}) \Bigg )^p\leq 
     \left ( \nicefrac{(296 \cdot C \cdot \log n)}{\eps^2} \right )^p  \cdot \opt_p^p, \label{eqn: s3-pivot}
    \end{align}
where the last equality is by Lemma \ref{lem: whp-neg-frac-cost} and subbing the values for $\delta,c,r$ chosen in Section \ref{sec:prelim}.

    \medskip

Continuing from line (\ref{eqn: combine-pivot-V_0'}), and substituting in from lines (\ref{eqn: s1-pivot}), (\ref{eqn: s2-pivot}), and (\ref{eqn: s3-pivot}), we conclude the proof
\begin{align*}
    \sum_{u \in V_0'} 
    \big |\textsf{Pivot}(u)|^p \leq  \left(\nicefrac{(5728 \cdot C'\cdot C^3 \cdot \log^4 n)}{\eps^8}\right)^p  \cdot \opt_p^p.
\end{align*}
\end{proof}

\subsubsection{Disagreements between $G[\widebar{V_0}]$ and $G[V_0']$}

The only disagreements occurring on edges going between $V_0'$ and $\widebar{V_0}$ are from positive edges. 
We bound the cost of disagreements $uv$ incident to $u \in \widebar{V_0}$, for $v \in V_0 \cap N_u^+$, in Lemma \ref{lem: type-t-lp}, then
we bound the cost of disagreements $uv$ incident to $u \in V_0$, for $v \in \widebar{V_0} \cap N_u^+$, in Lemma \ref{lem: type-t-lp-backward}.
Since $V_0' \subseteq V_0$, these lemmas immediately bound the cost of disagreements between $G[V_0']$ and $G[\widebar{V_0}]$.

\subsection{Cost between the Pre-clustering phase and the Pivot phase}\label{sec: between-phases}

It remains to bound the cost of edges that go between the Pre-clustering and Pivot phases. The disagreements $uv$ incident to $u$ can take several forms: 
 $u$ is pre-clustered and $v \in V_0'$,
  $u \in V_0'$ and $v$ is pre-clustered,
    $u \in \widebar{V_0}$ and $v$ is pre-clustered,
    $u$ is pre-clustered and $v \in \widebar{V_0}$.
We discuss each disagreement type in order of the above list. 

For $u$ pre-clustered and $v \in V_0'$, both $u$ and $v$ are in $V_0$, but $u \succ v$; recall we use the notation $u \succ v$ to mean $u$ is clustered before $v$, or in other words $u$ is either pre-clustered to a higher ordered center than $v$, or $u$ is pre-clustered and $v$ is not. These disagreements are already accounted for in Lemma \ref{lem: type1b-lp-backwards}.

Similarly, $u \in V_0'$ and $v$ pre-clustered, both nodes are in $V_0$ again. Though this time, $v \succ u$. These disagreements are already accounted for in Lemma \ref{lem: type1b-lp}.

The cost of the next type of disagreement, when $u \in \widebar{V_0}$ and $v$ is pre-clustered, will be bounded in Lemma \ref{lem: type-t-lp}. Then the cost of disagreements where $u$ is pre-clustered (thus $u \in V_0$) and $v \in \widebar{V_0}$ will be bounded in Lemma \ref{lem: type-t-lp-backward}.

\begin{lemma}\label{lem: type-t-lp}
Condition on the good event $B^c$ and fix $1 \leq p < \infty$. The $\ell_p$-cost for $u \in \widebar{V_0}$ of edges $uv$ in disagreement with respect to $\mathcal{C}_{\textsf{ALG}}$, for $v \in V_0$,  is bounded by  
\[\sum_{u \in \widebar{V_0}}|V_0 \cap N_u^+|^p \leq O \left (\left (\nicefrac{1}{\varepsilon^{4}} \cdot \log^2 n\right )^p \right )\cdot \textsf{OPT}_p^p.\]
As two consequences, we have that
$\sum_{u \in \widebar{V_0}} | \{v \in N_u^+ \mid v \text{ pre-clustered} \}|^p \leq O \left (\left (\nicefrac{1}{\varepsilon^{4}} \cdot \log^2 n\right )^p \right )\cdot \textsf{OPT}_p^p $ and $\sum_{u \in \widebar{V_0}} | \{v \in N_u^+ \cap V_0' \}|^p \leq O \left (\left (\nicefrac{1}{\varepsilon^{4}} \cdot \log^2 n\right )^p \right )\cdot \textsf{OPT}_p^p $.
\end{lemma}

\begin{proof}[Proof of Lemma \ref{lem: type-t-lp}]

It suffices to prove the first bound, since if $v$ is pre-clustered then $v \in V_0$, and if $v \in V_0'$ then $v \in V_0$. 

Define $D^+(u)$ to be the fractional cost of the positive edges incident to $u$ with respect to the (actual) correlation metric $d$, that is, $D^+(u) := \sum_{v \in N_u^+} d_{uv}$. We partition the sum based on how large $D^+(u)$ is, and see that
\begin{align}
    \sum_{u \in \widebar{V_0}}|V_0 \cap N_u^+|^p \leq \sum_{\substack{u \in \widebar{V_0}:~ D^+(u) > \frac{\varepsilon^2}{2C \log n}}} |N_u^+|^p  + \sum_{\substack{u \in \widebar{V_0}:~ D^+(u) < \frac{\varepsilon^2}{2C\log n}}} \Bigg ( \sum_{v \in N_u^+ \cap V_0} 1 \Bigg )^p. \label{eqn:eligible-ineligible}
\end{align}

Bounding the first term of Equation (\ref{eqn:eligible-ineligible}) is straightforward. Since we condition on the good event $B^c$, we know that $u \in \widebar{V_0}$ implies $|N_u^+| \leq C \cdot \log n / \varepsilon^2$. Recalling that $D^* = (D^*(u))_{u \in V}$ is the fractional cost vector for the adjusted correlation metric, the following holds:

\begin{align*}
\sum_{\substack{u \in \widebar{V_0}:~ D^+(u) > \nicefrac{\varepsilon^2}{2C \log n}}} |N_u^+|^p  &\leq  \left (\nicefrac{2C  \log n}{\varepsilon^2} \right )^p \cdot \sum_{u \in \widebar{V_0}} |N_u^+|^p \cdot (D^+(u))^p  
\leq \left (\nicefrac{2C  \log n}{\varepsilon^2} \right )^{2p} \cdot \sum_{u \in \widebar{V_0}} (D^+(u))^p \\
&\leq \left (\nicefrac{2C  \log n}{\varepsilon^2} \right )^{2p} \cdot \sum_{u \in \widebar{V_0}} (D^*(u))^p  \leq \left (\nicefrac{(4C^2 \cdot M \cdot \log^2 n)}{\varepsilon^{4}}\right )^p \cdot \textsf{OPT}_p^p.
\end{align*}
In the first inequality on the last line, we have used the fact that $d_{uv} \leq d^*_{uv}$ for all $uv \in E^+$, and the last inequality uses
the bound on $||D^*||_p$  from Lemma \ref{thm: all-norms-offline}, for $M$ the constant in Lemma \ref{thm: all-norms-offline}.

\medskip

Now consider the second term in Equation (\ref{eqn:eligible-ineligible}). We show this term is $0$ since we conditioned on $B^c$.  

Assume for sake of deriving a contradiction there exists $u \in \widebar{V_0}$ with $D^+(u) \leq \eps^2 / (2C \log n)$ whose positive neighborhood does \emph{not} form a perfect clique (i.e., $u$ is incident to at least one bad triangle).  
We know $u$ has one neighbor $v \in N^+_u$, where $ N^+_u \neq  N^+_v$.  Then either $N_v^+ \subset N_u^+$, in which case 
\[d_{uv} = 1 - \frac{|N^+_u \cap N^+_v|}{|N^+_u \cup N^+_v|} \geq 1 - \frac{|N^+_u |-1}{|N^+_u| } \geq 1 -  \frac{\frac{C \log n}{\eps^2} -1}{\frac{C \log n}{\eps^2}} \geq \frac{\eps^2}{2C \log n},\]
or $|N_v^+ \setminus N_u^+|>1$, in which case,
\[d_{uv} = 1 - \frac{|N^+_u \cap N^+_v|}{|N^+_u \cup N^+_v|} \geq 1 - \frac{|N^+_u |}{|N^+_u| +1} \geq 1 -  \frac{\frac{C \log n}{\eps^2} }{\frac{C \log n}{\eps^2} +1} \geq \frac{\eps^2}{2C \log n}.\]
Both equations use that  $|N^+_u| < \frac{C \log n}{\eps^2}$, which must be the case since $u \in \widebar{V_0}$ and we conditioned on the good event. So no such $v$ can exist (regardless of whether or not it is in $N_u^+ \cap V_0$), we as have contradicted the fact that $D^+(u) \leq \eps^2 / (2C \log n).$

Therefore, we we may assume $u$ is part of a perfect clique.  So for $w \in N_v^+$ and $v \in N_u^+$, we have $w \in N_u^+$. If this is the case, then $v \not \in V_0$ for each $v \in N_u^+$. This is because if $N_v^+ \cap S_d \neq \emptyset$, then $N_u^+ \cap S_d \neq \emptyset$, since $N_u^+ = N_v^+$. Thus, none of $u$'s positive neighbors $v$ will be in $V_0$, and $u$ will not contribute to the second sum in Equation (\ref{eqn:eligible-ineligible}).

\end{proof}

\begin{lemma}\label{lem: type-t-lp-backward}
Condition on the good event $B^c$ and fix $1 \leq p < \infty$. The $\ell_p$-cost for $u \in V_0$ of edges $uv$ in disagreement with respect to $\mathcal{C}_{\textsf{ALG}}$, for $v \in \widebar{V_0}$,  is bounded by  
 \[\sum_{u  \in V_0}|\widebar{V_0} \cap N_u^+|^p \leq O \left (\left( \nicefrac{1}{\eps^4} \cdot \log^2 n\right )^p \right )\cdot \textsf{OPT}_p^p .\]
    As two consequences, we have that
$\sum_{u \text{ pre-clustered}} | \{v \in N_u^+ \cap \widebar{V_0} \}|^p \leq O \left (\left( \nicefrac{1}{\eps^4} \cdot \log^2 n\right )^p \right )\cdot \textsf{OPT}_p^p$ and $\sum_{u \in V_0'} | \{v \in N_u^+ \cap \widebar{V_0} \}|^p \leq O \left (\left( \nicefrac{1}{\eps^4} \cdot \log^2 n\right )^p \right )\cdot \textsf{OPT}_p^p $.
\end{lemma}
\begin{proof}[Proof of Lemma \ref{lem: type-t-lp-backward}]

It suffices to prove the first bound, since if $u$ is pre-clustered then $u \in V_0$, and if $u \in V_0'$ then $u \in V_0$. 

Recall $D^+(u)$ is the fractional cost of the positive edges incident to $u$ with respect to the (actual) correlation metric $d$, that is, $D^+(u) := \sum_{v \in N_u^+} d_{uv}$. 
Since $u \in V_0$ and we conditioned on the good event, we know that $|N_u^+| \geq C \cdot \log n / \eps^2$, but we further partition $u \in V_0$ based on whether $|N_u^+| \geq 2C \cdot \log n / \eps^2$, and on how large $D^+(u)$ is
\begin{align}
    \sum_{u \in V_0}|\widebar{V_0} \cap N_u^+|^p &\leq \underbrace{\sum_{\substack{u \in V_0:~ |N^+_u| \geq \frac{2C \cdot \log n}{\eps^2}}} |N_u^+|^p }_{S_1} + \notag\\
    &\underbrace{\sum_{\overset{u \in V_0:~ |N^+_u| < \frac{2C \cdot \log n}{\eps^2},}{D^+(u) > \frac{\varepsilon^2}{4C \log n}}}  \big | N_u^+ \cap \widebar{V_0}\big |^p}_{S_2}+ \underbrace{\sum_{\overset{u \in V_0:~ |N^+_u| < \frac{2C\cdot \log n}{\eps^2},}{D^+(u) \leq \frac{\varepsilon^2}{4C \log n}}} \big | N_u^+ \cap \widebar{V_0}\big |^p}_{S_3}. \label{eqn:part-between-backward}
\end{align}

\paragraph*{Bounding $S_1$.}  
Since we conditioned on the good event, for $v \in \widebar{V_0}$ we have $|N_v^+| < C \log n / \eps^2$.
Therefore we find that $u \in V_0$ and $v \in \widebar{V_0} \cap N_u^+$ are quite far apart, with 
\[d_{uv}  = 1-\frac{|N^+_u\cap N^+_v|}{|N^+_u\cup N^+_v|} \geq 1 - \frac{\frac{C}{\epsilon^2 }\log n}{\frac{2C}{\epsilon^2 }\log n} \geq \frac{1}{2},\]
where we have used the fact that each $u$ included in the sum for $S_1$ has $|N_u^+| \geq 2 C \log n /\eps^2$.
We can then charge the disagreement directly to $d_{uv}$, since it is sufficiently large:
 \begin{align*}
\sum_{u  \in V_0:~ |N^+_u| \geq \frac{2C}{\eps^2\log n}}|\widebar{V_0} \cap N_u^+|^p  &=\sum_{u  \in V_0:~ |N^+_u| \geq \frac{2C}{\eps^2\log n}} \Bigg (\sum_{v \in \widebar{V_0} \cap N_u^+} 1 \Bigg)^p\\ 
&\leq 2^p \cdot \sum_{u  \in V_0} \big (D^+(u)\big )^p 
\leq 2^p \cdot \sum_{u  \in V_0} \big (D^*(u)\big )^p 
\leq (2 \cdot M)^p  \cdot \textsf{OPT}_p^p
\end{align*}
In the last line, we used the fact that $d_{uv} \leq d^*_{uv}$ for all $uv \in E^+$, and then use the bound  on $||D^*||_p$ from Lemma \ref{thm: all-norms-offline} ($M$ is the constant from Lemma \ref{thm: all-norms-offline}).

\paragraph*{Bounding $S_2$.} 
Bounding $S_2$ is almost identical to bounding the first sum in Equation (\ref{eqn:eligible-ineligible}). This is because $u$ being ineligible implies $|N_u^+|$ is small by $B^c$, and here the $u$ in the sum have small $|N_u^+|$ too.

\begin{align*}
\sum_{\overset{u \in V_0:~ |N^+_u| < \frac{2C}{\eps^2\log n},}{D^+(u) > \nicefrac{\varepsilon^2}{4C \log n}}} \big | N_u^+ \cap \widebar{V_0}\big |^p &\leq  \left (\nicefrac{4C  \log n}{\varepsilon^2} \right )^p \cdot \sum_{u  \in V_0} |N_u^+|^p \cdot (D^+(u))^p  \\
\vspace{-3mm}
&\leq \left (\nicefrac{4C  \log n}{\varepsilon^2} \right )^{2p} \cdot \sum_{u  \in V_0} (D^+(u))^p 
 \leq \left (\nicefrac{(16 \cdot C^2 \cdot M \cdot \log^2 n)}{\varepsilon^{4}}\right )^p \cdot \textsf{OPT}_p^p,
\end{align*}
where in the last inequality, we used that $d_{uv} \leq d^*_{uv}$ for all $uv \in E^+$, then apply
Lemma \ref{thm: all-norms-offline}.

\medskip

\paragraph*{Bounding $S_3$}  We show  $S_3=0$, since we conditioned on $B^c$.  

The proof follows exactly as that for bounding the second sum in Equation (\ref{eqn:eligible-ineligible}), just replacing $C \log n / \eps^2$ with $2C \log n / \eps^2$. 

\medskip
Summing together $S_1$, $S_2$ and $S_3$, we find that 
\[\sum_{u \in V_0} |\widebar{V_0} \cap N_u^+|^p\leq \left( \nicefrac{(18 \cdot C^2 \cdot M \cdot \log^2n)}{\eps^4}\right )^p \cdot \textsf{OPT}_p^p .\]
\end{proof}

\subsection{Proof of item \ref{item: thm-allp} for Theorem \ref{thm:main-all}}
We are ready to combine the results proven so far in this section.

\begin{proof}[Proof of item \ref{item: thm-allp} for Theorem \ref{thm:main-all}]
Let  $\mathcal{C}_{\textsf{ALG}}$ be the clustering output by Algorithm \ref{alg: main-alg}, and let $\text{cost}_p(\mathcal{C}_{\textsf{ALG}})$ be the $\ell_p$-norm of the disagreement vector of $\mathcal{C}_{\textsf{ALG}}$.
We further partition the edges in disagreement based on whether they are positive or negative, which phase in Algorithm \ref{alg: main-alg} they are clustered in, and (if at least one endpoint of an edge is pre-clustered) whether or not $u \succ v$. These cases are exhaustive; see Figure \ref{fig:summary-ellp}. Combining the terms from Lemmas \ref{lem: type1b-lp}, \ref{lem: type1b-lp-backwards}, \ref{lem: negative-lp}, \ref{lem: pivot-cost-lp-v0},\ref{lem: pivot-cost-lp-v''}, 
\ref{lem: type-t-lp}, and 
\ref{lem: type-t-lp-backward}, and then applying Jensen's inequality and taking the $p^{\text{th}}$ root,
we see that with high probability (as we recall the good event $B^c$ occurs with high probability)
\begin{align*}
||y_{\mathcal{C}_{\textsf{ALG}}}||_p  \leq O \left (\nicefrac{1}{\eps^8} \cdot \log^4 n \right) \cdot \opt_p.
\end{align*}
\end{proof}

\section{Conclusion}
We develop an algorithm for online correlation clustering which, given a sample of $\eps$-fraction of the nodes from the underlying instance,
returns a clustering that is simultaneously $O(\nicefrac{1}{\eps^6})$-competitive for the $\ell_1$-norm objective in expectation and $O(\nicefrac{\log n}{\eps^6})$-competitive for the $\ell_\infty$-norm objective with high probability. 
This is the first positive result for the $\ell_\infty$-norm in the online setting. 
We also prove lower bounds that match our upper bounds up to constants and powers of $\nicefrac{1}{\eps}$ for either norm. Finally, we show that our algorithm is also $O(\nicefrac{\log^4n}{\eps^8})$-competitive for each finite $\ell_p$-norm with high probability. Thus, we successfully translate the all-norms result of \cite{davies2023one} to the online setting. 

Our work highlights two key insights. First, it demonstrates the robustness of the adjusted correlation metric: even an estimated version suffices to guide near-optimal decisions in the AOS model. Second, it identifies structural properties that make problems amenable to this online model. Specifically, the ability to estimate key quantities from a small but uniformly sampled subset of the input is crucial for solving problems in the AOS model.  

Overall, our results suggest that the AOS model is a promising framework for problems where limited but well-distributed information allows for effective decision-making. In particular, $\ell_\infty$-norm clustering is an example of a problem when the AOS model is much stronger than the popular RO model. We remark that the model is still relatively new, and we believe the techniques from this paper can be of use to understand a wider range of problems in this setting. For instance, our idea of leveraging different subsamples independently to help mitigate correlation effects across estimating different quantities may be useful. 
It is of further interest to determine which problems with strong lower bounds on the competitive ratio in the strictly online setting and/or the random-order (RO) model admit small competitive ratios in the AOS model.

\printbibliography

\appendix

\section{Omitted Proofs and Constructions}\label{app: all-omits}

This section contains proofs and constructions that were omitted from the main body.

\subsection{Subsampling to simulate several independent samples }\label{app:omit-prelim}

In Lemma \ref{lem: simulate-sample}, we show how to construct four independent subsamples $S_p, S_d, S_b, S_r$ from the given sample $S$. The idea is to independently add each $v \in S$ to some subset of $\{S_p, S_d, S_b, S_r\}$, where the probability for each possible subset of subsamples is recursively defined so as to ensure the required independence properties.

\begin{lemma} \label{lem: simulate-sample}
For a fixed $0 < \eps < 1$, 
let $S \subseteq V$ independently contain each element of $V$ with probability $\eps$. From $S$, we construct 4 samples $S_p, S_d, S_b$, and $S_r$ such that:
\begin{enumerate}
\item  The probability of an element being in any one of the 4 samples is  $\eps^2/2$, i.e., for any $S_i \in \{S_p,S_d,S_b,S_r\}$ and any $v \in V$, we have that $\mathbb{P}(v \in S_i) = \eps^2/2$.
\item (Independence between elements) For any $\mathcal{S} \subseteq \{S_p, S_d, S_b, S_r\}$, the events $\{v \in \bigcap_{S_i \in \mathcal{S}} S_i \}_{v \in V}$ are mutually independent. That is, $\mathbb{P}(\bigcap_{v \in V'} \{v \in \bigcap_{S_i \in \mathcal{S}} S_i \}) = \prod_{v \in V'} \mathbb{P}(v \in \bigcap_{S_i 
 \in \mathcal{S}} S_i)$,  for any $V' \subseteq V$.
\item 
(Independence between samples) For any $v \in S$, and for any $\mathcal{S} \subseteq \{S_p,S_d,S_b,S_r\}$, the events $\{v \in S_i \}_{S_i \in \mathcal{S}}$ are mutually independent. That is, 
 $\mathbb{P}\left( \bigcap_{S_i \in \mathcal{S}} \{v \in S_i\} \right) = \prod_{S_i \in \mathcal{S}}\mathbb{P}(  v \in S_i).$ 
=\end{enumerate}
\end{lemma}

\begin{proof}[Proof of Lemma \ref{lem: simulate-sample}]
    We form the samples with the following procedure. 
    Fix $x>0$ 
    to be specified as a function of $\varepsilon$ later; it may be helpful to know that we will eventually show that $x = \mathbb{P}[v \in S_d \mid v \in S]$, where one could replace $S_d$ here with $S_b,S_p,$ or $S_r$.
    Consider each element $v \in S$ independently:
\begin{itemize}
\item Add $v$ to only $S_p$ (or $S_d$, $S_b$, or $S_r$) and to no other sample with probability $x-3\varepsilon x^2+3 \varepsilon^2 x^3-\varepsilon^3 x^4$. Thus the probability that $v$ is in exactly one sample is $4(x-3\varepsilon x^2+3 \varepsilon^2 x^3-\varepsilon^3 x^4)$.
\item Add $v$ to exactly $S_p$ and $S_d$ (or $S_p$ and $S_b$, $S_p$ and $S_r$, $S_d$ and $S_b$, $S_d$ and $S_r$, or $S_b$ and $S_r$) with probability $\varepsilon x^2-2 \varepsilon^2 x^3 + \varepsilon^3 x^4$. Thus the probability that $v$ is in exactly two samples is $6(\varepsilon x^2-2 \varepsilon^2 x^3 + \varepsilon^3 x^4)$.
\item Add $v$ to exactly $S_p$ and $S_d$ and $S_r$ (or $S_p$ and $S_d$ and $S_b$, $S_p$ and $S_r$ and $S_b$, or $S_d$ and $S_r$ and $S_b$) with probability $\varepsilon^2 x^3 - \varepsilon^3 x^4$. Thus the probability that $v$ is in exactly three samples is $4(\varepsilon^2 x^3 - \varepsilon^3 x^4)$.
\item Add $v$ to all 4 samples with probability $\varepsilon^3 x^4$. 
\item Add $v$ to none of the samples with probability 
\begin{align*}
    &1-\varepsilon^3 x^4-4(\varepsilon^2 x^3 - \varepsilon^3 x^4)-6(\varepsilon x^2-2 \varepsilon^2 x^3 + \varepsilon^3 x^4)-4(x-3\varepsilon x^2+3 \varepsilon^2 x^3-\varepsilon^3 x^4)\\
    &=1-4x+6 \varepsilon x^2-4 \varepsilon^2x^3+ \varepsilon^3 x^4.
\end{align*}
\end{itemize}

The above is a probability distribution so long as $x$ is chosen small enough relative to $\varepsilon$ so that all probabilities above are between 0 and 1. 
For instance, the choice of $x=\varepsilon/2$ works.
Further, since each $v$ is sampled into $S$ independently and is then considered independently in deciding where to allocate it among the subsamples in $ \{S_p, S_d, S_b, S_r\}$, property (2) in the lemma statement holds by construction. 

For a fixed $v \in S$, the probability that $v \in S_p \cap S_d \cap S_b$ is the probability that $v$ is in $S_p, S_d, S_b$ and not $S_r$ plus the probability that $v$ is in all 4 subsamples: 
\begin{align}\label{eq: triples}
\mathbb{P} [v \in S_p \cap S_d \cap S_b \mid v \in S] = \varepsilon^2 x^3 - \varepsilon^3 x^4 + \varepsilon^3 x^4 = \varepsilon^2 x^3.
\end{align}
This is symmetric for all triples of subsamples.

Then, for a fixed $v \in S$, the probability that $v \in S_p \cap S_d$ is the probability that $v$ is in $S_p$ and $ S_d$ but neither $S_b$ nor $S_r$, plus the probability that $v$ is in $S_p, S_d$ and exactly one of $S_b$ or $S_r$, plus the  probability that $v$ is in all 4 subsamples:
\begin{align}
\label{eq: pairs}
\mathbb{P} [v \in S_p \cap S_d \mid v \in S] = \varepsilon x^2-2 \varepsilon^2 x^3 +\varepsilon^3x^4 + 2(\varepsilon^2x^3-\varepsilon^3x^4) + \varepsilon^3x^4 = 
\varepsilon x^2.
\end{align}
This is symmetric for all pairs of subsamples.

Then, for a fixed $v \in S$, the probability that $v \in S_p$ is the probability that $v$ is in $S_p$ but none of the other subsamples, plus the probability that $v$ is in $S_p$ and exactly 1 of the other subsamples, plus the  probability that $v$ is in $S_p$ and exactly 2 of the other subsamples, plus the  probability that
$v$ is in all 4 subsamples:
\begin{align*}
    \mathbb{P} [v \in S_p \mid v \in S] &= x - 3 \varepsilon x^2+3 \varepsilon^2 x^3 - \varepsilon^3 x^4 + 3(\varepsilon x^2 - 2 \varepsilon^2 x^3+\varepsilon^3 x^4) + 3(\varepsilon^2 x^3 - \varepsilon^3 x^4) + \varepsilon^3 x^4 
    =  x. 
\end{align*} 
In particular, 
\begin{align}\label{eq: single}
    \mathbb{P}[v \in S_p] = \varepsilon x,
\end{align}
and this is symmetric for $S_d,S_b,$ and $S_r.$
We have proved that property (1) in the lemma statement is true for our choice of $x = \varepsilon/2$.

It remains to prove independence between subsamples (property (3) in the lemma statement). We begin with pairs of subsamples.
Using Equations (\ref{eq: pairs}) and (\ref{eq: single}), 
\[\mathbb{P}[v \in S_p \cap S_d] = \varepsilon^2 x^2 = \mathbb{P} [v \in S_p ]\cdot \mathbb{P} [v \in S_d ].\]
For triples, we use Equations (\ref{eq: triples}) and (\ref{eq: single}),
\[\mathbb{P}[v \in S_p \cap S_d \cap S_b] = \varepsilon^3 x^3 = \mathbb{P} [v \in S_p ]\cdot \mathbb{P} [v \in S_d ]\cdot \mathbb{P} [v \in S_b].\]
Lastly, we use the fact that the probability $v \in S$ is added to all 4 subsamples is $\varepsilon^3 x^4$ together with Equation (\ref{eq: single}),
\[\mathbb{P}[v \in S_p \cap S_d \cap S_b \cap S_r] = \varepsilon^4 x^4 = \mathbb{P} [v \in S_p ]\cdot \mathbb{P} [v \in S_d ]\cdot \mathbb{P} [v \in S_b]\cdot \mathbb{P} [v \in S_r].\]
\end{proof}

The independence properties satisfied by our construction yield the following corollary, which we apply heavily in the analysis.

\begin{corollary} \label{cor: subsample-independence}
Let $\mathcal{S} \subseteq \{S_p,S_d,S_b,S_r\}$. Let $f$ be any function that depends on the subsamples in $\mathcal{S}$ and let $\mathcal{S}' \subseteq \mathcal{S}$. Let $\{s_i'\}_{i : S_i \in \mathcal{S}}$ be any collection of subsets of $V$, that is, a collection of possible realizations of $\{S_i\}_{i \in \mathcal{S}'}$. Define $g(\{s_i'\}_{i : S_i \in \mathcal{S}'}) := \mathbb{E}\left[ f(\{s_i'\}_{i : S_i \in \mathcal{S'}}, \{S_i\}_{i : S_i \in \mathcal{S} \setminus \mathcal{S'}}\} )\right]$, that is, the expectation of $f$ when the subsamples in $\mathcal{S}'$ are fixed. Then 
\[\mathbb{E}\left[f(\mathcal{S}) \mid \mathcal{S}' \right] = g(\mathcal{S}').\]
In other words, to compute the conditional expectation, we can compute an unconditional expectation where we only integrate over the subsamples that are not part of the conditioning.
\end{corollary}

\begin{proof}[Proof of Corollary \ref{cor: subsample-independence}]
We use the following claim.  
\begin{claim} \label{clm: full-independence}
 Let $\mathcal{S}_1, \mathcal{S}_2 \subseteq \{S_p,S_d,S_b,S_r\}$ with $\mathcal{S}_1 \cap \mathcal{S}_2 = \emptyset$. Let $\{s_i\}_{i: S_i \in \mathcal{S}_1 \cup \mathcal{S}_2}$ be a collection of subsets of $V$. Then 
 \[\mathbb{P}(\{S_i = s_i\}_{i: S_i \in \mathcal{S}_1 \cup \mathcal{S}_2}) = \mathbb{P}(\{S_i = s_i\}_{i: S_i \in \mathcal{S}_1}) \cdot \mathbb{P}(\{S_i = s_i\}_{i: S_i \in \mathcal{S}_2}) \]
\end{claim}
The claim is a straightforward consequence of properties (2) and (3) in Lemma \ref{lem: simulate-sample}.  Having established the claim, the corollary then follows from the following well-known fact from probability theory.

\begin{fact} \label{fct: conditional-integrating}
Let $X,Y$ be \emph{independent, discrete} random vectors, and let $f$ be a function of $X$ and $Y$. Define $g(x) := \mathbb{E}[f(x,Y)]$ for every $x$ in the support of $X$. Then $\mathbb{E}[f(X,Y) \mid X] = g(X)$. 
\end{fact}

\end{proof}

\subsection{Proof that $\bar{d}$ and $\tilde{d}$ are semi-metrics}

Throughout this section, for  $z \in V$ and $S_*$ a chosen subsample, define the random variable $X_z$ to be 1 if $z \in S_*$ and 0 otherwise. Most often, the analysis will choose $S_*$ to be $S_d$.

Recall that 
Lemma \ref{lem: tri-inequality} states that $\bar{d}$ and $\tilde{d}$ are semi-metric. We prove this lemma next.

\trineq*

\begin{proof}[Proof of Lemma \ref{lem: tri-inequality}]
First, we prove that the
triangle inequality holds for $\bar{d}$, i.e., for vertices $u,v,w \in V$, $\bar{d}_{uv} \leq \bar{d}_{uw}+\bar{d}_{wv}$.
We already know that the correlation metric satisfies the triangle inequality by Lemma 2 in \cite{DMN23}.
When the sample $S_d$ intersects all of $N_u^+,N_v^+,N_w^+$, then
$\bar{d}$ is equal to the correlation metric on the subgraph induced by the sample.
To see this formally, recall $S_d$ is the distance sample,
and define the random variable $X_z$ to be 1 if $z \in S_d$ and 0 otherwise.
For $G[S_d]$ the subgraph induced on the vertices in $S_d$, we observe that
$$
\bar{d}_{uv} = 1-\frac{\sum_{z \in N_u^+ \cap N_v^+}X_z}{\sum_{z \in N_u^+ \cup N_v^+}X_z} 
=1-\frac{|(N_u^+ \cap N_v^+) \cap S_d|}{|(N_u^+ \cup N_v^+) \cap S_d|}.
$$
This is almost exactly the correlation metric in $S_d$, except that $u$ and $v$ may no longer be in the subgraph $G[S_d]$.
However, this subtle difference does not matter,
as the vertices in $S_d$ can still be partitioned based on their membership to $N_u^+ \cap S_d$ and $N_u^- \cap S_d$, or further partitioned based on their membership to $(N_u^+ \cap N_v^+)\cap S_d$, $(N_u^+ \cap N_v^-)\cap S_d$,
$(N_u^- \cap N_v^+)\cap S_d,$ or $(N_u^- \cap N_v^-)\cap S_d$, allowing the exact same proof as of Lemma 2 in \cite{DMN23} to go through. 

Next, we observe that when at least one of $N_u^+,N_v^+,N_w^+$ do not intersect $S_d$, the triangle inequality still holds. The only case to check is when $N_w^+ \cap S_d = \emptyset$, but $N_u^+ \cap S_d \neq \emptyset$ and $N_v^+ \cap S_d \neq \emptyset$.
Here, we see that
$\bar{d}_{uw} + \bar{d}_{wv} \geq 1$ because $|(N_u^+ \cap N_w^+) \cap S_d|$ and $|(N_v^+ \cap N_w^+) \cap S_d|$ are 0.

The proof that $\tilde{d}$ is a 10/7 semi-metric is the same as the proof of Lemma 1 in \cite{davies2023one}.
There are three steps to computing $\tilde{d}$. The first step is to take $\tilde{d} = \bar{d}$ (for which we know the triangle inequality holds).
Then in step 2, we round up negative edges with $\bar{d}\geq 7/10$, thus only gaining a factor of at most $10/7$ in an approximate triangle inequality. In the third and final step, some vertices are put in their own cluster. Here, if step 3 results in the left hand side of the approximate triangle inequality being changed to 1, then so is at least one term on the right hand side; so we do not gain any additional factor in the approximate triangle inequality from step 3.
\end{proof}

\subsection{The good event $B^c$} \label{sec: whp-good-events}

For the $\ell_p$-norm analysis with $p \in (1, \infty]$, we will condition on a ``good'' event occurring with high probability. The good event will be the complement of the following bad event, $B$.

\begin{definition}\label{def: good-event}
    Let the \emph{bad event} $B$ be the union of the following events, which depend on the randomness of the subsamples: 
    \begin{enumerate}
    \item $|N_u^+ \cap S_d| < \eps^2/4 \cdot |N_u^+|$ for some $u$ with $|N_u^+| \geq C \cdot \log n / \eps^2$. 
    \item For $t = \frac{r}{2 \delta}$,  $\ball_{\tilde{d}}^{S_p}(u,t) = \emptyset$ for some $u$ with  $|\ball_{\tilde{d}}(u,t)| \geq C' \cdot \log_{1/(1-\varepsilon^2/2)} n$. 
     \item   $\ball_{\tilde{d}}^{S_p}(u,c \cdot r) = \emptyset$ for some $u$ with  $|\ball_{\tilde{d}}(u,c \cdot r)| \geq C' \cdot \log_{1/(1-\varepsilon^2/2)} n$. 
    \item $|\ball_{\tilde{d}}^{S_b}(w,r)| \leq \nicefrac{\eps^2}{4} \cdot |\ball_{\tilde{d}}(w,r)|$ for some $w$ with $|\ball_{\tilde{d}}(w,r)| \geq 2C \cdot \log n / \eps^2$ 
    \item $|R_1(u) \cap S_r| <\nicefrac{\eps^2}{4} \cdot |R_1(u)|$ for some $u$ with $|R_1(u)| \geq C \cdot \log n / \varepsilon^2$. 
    \item $|N_u^+ \cap S_p| < \nicefrac{\eps^2}{4} \cdot |N_u^+|$ for some $u$ with $|N_u^+| \geq C \cdot \log n / \eps^2$. 
    \item $\bar{d}_{uv} > \frac{(1+C)}{3 \varepsilon^2} \cdot \log n\cdot d_{uv}$ or $1-\bar{d}_{uv} > \frac{(1+C)}{3 \varepsilon^2} \cdot \log n \cdot (1-d_{uv})$
    for some $u,v$ with $|N_u^+ \cup N_v^+| \geq C \cdot \frac{\log n}{\varepsilon^2}$.  
\end{enumerate}
Define the complement $B^c$ to be the \emph{good event}.
\end{definition}

To analyze the probability of the good event, we will use the following well-known Chernoff-Hoeffding bound. 

\begin{theorem}[Chernoff-Hoeffding] \label{thm: chernoff}
  Let $X = X_1 + \cdots + X_m$ where $\{X_1, \dots, X_m\}$ is a set of i.i.d. indicator random variables ($X_i \in \{0,1\}$ for $i \in [m]$). Define $\mu = \mathbb{E}[X]$. Then the following tail bounds hold:
\begin{align*}
\mathbb{P}(X \geq (1+\delta)\mu) &\leq e^{-\delta\mu/2} \hspace{0.3cm} \text{for }\delta \geq 2 \\
\mathbb{P}(X \leq (1-\delta)\mu) &\leq e^{-\delta^2 \mu/3} \text{for }0<\delta< 1.
\end{align*}
\end{theorem}

We will show that the good event $B^c$ occurs with high probability in Lemma \ref{lem:good-whp}. For almost every event composing $B$ in Definition \ref{def: good-event}, the fact that it happens with high probability is a consequence of previously proven statements. However, the last item (item 6) in Definition \ref{def: good-event} requires a bit more work. 
Specifically, in Lemma \ref{lem: whp-distances} we show that with high probability the estimated (unadjusted) correlation metric $\bar{d}$ is a good estimate of the (unadjusted) correlation metric\footnote{We note that we do not, however, obtain an analogous statement for the estimate $\tilde{d}$ of the adjusted correlation metric $d^*$. We discuss this more in Appendix \ref{sec: pre-clustering-cost-infinity}.
} $d$ under a technical condition, and thus the last item follows.

\begin{lemma} \label{lem: whp-distances}
Let $\bar{d}$ be the estimated correlation metric. Fix $u,v$ such that $|N_u^+ \cup N_v^+| \geq C \cdot \frac{\log n}{\varepsilon^2}$. Then, each of the following happens with probability at least $1-\frac{3}{n^{C/24}}$:
\[\bar{d}_{uv} 
\leq  \frac{ C+1}{3\varepsilon^2} \cdot \log n \cdot d_{uv} 
 \quad \textrm{and} \quad 1-\bar{d}_{uv} \leq   \frac{ C+1}{3 \varepsilon^2} \cdot \log n \cdot  (1-d_{uv}).\]
In particular, for fixed $u \in V$, if $|N_u^+| \geq C \cdot \frac{\log n}{\varepsilon^2} $, then with probability at least $1-\frac{3}{n^{C/24 - 1}}$,
\[\bar{D}(u) \leq \frac{C+1}{3 \varepsilon^2}\cdot \log n \cdot D(u) \leq \frac{8(C+1)}{3\varepsilon^2} \cdot \log n \cdot \textsf{OPT}_\infty. \]
\end{lemma}

We will use the following proposition, stating that the size of a sufficiently large random subset is well-concentrated, repeatedly. In all of our applications, we will take $U$ to be one of $S_p$, $S_d$, $S_b$, or $S_r$.

\begin{proposition} \label{prop: gen-chernoff}
    Let $T \subseteq V$, and $U$ be  a subset such that each element $v \in V$ is in $U$ independently with probability $p$. Suppose $|T| \geq (C \cdot \log n) / p$. Then, with probability at least $1-\frac{1}{n^{C/12}}$, 
    \[3 \cdot |T| \geq \frac{1}{p} \cdot |T \cap U| \geq \frac{1}{2} \cdot |T|. \]
\end{proposition}
\begin{proof}[Proof of Proposition \ref{prop: gen-chernoff}]
The proof follows from Theorem \ref{thm: chernoff}. 
Note that in what follows, $\mu = |T|p$.
For the upper bound,
\[
\mathbb{P}(|T \cap U| \geq 3|T|p) \leq e^{-2 |T|\cdot p/2} \leq e^{- C \log n} = 1/n^C.\]
For the lower bound,
\[
\mathbb{P}(|T \cap U| \leq |T|p/2) \leq e^{- |T|p/12} \leq e^{- C \log n/12} = \frac{1}{n^{C/12}}.
\]
\end{proof}

Now, we use Proposition \ref{prop: gen-chernoff} to prove Lemma \ref{lem: whp-distances}.
Recall that $X_w$ is defined to be 1 if $w \in S_d$ and 0 otherwise. Recall also that $q(\varepsilon) = \mathbb{P}(v \in S_d) = \varepsilon^2/2$ for any vertex $v$. 

\begin{proof}[Proof of Lemma \ref{lem: whp-distances}]
    We may rewrite Definition \ref{def: est-corr} as 
    \begin{equation} \label{eq: dbar-rvs}
    \bar{d}_{uv} = 1 - \frac{Y_{u,v}^{+,+}}{Y_{u,v}} =  \frac{Y_{u,v}^{+,-} + Y_{u,v}^{-,+}}{Y_{u,v}}
    \end{equation}
    for $Y_{u,v} > 0$, where
    \[Y_{u,v}^{+,-} = \frac{1}{q(\eps)} \cdot \sum_{w \in N_u^+ \cap N_v^-} X_w, \hspace{1cm} Y_{u,v}^{-,+} = \frac{1}{q(\eps)} \cdot \sum_{w \in N_u^- \cap N_v^+} X_w, \hspace{0.5cm}\]
    and
    \[Y_{u,v}^{+,+} = \frac{1}{q(\eps)} \cdot \sum_{w \in N_u^+ \cap N_v^+} X_w, \hspace{1cm} Y_{u,v} = \frac{1}{q(\eps)} \cdot \sum_{w \in N_u^+ \cup N_v^+} X_w. \]

Recall further that $\bar{d}_{uv}$ is defined to be 1 when $Y_{u,v} = 0$. We will show that, under the hypotheses of the proposition, the numerator and the denominator are each well-concentrated.

First we handle the denominator of $\bar{d}_{uv}$ in the right-hand side of (\ref{eq: dbar-rvs}). Applying Proposition \ref{prop: gen-chernoff} and the hypothesis that $|N_u^+ \cup N_v^+| \geq C \cdot \frac{\log n}{\varepsilon^2}$, we have 
\begin{equation} \label{eq: lower-bd-den}
\mathbb{P}\left(Y_{u,v} \leq \frac{1}{2} \cdot |N_u^+ \cup N_v^+| \right) \leq \frac{1}{n^{C/24}}.
\end{equation}
In particular, $Y_{u,v} > 0$ with high probability, meaning that $\bar{d}_{uv}$ is defined as in line (\ref{eq: dbar-rvs}) with high probability.

Next we handle the numerator of $\bar{d}_{uv}$ in the right-hand side of (\ref{eq: dbar-rvs}). Observe that 
\[\mathbb{E}[Y_{u,v}^{+,-}] = |N_u^+ \cap N_v^-| \quad \textrm{and} \quad \mathbb{E}[Y_{u,v}^{-,+}] = |N_u^- \cap N_v^+|.\]

Now we apply Theorem \ref{thm: chernoff} to $Y_{u,v}^{+,-}$ and $Y_{u,v}^{-,+}$. By symmetry, it suffices to bound the former random variable:
\begin{align}
\mathbb{P}\left(Y_{u,v}^{+,-} \geq (1+\delta) \cdot \mathbb{E}[Y_{u,v}^{+,-}] \right) = \mathbb{P}\Big(\sum_{w \in N_u^+ \cap N_v^-} X_w \geq (1+\delta) \cdot \mathbb{E}\big[\sum_{w \in N_u^+ \cap N_v^-} X_w\big] \label{eq: apply-chernoff} \Big).
\end{align}
We may assume without loss of generality that $|N_u^+ \cap N_v^-| \geq 1$
(as otherwise $\bar{d}_{uv}=d_{uv}=0$)
so that $\mu := \mathbb{E}\left[\sum_{w \in N_u^+ \cap N_v^-} X_w\right] \geq q(\varepsilon) = \varepsilon^2/2$. Thus, taking $\delta \geq \frac{C}{6} \cdot \frac{\log n}{\varepsilon^2}$ and applying Theorem \ref{thm: chernoff} to line (\ref{eq: apply-chernoff}), we have that 
\begin{equation} \label{eq: upper-bd-num1}
\mathbb{P}\left(Y_{u,v}^{+,-} \geq (1+\delta) \cdot |N_u^+ \cap N_v^-| \right) \leq 
e^{-\delta \mu/2}
 \leq   e^{-(C/24) \log n} = \frac{1}{n^{C/24}}.
\end{equation}
and by an identical argument, 
\begin{equation} \label{eq: upper-bd-num2}
\mathbb{P}\left(Y_{u,v}^{-,+} \geq (1+\delta) \cdot |N_u^- \cap N_v^+| \right)\leq \frac{1}{n^{C/24}}.
\end{equation}

Finally, combining (\ref{eq: dbar-rvs}), (\ref{eq: lower-bd-den}),  (\ref{eq: upper-bd-num1}), and (\ref{eq: upper-bd-num2}), we have that, with probability at least $1-\frac{3}{n^{C/24}}$, 
\[\bar{d}_{uv} \leq   \frac{C+1}{3 \varepsilon^2} \cdot \log n \cdot \frac{|N_u^+ \cap N_v^-| + |N_u^- \cap N_v^+|}{|N_u^+ \cup N_v^+|},\]
as desired. The proof that, with high probability, $1-\bar{d}_{uv} =  O(\log n / \varepsilon^2) \cdot (1-d_{uv})$ is similar, because $1-\bar{d}_{uv}$ simplifies to $|N_u^+ \cap N_v^+|/|N_u^+ \cup N_v^+|$, and the concentration bound for $Y_{u,v}^{+,+}$ is identical to that of, say, $Y_{u,v}^{+,-}$.

Theorem \ref{thm: corr-metric-cost} implies that with high probability $\bar{D}(u) \leq  \frac{ 8(C+1)}{3 \varepsilon^2} \cdot \log n \cdot \textsf{OPT}_\infty$.
\end{proof}

Now, we are ready to prove the good event occurs with high probability.

\begin{lemma}\label{lem:good-whp}
    The good event $B^c$ in Definition \ref{def: good-event} happens with probability at least \\ $1-\left (\frac{3}{n^{C/24-2}}+\frac{4}{n^{C/12-1}}+\frac{2}{n^{C'-1}}\right )$. 
\end{lemma}

\begin{proof}[Proof of Lemma \ref{lem:good-whp}]
 We will go item by item from Definition \ref{def: good-event} to upper bound the probability of the bad event $B$. 

\noindent \emph{Item 1.} Fix some $u$ with $|N_u^+| \geq \frac{C}{\varepsilon^2} \cdot \log n $. Applying Proposition \ref{prop: gen-chernoff} by choosing $T$ to be $N_u^+$ and $U$ to be $S_d$, with probability at least $1-\frac{1}{n^{C/12}}$
 \[3 \cdot |N_u^+| \geq \frac{2}{\varepsilon^2} \cdot |N_u^+ \cap S_d| \geq \frac12 \cdot |N_u^+|.\]
 Thus $|N_u^+ \cap S_d| < \varepsilon^2/4 \cdot |N_u^+|$ with probability less than $\frac{1}{n^{C/12}}$, and union bounding over all such potential vertices gives an upper bound of  $\frac{1}{n^{C/12-1}}$.

 \noindent \emph{Item 2.} Fix some $u$ with $|\ball_{\tilde{d}}(u,t)| > C' \cdot \log_{1/(1-q(\varepsilon))} n$. Note we consider $\tilde{d}$ fixed here, meaning that some choice of $S_d, S_r$ is fixed here. Then we can use Corollary \ref{cor: subsample-independence} (and in particular the independence of $S_p$ from  $S_d, S_r$) to see that 
 \[\mathbb{P} [S_p \cap \ball_{\tilde{d}}(u,t) = \emptyset \mid S_d, S_r] = (1-q(\varepsilon))^{|\ball_{\tilde{d}}(u,t)|} \leq (1-q(\varepsilon))^{C' \cdot \log_{1/(1-q(\varepsilon))} n} = \frac{1}{n^{C'}}, \]
 where the randomness is over $S_p$.
Taking a union bound over all such possible $u$, we upper bound the probability this happens for any $u$ by $\frac{1}{n^{C'-1}}.$

\noindent \emph{Item 3.} Same as proof of the previous item.

 \noindent \emph{Item 4.} Fix some $w$ with $|\ball_{\tilde{d}}(w,r)| \geq C \cdot \log n / \eps^2$. Note we consider $\tilde{d}$ fixed here, meaning that some choice of $S_d$ and $S_r$ are fixed here.
Since $S_b$ is independent of $S_d$ and $S_r$, as in Corollary \ref{cor: subsample-independence}, we can
apply Proposition \ref{prop: gen-chernoff} by letting $T$  be $|\ball_{\tilde{d}}(w,r)|$ and $U$ to be $S_b$, so that with probability at least $1-\frac{1}{n^{C/12}}$
 \[3 \cdot |\ball_{\tilde{d}}(w,r)| \geq \frac{2}{\varepsilon^2} \cdot |\ball_{\tilde{d}}(w,r) \cap S_b|  \geq \frac12 \cdot |\ball_{\tilde{d}}(w,r)|.\]
Thus  $|\ball_{\tilde{d}}(w,r) \cap S_b| \leq \nicefrac{\eps^2}{4} \cdot |\ball_{\tilde{d}}(w,r)|$ with probability less than $\frac{1}{n^{C/12}}$ for any one such $w$, and union bounding over any such possible $w$ gives an upper bound of $\frac{1}{n^{C/12}-1}$.

 \noindent \emph{Item 5.} Fix some $u$ with $|N_u^- \cap \{v \in V: \bar{d}_{uv} \leq 7/10 \}| \geq C \cdot \log n / q(\varepsilon)$, thus fixing some choice of $S_d$ here. Subsamples are independent by Corollary \ref{cor: subsample-independence}, and in particular $S_r$ and $S_d$ are independent. Thus we can apply Proposition \ref{prop: gen-chernoff} by letting $T$  be $|N_u^- \cap \{v \in V: \bar{d}_{uv} \leq 7/10 \}| $ and $U$ to be $S_r$, so that with probability at least $1-\frac{1}{n^{C/12}}$
  \begin{align*}
      3 \cdot |N_u^- \cap \{v \in V: \bar{d}_{uv} \leq 7/10 \}| &\geq \frac{2}{\varepsilon^2} \cdot |N_u^- \cap \{v \in V: \bar{d}_{uv} \leq 7/10 \}\cap S_r| \\
      &\geq \frac12 \cdot |N_u^- \cap \{v \in V: \bar{d}_{uv} \leq 7/10 \}|.  
      \end{align*}

It follows  that  
 $|N_u^- \cap \{v \in V: \bar{d}_{uv} \leq 7/10 \}| > \frac{4}{\eps^2} \cdot |N_u^- \cap S_r \cap \{v \in V: \bar{d}_{uv} \leq 7/10 \}|$  with probability less than $\frac{1}{n^{C/12}}$ and union bounding over all possible $w$ gives an upper bound of  with probability less than $\frac{1}{n^{C/12-1}}.$

 \noindent \emph{Item 6.} Same proof as item 1, just replacing $S_d$ with $S_p$.

 \noindent \emph{Item 7.} The probability that some fixed $u,v$ with $|N_u^+ \cup N_v^+| \geq C \cdot \frac{\log n}{\varepsilon^2}$ have $\bar{d}_{uv} > \frac{2(1+C)}{6 \varepsilon^2} \log n\cdot d_{uv}$ is less than $\frac{3}{n^{C/24}}$ by Lemma \ref{lem: whp-distances}. Union bounding over any such $u,v$ pair, we see that any such event occurs with probability at most $\frac{3}{n^{C/24-2}}.$

\medskip

Combining the probabilities, we see that the probability that event $B$ occurs is upper bounded by 
 \[\frac{3}{n^{C/24-2}}+\frac{4}{n^{C/12-1}}+\frac{3}{n^{C'-1}}.\]
For choices of constants where  $C' =5$ and $C = 100$, the probability that the good event $B^c$ occurs is at least $1-1/n$ for sufficiently large $n$. 
\end{proof}

\subsection{Positive fractional cost of correlation metric for finite $p$} \label{sec: pos-frac-cost}

We first bound the $\ell_p$-norm cost of the estimated adjusted correlation metric $\tilde{d}$ on positive edges. We restate the lemma for convenience.

\whpposfraccost*

\begin{proof}[Proof of Lemma \ref{lem: whp-pos-frac-cost}]
    Let $y$ be the vector of disagreements in a fixed clustering $\mathcal{C}$, let $C(u)$ denote the set of vertices in vertex $v$'s cluster, and let $\overline{C(u)} := V \setminus C(u)$. 

     We first bound the $\ell_p$-norm cost of the estimated \emph{unadjusted} correlation metric, $\bar{d}$, on positive edges:

\begin{claim}\label{clm: pos-corr-cost-whp}
    Let $1 \leq  p < \infty$. The following bounds hold:
    
    \begin{itemize}
    \item $\mathbb{E}\left[\sum_{u \in V_0} \sum_{v \in N_u^+ \cap V_0} \bar{d}_{uv} \right]\leq O (\nicefrac{1}{\eps^2}) \cdot \opt_1$, and
        \item Conditioned on the event $B^c$, $\sum_{u \in V_0} \Big(\sum_{v \in N_u^+ \cap V_0} \bar{d}_{uv}\Big)^p  
         \leq O \left (\left ( \nicefrac{ 1}{\eps^2} \cdot \log n \right )^p \right ) \cdot \opt_p^p.$
    \end{itemize}  
\end{claim}

\begin{proof}[Proof of Claim \ref{clm: pos-corr-cost-whp}]
    Define 

    \[E_1 = \sum_{u \in V_0} \left(\sum_{v \in N_u^+ \cap C(u) \cap V_0} \frac{|N_u^+ \cap N_v^- \cap S_d| + |N_u^- \cap N_v^+ \cap S_d|}{|(N_u^+ \cup N_v^+) \cap S_d|}\right)^p\]

    \[E_2 = \sum_{u \in V_0} \left(\sum_{v \in N_u^+ \cap \widebar{C(u)} \cap V_0} \frac{|N_u^+ \cap N_v^- \cap S_d| + |N_u^- \cap N_v^+ \cap S_d|}{|(N_u^+ \cup N_v^+) \cap S_d|}\right)^p.\]

  Notice that the numerator in $E_1$ is bounded above by $y(u) + y(v)$. This is because $v \in C(u)$ implies that $(u,w)$ or $(v,w)$ is a disagreement for every $w \in (N_u^+ \cap N_v^-) \cup (N_u^- \cap N_v^+)$. So we define 

    \[E_{1a} = \sum_{u \in V_0} \left(\sum_{v \in N_u^+ \cap C(u) \cap V_0} \frac{y(u)}{|(N_u^+ \cup N_v^+) \cap S_d|}\right)^p\]

    \[E_{1b} = \sum_{u \in V_0} \left(\sum_{v \in N_u^+ \cap C(u) \cap V_0} \frac{y(v)}{|(N_u^+ \cup N_v^+) \cap S_d|}\right)^p.\]

    By Jensen's inequality, we can, at a loss a factor of $3^{p-1}$,  bound the quantities $E_{1a}, E_{1b}$, and $E_2$ individually.

    Using that $\bar{d}_{uv} \leq 1$, we have $E_2 = \sum_{u \in V} \left(\sum_{v \in N_u^+ \cap \overline{C(u)} \cap V_0} \bar{d}_{uv}\right)^p \leq \sum_{u \in V} y(u)^p$, always. Next, 

    \begin{equation*}
        E_{1a} \leq \sum_{u \in V_0} \frac{1}{|N_u^+ \cap S_d |^p} \cdot  \left(\sum_{v \in N_u^+ } y(u)\right)^p \leq \sum_{u \in V} \frac{|N_u^+|^p}{|N_u^+ \cap S_d |^p} \cdot \mathbf{1}_{\{u \in V_0\}} \cdot   y(u)^p. 
    \end{equation*}
We can now apply Proposition \ref{prop: reciprocal-corr} to obtain a bound of $\nicefrac{14}{\eps^2} \cdot \sum_{u \in V}y(u)$ in expectation for $p=1$, and of $\left(\nicefrac{C \cdot \log n}{\varepsilon^2}\right)^p  \cdot \sum_{u \in V}y(u)^p$ conditioned on the event $B^c$ for finite $p \geq 1$. Finally we bound $E_{12}$: 

\begin{align*}
    E_{1b} &\leq \sum_{u \in V_0} \sum_{v \in N_u^+ \cap V_0 \cap C(u)} \frac{|N_u^+ \cap V_0 \cap C(u)|^{p-1}}{|(N_u^+ \cup N_v^+) \cap S_d|^p} \cdot y(v)^p \\
    &\leq \sum_{u \in V_0} \sum_{v \in N_u^+ \cap V_0 \cap C(u)} \frac{|N_u^+|^{p-1}}{|N_u^+ \cap S_d|^{p-1} \cdot |N_v^+ \cap S_d|} \cdot y(v)^p \\
    &\leq \left(\frac{C \cdot \log n}{\varepsilon^2}\right)^{p-1} \cdot \sum_{u \in V_0} \sum_{v \in N_u^+ \cap V_0 \cap C(u)} \frac{1}{|N_v^+ \cap S_d|} \cdot y(v)^p \\
    &\leq \left(\frac{C \cdot \log n}{\varepsilon^2}\right)^{p-1} \cdot \sum_{v \in V_0} y(v)^p \sum_{u \in V_0 \cap N_v^+} \frac{1}{|N_v^+ \cap S_d|} \\
    &\leq \left(\frac{C \cdot \log n}{\varepsilon^2}\right)^{p-1} \cdot \sum_{v \in V} y(v)^p \cdot  \frac{|N_v^+|}{|N_v^+ \cap S_d|} \cdot \mathbf{1}_{\{u \in V_0\}}
\end{align*}
where in the first line we have applied Jensen's inequality. Applying Proposition \ref{prop: reciprocal-corr} we obtain a bound of $\nicefrac{14}{\eps^2} \cdot \sum_{u \in V}y(u)$ in expectation for $p = 1$, and of $\left(\nicefrac{C \cdot \log n}{\varepsilon^2}\right)^{p} \cdot \sum_{v \in V} y(v)^p$ conditioned on the event $B^c$ for finite $p \geq 1$.
\end{proof}

Recall $R_1$ is the set of vertices isolated by $\tilde{d}$ (see Definition \ref{def: est-adj-corr}), and $R_2 = V \setminus R_1$. Next, we need to bound the contribution of vertices $u \in R_1$ to the fractional cost.  So, we need to bound $\sum_{u \in V_0 \cap R_1} |N_u^+|^p$. Define 
\begin{align*}
        V^1 &:= V_0 \cap \Big\{u \in V : |N_u^+ \cap S_d \cap \widebar{C(u)}| \geq \nicefrac{3}{20} \cdot |N_u^+ \cap S_d| \Big\} \\
        V^2 &:= V_0 \cap \Big\{u \in V : |N_u^+ \cap S_d \cap C(u)| \geq \nicefrac{17}{20} \cdot |N_u^+ \cap S_d| \Big\} \\
        V^{2a} &:=  V^2 \cap \left\{u \in V : |N_u^- \cap C(u)| \geq |N_u^+|\right\},  \qquad V^{2b} :=  V^2 \cap \left\{u \in V : |C(u)| \leq 2 \cdot |N_u^+|\right\}.
    \end{align*}

\begin{claim} \label{clm: r1v1-whp}
Let $1 \leq p < \infty$. The following bounds hold: 
    \begin{itemize}
    \item $\mathbb{E}\left[\sum_{u \in R_1 \cap V^1} |N_u^+|^p \right] \leq O(\nicefrac{1}{\eps^2}) \cdot \opt_1$, and
        \item Conditioned on the event $B^c$, we have $\sum_{u \in R_1 \cap V^1} |N_u^+|^p \leq  O \left (\left( \nicefrac{1 }{ \eps^2} \cdot \log n\right)^p \right)\cdot \opt_p^p$.
    \end{itemize}  
\end{claim}

\begin{proof}[Proof of Claim \ref{clm: r1v1-whp}]
    Using the definition of $V^1$, we have 
    \begin{align*}
    \sum_{u \in R_1 \cap V^1} |N_u^+|^p &\leq (20/3)^p \cdot \sum_{u \in V_0} |N_u^+|^p \cdot \frac{|N_u^+ \cap S_d \cap \overline{C(u)}|^p}{|N_u^+ \cap S_d|^p} \\
    &\leq (20/3)^p \cdot \sum_{u \in V} \frac{|N_u^+|^p}{|N_u^+ \cap S_d|^p} \cdot \mathbf{1}_{\{u \in V_0\}} \cdot y(u)^p.
    \end{align*}
We can now apply Proposition \ref{prop: reciprocal-corr} to obtain a bound of $\nicefrac{280}{3\eps^2} \cdot \sum_{u \in V} y(u)$ in expectation for $p=1$, and of $\left(\frac{20}{3} \cdot \frac{C \cdot \log n}{\eps^2}\right)^p \cdot \sum_{u \in V} y(u)^p$ conditioned on the event $B^c$ for finite $p \geq 1$.
\end{proof}

\begin{claim} \label{clm: r1v2a-whp}
   Let $1 \leq p < \infty$. It is always the case that $\sum_{u \in R_1 \cap V^{2a}} |N_u^+|^p \leq  \opt_p^p$.
\end{claim}

\begin{proof}[Proof of Claim \ref{clm: r1v2a-whp}]
    By definition of $V^{2a}$, $\sum_{u \in R_1 \cap V^{2a}} |N_u^+|^p \leq \sum_{u \in R_1 \cap V^{2a}} |N_u^- \cap C(u)|^p \leq \sum_{u \in V} y(u)^p$.
\end{proof}

\begin{claim}\label{clm: r1v2b-whp}
Let $1 \leq p < \infty$. The following bounds hold: 
\begin{itemize}
    \item $\mathbb{E}\left[\sum_{u \in R_1 \cap V^{2b}} |N_u^+| \right] \leq O \left (\nicefrac{1}{\eps^{4}} \right ) \cdot \opt_1$, and
     \item Conditioned on the event $B^c$, we have $\sum_{u \in R_1 \cap V^{2b}} |N_u^+|^p \leq O \left (\left(\nicefrac{1 }{\eps^4} \cdot \log^2 n \right)^p \right ) \cdot  \opt_p^p$.
\end{itemize}
\end{claim}

\begin{proof}[Proof of Claim \ref{clm: r1v2b-whp}] 
   Recall $R_1(u):= N_u^- \cap \{v: \bar{d}_{uv} \leq 7/10\}$ (Definition \ref{def: est-adj-corr}), and that for $u \in R_1$, $|R_1(u)| \geq \frac{10}{3} \cdot |N_u^+ \cap S_d|$. For convenience, define $N_{u,v}:= N_u^+ \cap N_v^+ \cap C(u)$. 
    
    Define $\varphi(u,w) := |R_1(u) \cap N_w^+|$. Note that for $w \in N_u^+$, we have $\varphi(u,w) \leq y(u) + y(w)$, since for any $v \in R_1(u) \cap N_w^+$, either the edge $vw$ is a disagreement or the edge $vu$ is a disagreement. We will charge the cost of $u \in R_1$ to these disagreements, using a double-counting argument. First we lower bound the number of such disagreements that arise in this way for fixed $u$:

    \begin{align*}
        \sum_{w \in N_u^+ \cap C(u) \cap S_d} \varphi(u,w) = \sum_{v \in R_1(u)} |N_{u,v} \cap S_d| \geq \sum_{v \in R_1(u)} \frac{3}{20} \cdot |N_u^+ \cap S_d| \geq \frac{1}{2} \cdot |N_u^+ \cap S_d|^2 
    \end{align*}
    where in the first inequality we have used Fact \ref{fct: overlapping-nbhds-Sd}, and in the last inequality we have used the lower bound on $R_1(u)$ from above. 

    Next, applying Jensen's inequality, we have 
    \[|N_u^+ \cap C(u) \cap S_d|^{p-1} \cdot \hspace{-10pt}\sum_{w \in N_u^+ \cap C(u) \cap S_d} \hspace{-5pt}\varphi(u,w)^p \geq \left(\sum_{w \in N_u^+ \cap C(u) \cap S_d} \varphi(u,w) \right)^p \geq \frac{1}{2^p} \cdot |N_u^+ \cap S_d|^{2p}.\]

    Now we may bound the contribution of $u \in R_1 \cap V_{2b}$: 

    \begin{align*}
        \sum_{u \in R_1 \cap V^{2b}} |N_u^+|^p &\leq 2^p \cdot \sum_{u \in R_1 \cap V^{2b}} \frac{|N_u^+|^p}{|N_u^+ \cap S_d|^{2p}} \cdot |N_u^+ \cap C(u) \cap S_d|^{p-1} \cdot \sum_{w \in N_u^+ \cap C(u) \cap S_d} \varphi(u,w)^p.
    \end{align*}

As $\varphi(u,w) \leq y(u) + y(w)$, it suffices by Jensen's inequality to individually bound the following terms, at a loss of an additional factor of $2^{p-1}$: 
        \[  \sum_{u \in R_1 \cap V^{2b}} \frac{|N_u^+|^p}{|N_u^+ \cap S_d|^{2p}} \cdot |N_u^+ \cap C(u) \cap S_d|^{p-1} \cdot \sum_{w \in N_u^+ \cap C(u) \cap S_d} y(u)^p \]
        and 
        \[ \sum_{u \in R_1 \cap V^{2b}} \frac{|N_u^+|^p}{|N_u^+ \cap S_d|^{2p}} \cdot |N_u^+ \cap C(u) \cap S_d|^{p-1} \cdot \sum_{w \in N_u^+ \cap C(u) \cap S_d} y(w)^p.\]

The first term is upper bounded by 

\begin{align*}
\sum_{u \in V_0: |C(u)| \leq 2 \cdot |N_u^+|} \frac{|N_u^+|^p}{|N_u^+ \cap S_d|^{2p}} \cdot |C(u)|^p \cdot y(u)^p &\leq 2^p \cdot \sum_{u \in V} \frac{|N_u^+|^{2p}}{|N_u^+ \cap S_d|^{2p}} \cdot \mathbf{1}_{\{u \in V_0\}} \cdot y(u)^p. 
\end{align*}

Applying Proposition \ref{prop: reciprocal-corr}, we obtain a bound of $\nicefrac{664}{\eps^4} \cdot \sum_{u \in V} y(u)$ in expectation for $p=1$, and of $2^p \cdot \left(\frac{C^2 \cdot \log^2 n}{\varepsilon^4}\right)^p \cdot \sum_{u \in V} y(u)^p$ conditioned on the event $B^c$ for finite $p \geq 1$. The second term is upper bounded by

\begin{align*}
    & \quad \sum_{w \in S_d} y(w)^p \sum_{u \in V_0 \cap C(w): |C(u)| \leq 2 \cdot |N_u^+|} \frac{|N_u^+|^p}{|N_u^+ \cap S_d|^{2p}} \cdot |N_u^+ \cap C(u) \cap S_d|^{p-1} \\
    &\leq \sum_{w \in S_d} y(w)^p \sum_{u \in V_0 \cap C(w): |C(u)| \leq 2 \cdot |N_u^+|} \frac{|N_u^+|^p}{|N_u^+ \cap S_d|^{2p}} \cdot |C(w)|^{p-1} \\
    &\leq \sum_{w \in S_d} y(w)^p \sum_{u \in V_0 \cap C(w): |C(w)| \leq 2 \cdot |N_u^+|} \frac{|N_u^+|^{2p}}{|N_u^+ \cap S_d|^{2p}} \cdot \frac{|C(w)|^{p-1}}{|N_u^+|^p} \\
    &\leq 2^p \cdot  \sum_{w \in V} y(w)^p \sum_{u \in C(w)} \frac{|N_u^+|^{2p}}{|N_u^+ \cap S_d|^{2p}} \cdot \mathbf{1}_{\{u \in V_0\}} \cdot \frac{|C(w)|^{p-1}}{|C(w)|^p}. 
\end{align*}
Again applying Proposition \ref{prop: reciprocal-corr}, we obtain the same upper bounds as above for the first term. 
\end{proof}

In total, we combine the results from Claims \ref{clm: r1v1-whp}, \ref{clm: r1v2a-whp}, and \ref{clm: r1v2b-whp} to see
\begin{align}
     \mathbb{E} \left [\sum_{u \in V_0 \cap R_1} |N_u^+| \right ]\leq  \left (\nicefrac{1}{\eps^4}\right )\cdot \opt_1  
     \label{eqn: bound-sum-R1-p1}
\end{align}
and conditioning on $B^c$, for finite $p$
\begin{align}
     \sum_{u \in V_0 \cap R_1} |N_u^+|^p \leq    O\left (\left ( \nicefrac{1}{\eps^4}  \cdot \log^2 n \right )^p\right )
     \cdot \opt_p^p.
     \label{eqn: bound-sum-R1-pfinite}
\end{align}

Finally, for $u \in R_2$, we need to bound the contribution of $uv \in E^+$ such that $v \in R_1$, i.e., of those $v$ for which $\tilde{d}_{uv} = 1 \neq \bar{d}_{uv}$. We note that this case is not needed for $p=1$, since Claims \ref{clm: pos-corr-cost-whp}, \ref{clm: r1v1-whp}, \ref{clm: r1v2a-whp}, and \ref{clm: r1v2b-whp} suffice to compute the $\ell_1$-norm of the fractional cost.

\begin{claim} \label{clm: r2-whp}
Let $1 \leq p < \infty$.     Conditioned on the event $B_c$, we have
    \[\sum_{u \in R_2 \cap V_0} |N_u^+ \cap R_1 \cap V_0|^p \leq O \left (\left ( \nicefrac{ 1 }{\eps^6} \cdot  \log^3 n \right )^p \right) \cdot \opt_p^p.\]
\end{claim}

\begin{proof}[Proof of Claim \ref{clm: r2-whp}]
    Observe that \[\sum_{u \in R_2 \cap V_0} |\{v : v \in N_u^+ \cap R_1 \cap V_0 \mbox{ and } \bar{d}_{uv} \geq \nicefrac{1}{4}\}|^p \leq 4^p \cdot \sum_{u \in V_0} \left(\sum_{v \in V_0 \cap N_u^+} \bar{d}_{uv} \right)^p.\]
    By Claim \ref{clm: pos-corr-cost-whp}, this is bounded for $p=1$ in expectation by $\frac{116}{\eps^2} \cdot \opt_1$ and for finite $p$ by $ \left (\nicefrac{12C \cdot \log n}{\eps^2} \right )^p \cdot \opt_p^p$, since we conditioned on $B^c$ . 

Given the above observation, with loss of a factor of $2^{p-1}$ due to Jensen's inequality, it suffices to bound the analogous quantity with $\bar{d}_{uv} \leq \nicefrac{1}{4}$. We create a bipartite auxiliary graph $H = (R_2 \cap V_0, R_1 \cap V_0, F)$.  An edge $uv \in F$ for $u \in R_2 \cap V_0$ and $v \in R_1 \cap V_0$ if $uv \in E^+$ and $\bar{d}_{uv} \leq 1/4$. It then suffices to show 
\[ \sum_{u \in R_2 \cap V_0} \text{deg}_{H}(u)^p = O\left(\nicefrac{\log n}{\eps^2}\right)^p \cdot \sum_{v \in R_1 \cap V_0} |N_v^+|^p,\]
since then we may bound the right-hand side via Claims \ref{clm: r1v1-whp}, \ref{clm: r1v2a-whp}, and \ref{clm: r1v2b-whp} (specifically, we combine them using Equation (\ref{eqn: bound-sum-R1-p1})). We will bound via double counting the quantity $\sum_{f=uv \in F} \left(\text{deg}_H(u) + \text{deg}_H(v) \right)^{p-1}$. Let $N_H(\cdot)$ denote the neighborhoods in $H$ of the vertices. 
\begin{align}
     \sum_{f=uv \in F} \left(\text{deg}_H(u) + \text{deg}_H(v) \right)^{p-1}  
    &\leq \sum_{v \in R_1 \cap V_0} \sum_{u \in N_{H}(v)} \left(\text{deg}_H(v) + \text{deg}_H(u) \right)^{p-1} \notag \\
    &\leq \sum_{v \in R_1 \cap V_0} \sum_{u \in N_{H}(v)} \left( |N_v^+| + \nicefrac{C \cdot \log n}{\eps^2} \cdot |N_u^+ \cap S_d| \right)^{p-1} \label{eq: deg-bd} \\ &\leq \sum_{v \in R_1 \cap V_0} \sum_{u \in N_{H}(v)} \left( |N_v^+| + \nicefrac{C \cdot \log n}{\eps^2} \cdot \nicefrac{4}{3} \cdot |N_v^+ \cap S_d| \right)^{p-1} \label{eq: closeness_prop} \\ 
    &\leq \sum_{v \in R_1 \cap V_0} \sum_{u \in N_{H}(v)} (1+ \nicefrac{4}{3} \cdot \nicefrac{C \cdot \log n}{\eps^2})^{p-1}  \cdot |N_v^+|^{p-1} \notag \\
    &\leq (1+ \nicefrac{4}{3} \cdot \nicefrac{C \cdot \log n}{\eps^2})^{p-1}  \cdot \sum_{v \in R_1 \cap V_0} |N_v^+| \cdot |N_v^+|^{p-1} \notag\\ 
    &\leq ( 3 \cdot \nicefrac{C \cdot \log n}{\eps^2})^{p-1}  \cdot \sum_{v \in R_1 \cap V_0} |N_v^+|^p \notag
\end{align}
 where (\ref{eq: deg-bd}) follows from Proposition \ref{prop: reciprocal-corr} and (\ref{eq: closeness_prop}) follows from Fact \ref{fact:sim-sizes}. It now just remains to lower bound the sum:
\begin{align*}
     \sum_{f=uv \in F} \left(\text{deg}_H(u) + \text{deg}_H(v) \right)^{p-1} 
    &= \sum_{u \in R_2 \cap V_0} \sum_{v \in N_H(u)} \left(\text{deg}_H(u) + \text{deg}_H(v) \right)^{p-1} \\
    &\geq \sum_{u \in R_2 \cap V_0} \sum_{v \in N_H(u)} \text{deg}_H(u)^{p-1} \\
    &= \sum_{u \in R_2 \cap V_0} \text{deg}_H(u)^{p},
\end{align*}
 which is what we sought to show. 
\end{proof}

Combining Claims \ref{clm: pos-corr-cost-whp}, \ref{clm: r1v1-whp}, \ref{clm: r1v2a-whp}, \ref{clm: r1v2b-whp}, and \ref{clm: r2-whp} completes the proof of Lemma \ref{lem: whp-pos-frac-cost} (using only the first four claims for $p=1$) with 
$\mathbb{E}\left[\sum_{u \in V_0} \sum_{v \in N_u^+ \cap V_0} \tilde{d}_{uv} \right] \leq \nicefrac{1451}{\eps^4} \cdot \opt_1$, and conditioned on the event $B^c$, $\sum_{u \in V_0} \left(\sum_{v \in N_u^+ \cap V_0} \tilde{d}_{uv}\right)^p\leq \left ( \nicefrac{110 \cdot C^3 \cdot \log^3 n }{\eps^6}\right ) ^p \cdot \opt_p^p.$
\end{proof}

\section{Tight analysis of Algorithm \ref{alg: main-alg} for $p=1$}
\label{app: omit-proofs-l1}

While the proof of item \ref{item: thm-allp} of Theorem \ref{thm:main-all} is valid when $p=1$, the analysis is rather lossy for this case. We know there exist $O(1)$ algorithms for $\ell_1$-norm correlation clustering in the AOS model \cite{LattanziMVWZ21}, and we will show Algorithm \ref{alg: main-alg} achieves this too. 

\begin{figure}[h]
\begin{minipage}{0.65\textwidth}
\hfill
\includegraphics[height=3.5cm]{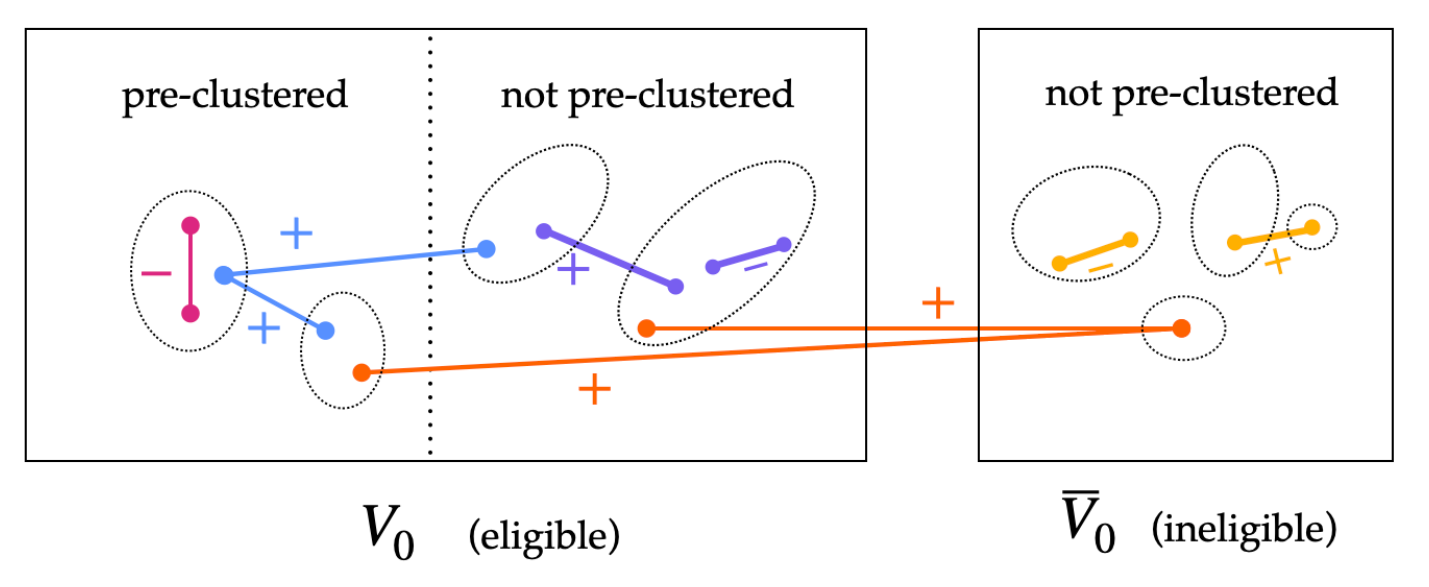}
\end{minipage}
\hfill 
\begin{minipage}{0.28\textwidth}
\footnotesize{\textcolor{myblue}{Lemma \ref{lem: type1b}}  \\ \textcolor{mypink}{Lemma \ref{lem: expected-negative}} \\
\textcolor{myyellow}{Lemma \ref{lem: pivot-bar-V0-ell1}} \\
\textcolor{mypurple}{Lemma \ref{lem: pivot-v0'-ell1}}}\\
\textcolor{myorange}{Lemma \ref{lem: type-t}} 
\end{minipage}
\captionsetup{width=.9\textwidth}
\caption{An overview of the cost analysis of Algorithm \ref{alg: main-alg} for $p=1$. As in Figure \ref{fig:summary-ellp}, solid edges are disagreements and dashed ovals are clusters, and different types of disagreements are color coded with the corresponding lemmas where we handle their charging arguments.
}
\label{fig:summary-ell1}
\end{figure}

As in the finite $\ell_p$-norm analysis (Section \ref{sec: lp-norm-analysis}), we partition our analysis into the cost of the disagreements incurred during the Pre-clustering phase (Section \ref{sec: preclustering-charging-1}), the cost of the disagreements incurred during the two individual  Pivot phases  (Section \ref{sec: pivot-charging-1}), and those cost incurred between the Pre-clustering phase and the Pivot phases (Section \ref{sec: between-phases-ell1}). See Figure \ref{fig:summary-ell1} for references to the lemmas for each type of disagreement. Note there are fewer cases than in Figure \ref{fig:summary-ellp} for the finite $\ell_p$-norm analysis, because we need not consider, e.g., whether $u \succ v$ or $v \succ u$ for the $\ell_1$-norm objective.

The analysis in Section \ref{sec: preclustering-charging-1} is for disagreements where both endpoints are eligible and at least one is pre-clustered (see the \textcolor{myblue}{blue} and \textcolor{mypink}{pink} edges in Figure \ref{fig:summary-ell1}). The analysis here is similar to the corresponding analysis (Section \ref{sec: pre-clustering-ellp}) for $\ell_p$-norms, $p \in (1,\infty)$, but involves more probabilistic subtleties in order to obtain an upper bound that both holds in expectation and nearly matches the lower bound. 

The analysis in Sections \ref{sec: pivot-charging-1} and \ref{sec: between-phases-ell1} looks most dissimilar to the corresponding $\ell_p$-norm analyses (Sections \ref{sec: pivot-phase-lp} and \ref{sec: between-phases}). For edges that have at least one endpoint in $\widebar{V_0}$ (see the \textcolor{myorange}{orange} edges and \textcolor{myyellow}{yellow} edges in Figure \ref{fig:summary-ell1}), the key idea will be that the expected positive degree of a vertex in $G[\widebar{V_0}]$ is small. This will help us charge the cost of disagreements to bad triangles (Definition \ref{def: bad-tri}) and to the (true) correlation metric $d$ (Definition \ref{def: corr}). 

Finally, for the disagreements incurred from running Modified Pivot on $G[V_0']$, we will again charge to both bad triangles and to $\tilde{d}$, as we did for the $\ell_p$-norm analysis (Lemma \ref{lem: pivot-cost-lp-v''}), but the analysis will be simpler as for the $\ell_1$-norm we may charge on a per-edge basis rather than a per-vertex basis.

\subsection{Cost of Pre-clustering phase} \label{sec: preclustering-charging-1}

In this section, we bound the cost of the pre-clustering phase in Algorithm \ref{alg: main-alg}. By this we mean that we bound the cost of disagreeing edges that have at least one endpoint that is pre-clustered. We refer the reader to Section \ref{sec: pre-clustering-ellp} for a reminder on notations and definitions related to the pre-clustering phase.

Define $t := r/(2\delta)$ to be a threshold parameter which will be used in our analysis.

\subsubsection{Cost of positive edges}

We begin in Lemma \ref{lem: type1b} by bounding the cost of positive edges that are disagreements in $\mathcal{C}_{\textsf{ALG}}$ where both endpoints are eligible and at least one is pre-clustered. This lemma is an analog of Lemmas \ref{lem: type1b-lp} / \ref{lem: type1b-lp-backwards} in the cost analysis for $\ell_p$-norms, $p \in (1, \infty)$. Here, we need not consider separate lemmas because the $\ell_1$-norm objective is equivalent to minimizing the total number of edges that are disagreements. 

\begin{lemma} \label{lem: type1b}
    The expected cost of disagreements $uv \in E^+$ where $u$ and $v$ are eligible and at least one is pre-clustered is      
    \[\mathbb{E}\Big [\sum_{u \in V_0} |\{v \in N_u^+ : v \succ u\}| \Big]  \leq O \left(\nicefrac{1}{\eps^6} \right) \cdot \opt_1.
    \]

\end{lemma}

\begin{proof}[Proof of Lemma \ref{lem: type1b}]
Observe first that each $v$ in the statement of the lemma is in $V_0$, since $v \succ u$ implies $v$ is pre-clustered. As in Lemma \ref{lem: type1b-lp}, for fixed $u \in V_0$ partition the disagreements we wish to bound into two sets: 

\[E_1(u) := \{v \in N_u^+ : v \succ u\} \cap \ball_{\tilde{d}}(u,t)\]
and
\[E_2(u) := \{v \in N_u^+ : v \succ u\} \setminus E_1(u).\]

So we wish to bound 

\[\mathbb{E}\Big [\sum_{u \in V_0} |\{v \in N_u^+ : v \succ u\}| \Big] = \mathbb{E} \Big[\sum_{u \in V_0} |E_1(u)| \Big] + \mathbb{E} \Big[\sum_{u \in V_0} |E_2(u)| \Big]  \]

By Lemma \ref{lem: whp-pos-frac-cost}, the second term is at most 

\begin{equation} \label{eq: E2-ell1}
\mathbb{E} \Big[\sum_{u \in V_0} |E_2(u)| \Big] \leq \frac{1}{t} \cdot \mathbb{E}\Big[\sum_{u \in V_0} \sum_{v \in N_u^+ \cap V_0} \tilde{d}_{uv}  \Big] \leq O(\nicefrac{1}{\eps^4}) \cdot \opt_1.
\end{equation}

To bound the first term, $\mathbb{E}[\sum_{u \in V_0} |E_1(u)|]$, we case on whether $u$ has a close neighbor sampled by $S_p$. We note that the sums below are taken over \emph{random} sets. 
\begin{align*} &\mathbb{E}\Big[\sum_{u \in V_0}  |E_1(u)| \Big] = \underbrace{\mathbb{E}\Big[\sum_{u \in V_0: \ball_{\tilde{d}}^{S_p}(u,t)=\emptyset} |E_1(u)| \Big]}_{E_{1a}} + \underbrace{\mathbb{E}\Big[\sum_{u \in V_0: \ball_{\tilde{d}}^{S_p}(u,t) \neq \emptyset} |E_1(u)| \Big]}_{E_{1b}}
\end{align*}

\begin{figure}
    \centering
  \captionsetup{width=.9\textwidth}
    \includegraphics[width=6cm]{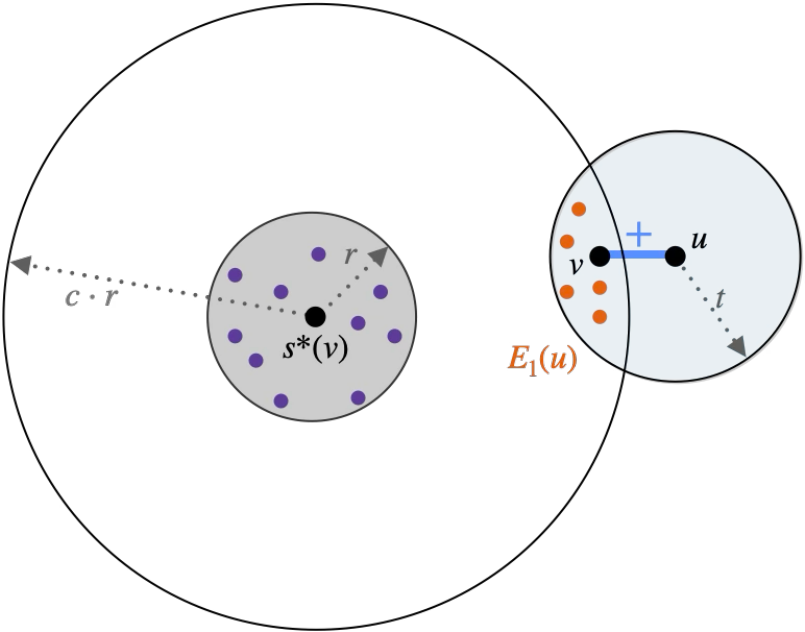}
    \caption{Bounding $E_{1b}$ in the proof of Lemma \ref{lem: type1b}. The orange points $v \in E_1(u)$  are clustered before $u$, so edges $uv$ are disagreements. To bound the cost of the orange points, it suffices to show that there are more purple points in expectation, and that the cost of the purple points can be charged to the fractional cost $\tilde{D}_0(u)$ of $u$.}
    \label{fig: type1b}
\end{figure}

\paragraph*{Bounding $E_{1a}$.} As in the proof of Lemma \ref{lem: type1b-lp}, bounding $E_{1a}$ is simpler because the expected size of $E_1(u)$ must be small. We repeat the following claim, introduced in the proof of Lemma \ref{lem: type1b-lp}, that shows there is sufficient fractional cost incident to $u$ to which to charge $E_1(u)$.

Recall 
$\tilde{D}_0(u):= \sum_{v \in N_u^+ \cap V_0} \tilde{d}_{uv} + \sum_{v \in N_u^- \cap V_0} (1-\tilde{d}_{uv}). $

\etwotwoprop*

\begin{restatable}{claim}{etwotwo} \label{claim: E1a-ell1}
For $t = \nicefrac{r}{2 \delta}$, it is the case that  

\[E_{1a} := \mathbb{E}\Big[\sum_{u \in V_0: \ball_{\tilde{d}}^{S_p}(u,t)=\emptyset} |E_1(u)| \Big] \leq O(\nicefrac{1}{\varepsilon^6} )\cdot \textsf{OPT}_1.
\]
\end{restatable}

\begin{proof}[Proof of Claim \ref{claim: E1a-ell1}]
In this case we may upper bound $|E_1(u)|$ by $|\ball_{\tilde{d}}(u,t)|$. Applying Claim \ref{clm: sufficient-charge}, we obtain: 

\begin{align*}
    E_{1a} & \leq \mathbb{E} \Big [
    \sum_{u \in V_0: \ball_{\tilde{d}}^{S_p}(u,t) =\emptyset} |\ball_{\tilde{d}}(u,t)|
    \Big] \leq 5 \cdot \mathbb{E} \Big [
    \sum_{u \in V_0: \ball_{\tilde{d}}^{S_p}(u,t) =\emptyset} |\ball_{\tilde{d}}(u,t)| \cdot \tilde{D}_0(u)
    \Big] 
    \\
&= 5 \cdot \mathbb{E}\Big[ \mathbb{E}\Big[ \sum_{u \in V_0} |\ball_{\tilde{d}}(u,t)| \cdot \tilde{D}_0(u) \cdot  \mathbf{1}_{\{\ball_{\tilde{d}}^{S_p}(u,t) = \emptyset\}}  \Big | S_d, S_r \Big]\Big] \\
    &\leq 5 \cdot  \mathbb{E}\Big[\sum_{u \in V_0} |
    \ball_{\tilde{d}}(u,t)| \cdot \tilde{D}_0(u) \cdot \mathbb{E}[ \mathbf{1}_{\{\ball_{\tilde{d}}^{S_p}(u,t) = \emptyset\}} \mid S_d, S_r ] \Big] \\
    &=5 \cdot \mathbb{E}\Big[\sum_{u \in V_0} |\ball_{\tilde{d}}(u,t)| \cdot \tilde{D}_0(u)  \cdot (1-q(\varepsilon))^{|\ball_{\tilde{d}}(u,t)|} \Big] \notag
   \\
   &= \nicefrac{10}{\eps^2} \cdot \mathbb{E}\Big[\sum_{u \in V_0} \tilde{D}_0(u) \Big] = \nicefrac{10}{\varepsilon^2} \cdot \nicefrac{2128} {\varepsilon^4} \cdot \textsf{OPT}_1 = \nicefrac{21280}{\varepsilon^6} \cdot \textsf{OPT}_1. \notag
\end{align*}
In the second line, the outer expectation is over $S_d$ and $S_r$ while the inner is over $S_p$. Then in the second to last line, we use Corollary \ref{cor: subsample-independence}, and in the last line we use Corollary \ref{cor: bounded-exp-frac-cost}.  
\end{proof}

\paragraph*{Bounding $E_{1b}$.} 

For each $u \in V_0$ with $\ball_{\tilde{d}}^{S_p}(u,t) \neq \emptyset$, choose a fixed but arbitrary $z(u) \in \ball_{\tilde{d}}^{S_p}(u,t)$. Note that because $z(u) \in S_p$ and $\tilde{d}_{u, z(u)} \leq t \leq c \cdot r$, we know that $z(u)$ is a \emph{candidate} for clustering $u$, so in particular $s^*(u)$ exists. 
Note $z(u)$ is a random variable depending on $S_p, S_d,$ and $ S_r$, but \emph{not} on $S_b$.

Define

\[B(u) := \bigcup_{v \in E_1(u)} \ball^{S_b}_{\tilde{d}}(s^*(v), r),\] 
that is, $B(u)$ is the union of balls in $S_b$, cut out around the vertices that cluster the vertices in $E_1(u)$. 

We repeat the following three claims from the proof of Lemma \ref{lem: type1b-lp}.

\upperbdCu*

\lowerbdAu*

\upperbdAu*

\begin{claim} \label{clm: E1b-ell1}
    For $t = \nicefrac{r}{2 \delta}$, it is the case that  
\[E_{1b} := \mathbb{E}\Big[\sum_{u \in V_0: \ball_{\tilde{d}}^{S_p}(u,t) \neq \emptyset} |E_1(u)| \Big] \leq O(\nicefrac{1}{\eps^6}) \cdot \textsf{OPT}_1.
\]
\end{claim}

The following proof will look a bit different from the analogous claim in the proof of Lemma \ref{lem: type1b-lp}, Claim \ref{claim: E1b-lp}. There, we split $E_{1b}$ based on whether $|\ball_{\tilde{d}}(z(u), r)|$ was above or below a threshold of $\Theta(\nicefrac{\log n}{\eps^2})$. Here we will need to be less lossy. 

\begin{proof}[Proof of Claim \ref{clm: E1b-ell1}]
We split the sum based on whether or not the ball around $u$'s candidate cluster center $z(u)$, restricted to points in $S_b$, is non-empty. (Note this ball may be empty because $z(u)$ is not necessarily in $S_b$.) We then apply Claim \ref{clm: upper-bd-Cu}.

\begin{align}
    E_{1b} 
    &\leq \mathbb{E}\Big[\sum_{u \in V_0: \ball_{\tilde{d}}^{S_p}(u,t) \neq \emptyset}  |\ball_{\tilde{d}}(z(u), r)| \Big] \notag \\
    &\leq \mathbb{E}\Big[\sum_{u \in V_0: \ball_{\tilde{d}}^{S_p}(u,t) \neq \emptyset}  |\ball_{\tilde{d}}(z(u), r)| \cdot \mathbf{1}_{\{|\ball^{S_b}_{\tilde{d}}(z(u), r)| \geq 1\}} \Big] \label{eq: E1b-1-ell1} \\
    & \qquad +  \mathbb{E}\Big[\sum_{u \in V_0: \ball_{\tilde{d}}^{S_p}(u,t) \neq \emptyset}  |E_1(u)| \cdot \mathbf{1}_{\{|\ball^{S_b}_{\tilde{d}}(z(u), r)| = 0\}} \Big] \label{eq: E1b-2-ell1}
\end{align}
 Define  $\frac{1}{|\ball^{S_b}_{\tilde{d}}(z(u), r)|} \cdot \mathbf{1}_{\{|\ball^{S_b}_{\tilde{d}}(z(u), r)| \geq 1\}}$ to be 0 when $|\ball^{S_b}_{\tilde{d}}(z(u), r)| = 0$. Then we upper bound the quantity in line (\ref{eq: E1b-1-ell1}): 
\begin{align*}
   & \quad \mathbb{E}\Big[\sum_{u \in V_0: \ball_{\tilde{d}}^{S_p}(u,t)}  |\ball_{\tilde{d}}(z(u), r)| \cdot \mathbf{1}_{\{|\ball^{S_b}_{\tilde{d}}(z(u), r)| \geq 1\}} \Big] \\
    &\leq  \mathbb{E}\Big[ \sum_{u \in V_0: \ball_{\tilde{d}}^{S_p}(u,t)} 8 \cdot \tilde{D}_0(u) \cdot |\ball_{\tilde{d}}(z(u), r)| \cdot \frac{1}{|\ball^{S_b}_{\tilde{d}}(z(u), r)|} \cdot \mathbf{1}_{\{|\ball^{S_b}_{\tilde{d}}(z(u), r)| \geq 1\}} \Big] \\
    &= \mathbb{E}\Bigg[\sum_{u \in V_0: \ball_{\tilde{d}}^{S_p}(u,t)} 8 \cdot \tilde{D}_0(u) \cdot |\ball_{\tilde{d}}(z(u), r)| \cdot\mathbb{E}\Big[\frac{1}{|\ball^{S_b}_{\tilde{d}}(z(u), r)|} \cdot \mathbf{1}_{\{|\ball^{S_b}_{\tilde{d}}(z(u), r)| \geq 1\}} \Big | S_d, S_r, S_p \Big] \Bigg].
\end{align*}
In the second line we have applied Claims \ref{clm: lower-bd-Au} and \ref{clm: upper-bd-Au}. In the last line we have pulled out known factors. Finally, applying Lemma \ref{lem: exp-reciprocal} 
with $A = |\ball_{\tilde{d}}(z(u), r)|$, $S_* = S_b$, and $\mathcal{B} = \{S_d,S_r,S_p\}$, we continue upper bounding (\ref{eq: E1b-1-ell1}). 
\begin{align*}
  &\quad \mathbb{E}\Big[\sum_{u \in V_0: \ball_{\tilde{d}}^{S_p}(u,t)}   \tilde{D}_0(u) \cdot |\ball_{\tilde{d}}(z(u), r)| \cdot \frac{112}{\eps^2 \cdot |\ball_{\tilde{d}}(z(u), r)|} \Big] \\
  &= \nicefrac{112}{\varepsilon^2} \cdot \mathbb{E}\Big[ \sum_{u \in V_0} \tilde{D}_0(u) \Big] = \nicefrac{112}{\varepsilon^2} \cdot \nicefrac{2128} {\varepsilon^4} \cdot  \textsf{OPT}_1 = \nicefrac{238336}{\eps^6 \cdot \opt_1},
\end{align*}
where in the last line we have used Corollary \ref{cor: bounded-exp-frac-cost}.

Finally, we upper bound the quantity in line (\ref{eq: E1b-2-ell1}): 
\begin{align*}
  & \quad \mathbb{E}\Big[\sum_{u \in V_0: \ball_{\tilde{d}}^{S_p}(u,t) \neq \emptyset}  |E_1(u)| \cdot \mathbf{1}_{\{|\ball^{S_b}_{\tilde{d}}(z(u), r)| = 0\}} \Big] \\
      &\leq 5 \cdot \mathbb{E}\Big[\sum_{u \in V_0: \ball_{\tilde{d}}^{S_p}(u,t) \neq \emptyset}  \tilde{D}_0(u) \cdot |\ball_{\tilde{d}}(z(u), r)| \cdot \mathbf{1}_{\{|\ball^{S_b}_{\tilde{d}}(z(u), r)| = 0\}} \Big]  \\
     &= 5 \cdot 
  \mathbb{E}\Bigg [\sum_{u \in V_0: \ball_{\tilde{d}}^{S_p}(u,t) \neq \emptyset} \tilde{D}_0(u)  \cdot |\ball_{\tilde{d}}(z(u), r)| \cdot \mathbb{E}\Big[\mathbf{1}_{\{|\ball^{S_b}_{\tilde{d}}(z(u), r)| =0\}} \Big | S_d, S_r, S_p \Big]  \Bigg] \\
    &  = 5 \cdot \mathbb{E}\Big[\sum_{u \in V_0: \ball_{\tilde{d}}^{S_p}(u,t)} \tilde{D}_0(u)  \cdot |\ball_{\tilde{d}}(z(u), r)| \cdot (1-q(\varepsilon))^{|\ball_{\tilde{d}}(z(u), r)|} \Big] 
     \\
     &= \nicefrac{10}{\eps^2} \cdot \mathbb{E}\Big[\sum_{u \in V_0} \tilde{D}_0(u) \Big] = \nicefrac{10}{\varepsilon^2} \cdot \nicefrac{2128} {\varepsilon^4} \cdot \textsf{OPT}_1 = \nicefrac{21280}{\varepsilon^6} \cdot \textsf{OPT}_1.
\end{align*}
In the first line we have used Claims \ref{clm: sufficient-charge} and \ref{clm: upper-bd-Cu}. In the third line, we have taken out known quantities, as
the outer expectation is over $S_d,S_r,S_p$ and the inner expectation is over $S_b$. Then in the penultimate line we have used Corollary \ref{cor: subsample-independence}, and in the last line we have used Corollary \ref{cor: bounded-exp-frac-cost}.

Combining the upper bounds on (\ref{eq: E1b-1-ell1}) and (\ref{eq: E1b-2-ell1}) finishes the claim.

\end{proof}

Combining the upper bounds in line (\ref{eq: E2-ell1}) and Claims \ref{claim: E1a-ell1} and \ref{clm: E1b-ell1} finishes the lemma.
\end{proof}

\subsubsection{Cost of negative edges} \label{subsec: l1-neg-edge}

\begin{lemma}\label{lem: expected-negative}
    The expected cost of negative disagreements incurred during the Pre-clustering phase is 
    \[\mathbb{E}\Big[\sum_{u \in V_0} |  \big \{v \in N_u^- : v \text{ clustered with } u \big \} | \Big]
    \leq  O(\nicefrac{1}{\eps^2}) \cdot \textsf{OPT}_1.
    \]
\end{lemma}

\begin{proof}[Proof of Lemma \ref{lem: expected-negative}]
    If $u,v \in V_0$ are clustered together, there exists $s^* \in S_p$ such that $s^* = s^*(u) = s^*(v)$ (possibly with $s^* = u$ or $s^* = v$). By the approximate triangle inequality (Lemma \ref{lem: tri-inequality}),
    \[\tilde{d}_{uv} \leq \delta \cdot (\tilde{d}_{us^*} + \tilde{d}_{vs^* }) \leq 2\delta c r \leq \nicefrac{7}{10}.\] 
    Since $\tilde{d}_{uv} < 1$, we know that $\tilde{d}_{uv} = \bar{d}_{uv}$ and also that $u \in R_2$.

    Thus, to prove the lemma, it suffices to bound $\mathbb{E}\left[\sum_{u \in V_0 \cap R_2} |R_1(u)| \right]$.
    Since we bound this exact quantity in the proof of Lemma \ref{lem: whp-neg-frac-cost}, the proof is complete. 
\end{proof}

\subsection{Cost of Pivot phase} \label{sec: pivot-charging-1}
Let $G' = (V', E')$ be the subgraph induced by the unclustered vertices, where $V'$ is as defined in Algorithm \ref{alg: main-alg}. In this section, we bound the expected cost of disagreements in $G'$. 
Recall that $V_0$ is the set of eligible vertices, i.e., those vertices $v \in V$ such that $|N_v^+ \cap S_d| \neq \emptyset$. So $V'$ contains the vertices that are \textit{not} eligible, as well as vertices that are eligible but that are far from all vertices in $S_p$:

\[V' = \widebar{V_0} \cup V_0' \text{, where } V_0' = V_0 \cap \{v \in V: \tilde{d}_{vu_i} > c \cdot r  \text{ for all } u_i \in S_p\} .\]

Algorithm \ref{alg: main-alg} runs the standard Pivot algorithm on $G[\widebar{V_0}]$, and runs Modified Pivot on $G[V_0']$. In Lemma \ref{lem: pivot-bar-V0-ell1}, we bound the disagreements incurred by Pivot on $G[\widebar{V_0}]$, and in Lemma \ref{lem: pivot-v0'-ell1}, we bound the disagreements incurred by Modified Pivot on $G[V_0']$. 

\subsubsection{Disagreements in $G[\widebar{V_0}]$}

The key idea is that if the expected positive degree of every vertex in a graph is small, then the cost of Pivot on that graph is small as well. 

Fix an optimal clustering $\mathcal{C}_\opt$ (on the entire graph $G$) for the $\ell_1$-norm objective, and let $E^* \subseteq E$ be the disagreements in $G$ with respect to $\mathcal{C}_\opt$. For any $V_H \subseteq V$, let $H = (V_H, E_H)$  be the subgraph induced by $V_H$. Then $E^* \cap E_H$ is the set of disagreements in $H$ with respect to $\mathcal{C}_\opt$. Note that $E^*$ is deterministic, but in context $E_H$ may not be. Let $\textsf{cost}_{H}(\textsf{Pivot})$ be the number of disagreements incurred by running Pivot on $H$. For $i \in V_H$, let $\textsf{deg}^+_{H}(i)$ be the positive degree of vertex $i$ in $H$. The following lemma is proved in \cite{LattanziMVWZ21}:

\begin{lemma}[Lemmas 2 and 3 in \cite{LattanziMVWZ21}] \label{lem: pivot-degree-bd}
    Let $\mathcal{C}_\opt, E^*, H, V_H, E_H$ be as above. Then the expected number of disagreements that (unmodified) Pivot\footnote{We do \emph{not} assume that the vertices arrive in random order here, unlike, e.g., in the 3-approximation of \cite{ACN-pivot}.} makes on $H$, $\mathbb{E}[\textsf{cost}_{H}(\textsf{Pivot})]$, is at most
    \[\mathbb{E}\Big[\sum_{uv \in E_H \cap E^*}(\textsf{deg}_{H}^+(u) + \textsf{deg}_{H}^+(v))\Big].\]
\end{lemma}

We bound the expected cost of the disagreements within $G[\widebar{V_0}]$ in the following claim.

\begin{lemma}\label{lem: pivot-bar-V0-ell1}
    The expected cost of disagreements in $G[\widebar{V_0}]$ is 
    $O(\nicefrac{1}{\varepsilon^2})\cdot \textsf{OPT}_1$.
\end{lemma}

\begin{proof}[Proof of Lemma \ref{lem: pivot-bar-V0-ell1}]
    To prove the claim, we apply Lemma \ref{lem: pivot-degree-bd} with $H = G[\widebar{V_0}]$. We have 
    \begin{align*}
        \mathbb{E}[\textsf{cost}_{G[\widebar{V_0}]}(\textsf{Pivot})] &\leq \sum_{uv \in E^*} \mathbb{E}\left[\mathbf{1}_{\{u \in \widebar{V_0}\}} \cdot |N_u^+|+ \mathbf{1}_{\{v \in \widebar{V_0}\}} \cdot |N_v^+|\right] \\
        &= \sum_{uv \in E^*} \left(|N_u^+| \cdot \mathbb{P}(u \in \widebar{V_0}) + |N_v^+| \cdot \mathbb{P}(v \in \widebar{V_0})\right)\\
        &= \sum_{uv \in E^*} \left(|N_u^+| \cdot (1-q(\varepsilon))^{|N_u^+|} + |N_v^+| \cdot (1-q(\varepsilon))^{|N_v^+|}\right) \\
        &= \sum_{uv \in E^*} \nicefrac{2}{q(\varepsilon)} = \nicefrac{4}{\varepsilon^2}\cdot \nicefrac{1}{2} \cdot \textsf{OPT}_1 = \nicefrac{2}{\eps^2} \cdot \opt_1.
    \end{align*}
\end{proof}

\subsubsection{Disagreements in $G[V_0']$}

Next, we bound the expected cost of the disagreements in $G[V_0']$. Recall that on this graph, we run a \emph{modified} version of the Pivot algorithm, where clusters are formed by vertices grabbing their \emph{close}, positive neighbors (rather than just their positive neighbors as is done in classic Pivot). (See the definition of $E_c$ in the second \textsf{else} statement in Algorithm \ref{alg: main-alg}). 

\begin{lemma}
\label{lem: pivot-v0'-ell1}
    The expected cost of disagreements in $G[V_0']$ is at most 
     $O( \nicefrac{1}{\eps^6}) \cdot \textsf{OPT}_1.$
\end{lemma}

\begin{proof}[Proof of Lemma \ref{lem: pivot-v0'-ell1}]
    Let $E_0'$ denote the edge set of $G[V_0']$. Let $E_{\textsf{ALG}} \subseteq E_0'$ denote the disagreements made by $\mathcal{C}_{\textsf{ALG}}$ on $G[V_0']$, that is, these are the disagreements incurred by running the modified version of Pivot on $G[V_0']$. Let $E^*$ is the set of disagreements in an optimal solution $\mathcal{C}_\opt$ for the $\ell_1$-norm.

   All disagreements in $E_{ALG} \cap E^*$ can be charged directly to $\textsf{OPT}_1$. We partition the remaining disagreements in $E_{\textsf{ALG}}$ into three sets: 

      \[S_1:= \{uv \in E^+ \cap E_{\textsf{ALG}} \cap \widebar{E^*} \mid \tilde{d}_{uv} > c\cdot r \}, \qquad S_2 := \{uv \in E^+ \cap E_{\textsf{ALG}} \cap \widebar{E^*} \mid \tilde{d}_{uv} \leq c \cdot r\},\]
      \[S_3 := E^- \cap E_{\textsf{ALG}} \cap \widebar{E^*}.\]

      So

    \begin{equation} \label{eqn: ell1-v0'-total}
      |E_{\textsf{ALG}}| = |E_{\textsf{ALG}} \cap E    ^*| + |S_1| + |S_2| + |S_3| \leq \nicefrac{1}{2} \cdot \opt_1 +  |S_1| + |S_2| + |S_3|.
    \end{equation}

    \paragraph*{Bounding $S_1$.}

    We charge the cost of all $uv \in S_1$ to $\tilde{d}$. Specifically, 
    \begin{equation} \label{eq: ell1-pivot-s1}
    \mathbb{E}[|S_1|] \leq \nicefrac{1}{c \cdot  r} \nicefrac{1}{2} \cdot \mathbb{E}\Big[\sum_{u \in V_0} \sum_{v \in N_u^+ \cap V_0} \tilde{d}_{uv} \Big] = \nicefrac{2962}{\eps^4} \cdot  \textsf{OPT}_1,
    \end{equation}
    where we have applied Lemma \ref{lem: whp-pos-frac-cost} in the last equality.

\paragraph*{Bounding $S_2$.}
     For $uv \in S_2$, we have  $\tilde{d}_{uv} \leq c \cdot r$. So there are only two ways for $uv$ to be in $E_{\textsf{ALG}}$: 

    \begin{itemize}
    \item The first is if $uv$ is on a bad triangle (recall Definition \ref{def: bad-tri}) $uvw$ where, WLOG, $wu \in E^+ \cap E_0'$, $wv \in E^- \cap E_0'$, and Pivot clusters $w$ and $u$ together, but not $v$. Since $uvw$ is a bad triangle, and $uv \not \in E^*$ by assumption, we have that either $wu \in E^*$ or $wv \in E^*$. We  charge the cost of the disagreement $uv$ made by $\mathcal{C}_\textsf{ALG}$ to whichever one is in $E^*$ (choosing one arbitarily if both $wu$ and $wv$ are in $E^*$). 

    \item The second is if $uv$ is on a triangle $uvw$ of \textit{all} positive edges such that, WLOG, $\tilde{d}_{wv}  < c  \cdot r$ but $\tilde{d}_{wu} \geq c \cdot r$, and Pivot clusters $w$ and $v$ together, but not $u$. Since  $\tilde{d}_{wu} \geq c \cdot r$, we can charge the cost of the disagreement $uv$ made by $\mathcal{C}_{\textsf{ALG}}$ to $\nicefrac{1}{c \cdot r} \cdot \tilde{d}_{wu}$. 
    \end{itemize}

    We need to show that the edges in $E^*$ are not charged too many times in the former case, and that $\tilde{d}$ is not charged too many times in the latter case. Call the subset of disagreements in $S_2$ satisfying the former case $S_{2a}$, and those satisfying the latter case $S_{2b}$. 

\begin{claim} \label{clm: ell1-pivot-s2a}
$\mathbb{E}[|S_{2a}|] \leq O(\nicefrac{1}{\eps^2}) \cdot \opt_1$.
\end{claim}

\begin{proof}[Proof of Claim \ref{clm: ell1-pivot-s2a}]
    Fix $xy \in E^* \cap E_0'$. We need to compute how many times $xy$ can be charged by edges in $S_{2a}$. For an edge in $S_{2a}$ to charge $xy$, it must be that either one of its endpoints is $x$ and the other is in $\ball_{\tilde{d}}(x, c \cdot r)$; or, one of its endpoints is $y$ and the other endpoint is in $\ball_{\tilde{d}}(y, c \cdot r)$. Therefore,

    \begin{align*}
       \mathbb{E}[|S_{2a}|] &\leq \mathbb{E}\Big[\hspace{-3mm} \sum_{xy \in E^* \cap E_0'} \hspace{-3mm} \left(|\ball_{\tilde{d}}(x, c \cdot r)| 
     + |\ball_{\tilde{d}}(y, c  \cdot r)|\right) \Big]\\
     &\leq \sum_{xy \in E^*} \mathbb{E}\left[\mathbf{1}_{\{x \in V_0'\}} \cdot |\ball_{\tilde{d}}(x, c \cdot r)| + \mathbf{1}_{\{y \in V_0'\}} \cdot |\ball_{\tilde{d}}(y, c  \cdot r)|  \right] \\
     &\leq \sum_{xy \in E^*} \mathbb{E}\Big[\mathbf{1}_{\{\ball_{\tilde{d}}^{S_p}(x, c \cdot r) = \emptyset \}} \cdot |\ball_{\tilde{d}}(x, c  \cdot r)|\Big] + \mathbb{E}\Big[\mathbf{1}_{\{\ball_{\tilde{d}}^{S_p}(y,c \cdot r) \neq \emptyset\}} \cdot |\ball_{\tilde{d}}(y, c  \cdot r)|  \Big] \\
     &\leq \sum_{xy \in E^*} \Bigg(\mathbb{E}\Big[|\ball_{\tilde{d}}(x, c  \cdot r)| \cdot \mathbb{E}\Big[\mathbf{1}_{\{\ball_{\tilde{d}}^{S_p}(x, c \cdot r) = \emptyset\}} | S_d, S_r \Big]\Big] \\
     &\qquad \qquad + \mathbb{E}\Big[|\ball_{\tilde{d}}(y, c  \cdot r)| \cdot \mathbb{E}\Big[\mathbf{1}_{\{\ball_{\tilde{d}}^{S_p}(y, c \cdot r) = \emptyset\}} | S_d, S_r \Big]\Big]\Bigg) \\
     &= \sum_{xy \in E^*} \mathbb{E}[|\ball_{\tilde{d}}(x, c \cdot r)| \cdot (1-q(\eps))^{|\ball_{\tilde{d}}(x, c \cdot r)|}] + \mathbb{E}[|\ball_{\tilde{d}}(y, c \cdot r)| \cdot (1-q(\eps))^{|\ball_{\tilde{d}}(y, c \cdot r)|}] \\
     &\leq \nicefrac{4}{\eps^2} \cdot \nicefrac{1}{2} \cdot \opt_1 = \nicefrac{2}{\eps}^2 \cdot \opt_1,
    \end{align*}
    where in the penultimate line we used Corollary \ref{cor: subsample-independence}. 
\end{proof}

    \begin{claim} \label{clm: ell1-pivot-s2b}
       $\mathbb{E}[|S_{2b}|] \leq O(\nicefrac{1}{\eps^6} )\cdot \opt_1$.
    \end{claim}

\begin{proof}[Proof of Claim \ref{clm: ell1-pivot-s2b}]

    Fix $xy \in \{e \in E_0' \mid \tilde{d}_e \geq c \cdot r\}$. We need to upper bound how many times $xy$ is charged by edges in $S_{2b}$. The situation is the same as in Claim \ref{clm: ell1-pivot-s2a} for $S_{2a}$: For an edge in $S_{2b}$ to charge $xy$, it must be that either one of its endpoints is $x$ and the other is in $\ball_{\tilde{d}}(x, c \cdot r)$; or, one of its endpoints is $y$ and the other endpoint is in $\ball_{\tilde{d}}(y, c \cdot r)$. Therefore,

    \begin{align*}
    \mathbb{E}[|S_{2b}|] 
    &\leq \mathbb{E}\Big[\hspace{-3mm} \sum_{\substack{xy \in E^+ \cap E_0':\\ \tilde{d}_{xy} \geq c \cdot r}} \hspace{-3mm}\left(|\ball_{\tilde{d}}(x, c \cdot r)| + |\ball_{\tilde{d}}(y, c \cdot   r)|\right)\Big] \\
    &\leq \sum_{xy \in E^+} \nicefrac{1}{c \cdot r} \cdot \mathbb{E}\left[\mathbf{1}_{\{x \in V_0'\}} \cdot \tilde{d}_{xy} \cdot |\ball_{\tilde{d}}(x, c \cdot r)| + \mathbf{1}_{\{y \in V_0'\}} \cdot \tilde{d}_{xy} \cdot |\ball_{\tilde{d}}(y, c  \cdot r)| \right] \\
    &\leq \nicefrac{1}{c \cdot r} \cdot \sum_{xy \in E^+}  \Bigg(\mathbb{E}\Big[\mathbf{1}_{\{\ball_{\tilde{d}}^{S_p}(x, c \cdot r) = \emptyset \}} \cdot \tilde{d}_{xy} \cdot |\ball_{\tilde{d}}(x, c \cdot r)|\Big] \\
    &\qquad \qquad + \mathbb{E}\Big[\mathbf{1}_{\{\ball_{\tilde{d}}^{S_p}(y, c \cdot r) = \emptyset \}} \cdot \tilde{d}_{xy} \cdot |\ball_{\tilde{d}}(y, c \cdot  r)|  \Big] \Bigg)\\
    &= \sum_{xy \in E^+} \frac{2}{ c \cdot r \cdot q(\varepsilon)} \cdot \tilde{d}_{xy} = \nicefrac{11845}{\eps^6} \cdot \opt_1
    \end{align*}
    where in the last line we have conditioned on $S_d, S_r$, applied Corollary \ref{cor: subsample-independence} in the same way as in the proof of Claim \ref{clm: ell1-pivot-s2a}, and then finished off with Lemma \ref{lem: whp-pos-frac-cost}.
\end{proof}

\paragraph*{Bounding $S_3$.}
    Finally, we bound $\mathbb{E}[|S_3|]$. Let $uv \in S_3$. Then the only way that $u,v$ can be clustered together is if $u,v$ are clustered by the same pivot $w$, where $w \in N_u^+ \cap N_v^+$ and $\tilde{d}_{uw}, \tilde{d}_{vw} \leq c \cdot r$. But then $1-\tilde{d}_{uv} \geq 1- 2\delta c r$, so we have 
    \begin{equation} \label{eq: ell1-pivot-s3}
        \mathbb{E}[|S_3|] \leq \frac{1}{1-2\delta cr} \cdot \mathbb{E}\left[\sum_{u \in V_0} \sum_{v \in N_u^- \cap V_0} (1-\tilde{d}_{uv})\right] = 
    \nicefrac{1129}{\eps^2} \cdot \textsf{OPT}_1, 
    \end{equation}

    where the last equality is by Lemma \ref{lem: whp-neg-frac-cost}. 

    Continuing from line (\ref{eqn: ell1-v0'-total}), and substituting in from lines (\ref{eq: ell1-pivot-s1}) and (\ref{eq: ell1-pivot-s3}) and Claims \ref{clm: ell1-pivot-s2a} and \ref{clm: ell1-pivot-s2b}, 
     \[\mathbb{E}[|E_{\textsf{ALG}}|] \leq (\nicefrac{1}{2} + \nicefrac{2962}{\eps^2} + \nicefrac{2}{\eps^2} + \nicefrac{11845}{\eps^6} + \nicefrac{1129}{\eps^2}) \cdot \opt_1.\]
\end{proof}

\subsection{Cost between the Pre-clustering phase and the Pivot phase} \label{sec: between-phases-ell1}

In the next lemma, Lemma \ref{lem: type-t}, we bound the cost of disagreeing positive edges where one endpoint is eligible and the other is ineligible. Recall a vertex $u$ is ineligible ($u \in \widebar{V_0})$ when its positive neighborhood is \emph{not} sampled by $S_d$; in this case, the distances $\tilde{d}_{uv}$ are not meaningful, so we will not be able to  charge the cost of these disagreements to $\tilde{d}$. Instead, we will use two different surrogates for optimal: bad triangles (Definition \ref{def: bad-tri}) and the fractional cost of the (true) correlation metric $d$ (Definition \ref{def: corr}). These will be useful for charging because every clustering must make at least one disagreement on each bad triangle, and the fractional cost of $d$ --restricted to positive edges -- is $O(1)$ approximate to optimal for the $\ell_1$-norm objective (Lemma \ref{lem: bdd-pos-frac}).

\begin{lemma}\label{lem: type-t}
The expected cost
of disagreements $uv \in E^+$ with $u$ ineligible and $v$ eligible  is 
    \[\mathbb{E}\Big[\sum_{u \in \widebar{V_0}} |N_u^+ \cap V_0| \Big]  \leq   O(\nicefrac{1}{\eps^4} ) \cdot \textsf{OPT}_1.\]
\end{lemma}

\begin{proof}[Proof of Lemma \ref{lem: type-t}]

First note that the second bound in the statement of the lemma follows from the first bound, since $v \succ u$ implies $v$ is pre-clustered, so $v \in V_0$. 

Define $D^+(u)$ to be the fractional cost of the positive edges incident to $u$ with respect to the (actual) correlation metric $d$, that is, $D^+(u) := \sum_{v \in N_u^+} d_{uv}$. It suffices to bound
\[\mathbb{E}\Big[\sum_{\substack{u \in \widebar{V_0}:~ D^+(u) > 1/4}} |N_u^+| \Big]  + \mathbb{E}\Big[\sum_{\substack{u \in \widebar{V_0}:~ D^+(u) \leq 1/4}} |N_u^+ \cap V_0| \Big], \]
where the expectation is taken over the randomness of the sample $S_d$.
Bounding the first term is straightforward. We use Corollary \ref{cor: subsample-independence} to see that $\mathbb{P}[u \in \widebar{V_0}] = \prod_{v \in N_u^+}  \mathbb{P}(v \not \in S_d) = (1-q(\varepsilon))^{|N_u^+|}$. Then,  we charge the cost of $|N_u^+|$ to $D^+(u)$, and bound the fractional cost of positive edges with Lemma \ref{lem: bdd-pos-frac}.

\begin{align*}
\mathbb{E}\Big[&\sum_{\substack{u \in \widebar{V_0}:~ D^+(u) > 1/4}} |N_u^+| \Big] \leq 4 \cdot \mathbb{E}\left[\sum_{u \in V} |N_u^+| \cdot D^+(u) \cdot \mathbf{1}_{\{u \in \widebar{V_0}\}}  \right] \\
&= 4 \cdot \sum_{u \in V} D^+(u) \cdot |N_u^+| \cdot (1-q(\varepsilon))^{|N_u^+|}  \leq \nicefrac{8}{\eps^2} \cdot \sum_{u \in V} D^+(u) \leq \nicefrac{24}{\eps^2} \cdot \textsf{OPT}_1.
\end{align*}

To bound the second term, fix $u \in \widebar{V_0}$. Suppose that $u$ is part of a perfect clique, i.e., $u$ is incident to no bad triangles. Then if $w \in N_v^+$ and $v \in N_u^+$, then $w \in N_u^+$. If this is the case, then $v \in \widebar{V_0}$ for each $v \in N_u^+$; for, if $N_v^+ \cap S_d \neq \emptyset$, then $N_u^+ \cap S_d \neq \emptyset$, since $N_u^+ = N_v^+$. Thus, none of $u$'s positive neighbors $v$ will be in $V_0$, and therefore $u$ will not contribute to the second sum above.

Fix an optimal clustering $\mathcal{C}_\opt$ for the $\ell_1$-norm objective, and let $E^*$ be the disagreements with respect to $\mathcal{C}_\opt$. By the above, we may assume that the second sum above is further restricted to $u$ such that $u$ is incident to a bad triangle. Thus we may arbitrarily map each $u$ to a bad triangle $T(u)$ that is incident to $u$. In turn, since $\mathcal{C}_\opt$ must make a disagreement on $T(u)$, we may map $u$ to $e(u) \in E^*$ such that $e(u)$ is on $T(u).$

It now suffices to bound
$\mathbb{E}[\sum_{\substack{u \in \widebar{V_0}:~ D^+(u) \leq 1/4}} |N_u^+| ]$ (where the sum is restricted to $u$ that is incident to a bad triangle, though we omit this for ease).  
The idea is to charge the cost of $|N_u^+|$ to the disagreement $e_u$. We have 
\begin{align*}
    \mathbb{E}\Bigg[\sum_{\substack{u \in \widebar{V_0}:\\ D^+(u) \leq 1/4}} |N_u^+|\Bigg] &= \sum_{e \in E^*} \sum_{\substack{u: ~ e = e(u):\\  D^+(u) \leq 1/4}} \mathbb{E}\left[|N_u^+| \cdot \mathbf{1}_{\{u \in \widebar{V_0}\}}\right] = \sum_{e \in E^*} \sum_{\substack{u: ~ e = e(u),\\  D^+(u) \leq 1/4}} |N_u^+| \cdot (1-q(\varepsilon))^{|N_u^+|},
\end{align*}
where again we use Corollary \ref{cor: subsample-independence} to see that $\mathbb{P}(u \in \widebar{V_0}) = (1-q(\varepsilon))^{|N_u^+|}$.
We now bound the inner sum. Fix $e = ab \in E^*$ such that $e$ is on a bad triangle $T$.  Now consider $u$ such that $D^+(u) \leq 1/4$ and $T = T(u)$. By definition, $u$ is on $T(u)$. Since $T$ is a bad triangle, $u \in N_a^+$ or $u \in N_b^+$ (or both) (where we may have $u=a$ or $u=b$, due to self-loops).

Consider $u \in N_a^+$. Since $D^+(u) \leq 1/4$, we have $d_{ua} \leq 1/4$. It is then straightforward to see from  $d_{ua} = 1-\nicefrac{|N_u^+ \cap N_a^+|}{|N_u^+ \cup N_a^+|}$ that $|N_u^+| \leq \nicefrac{4}{3} \cdot |N_a^+|$ and $|N_a^+| \leq \nicefrac{4}{3} \cdot |N_u^+|$. The analogous conclusion holds when $u \in N_b^+$. Continuing from the above,
\begin{align*}
&\qquad \sum_{e \in E^*} \sum_{\substack{u:~ e = e(u),\\  D^+(u) \leq 1/4}} |N_u^+| \cdot (1-q(\varepsilon))^{|N_u^+|} \\
&\leq \sum_{ab \in E^*} \Big(\sum_{\substack{u \in N_a^+:~ d_{ua} \leq 1/4}}  |N_u^+| \cdot (1-\nicefrac{\eps^2}{2})^{|N_u^+|} + \sum_{\substack{u \in N_b^+:~ d_{ub} \leq 1/4}}  |N_u^+| \cdot (1-\nicefrac{\eps^2}{2})^{|N_u^+|} \Big) \\
&\leq \sum_{ab \in E^*} \Big(\sum_{u \in N_a^+} \frac{4}{3} \cdot |N_a^+| \cdot (1-\nicefrac{\eps^2}{2})^{\frac{3}{4} \cdot |N_a^+|} + \sum_{u \in N_b^+} \frac{4}{3} \cdot |N_b^+| \cdot (1-\nicefrac{\eps^2}{2})^{\frac{3}{4} \cdot |N_b^+|} \Big) \\
&\leq \frac{4}{3} \cdot \sum_{ab \in E^*} \left(|N_a^+|^2 \cdot (1-\nicefrac{\eps^2}{2})^{\frac{3}{4} \cdot |N_a^+|}  + |N_b^+|^2 \cdot (1-\nicefrac{\eps^2}{2})^{\frac{3}{4} \cdot |N_b^+|}   \right) = \nicefrac{4}{3} \cdot \sum_{e \in E^*} \nicefrac{16}{\varepsilon^4} = \nicefrac{64}{3\eps^4} \cdot\textsf{OPT}_1.
\end{align*}
Note in the last line we use the fact that $n^2 \cdot (1-x)^{n \cdot 3/4} \leq 2/x^2$.
This concludes the proof.
\end{proof}

\subsection{Proof of item \ref{item: thm-1} for Theorem \ref{thm:main-all}}\label{sec: thm1-pf}

In this subsection, we show that the cases considered in the preceding lemmas are exhaustive, and thus that we can prove item \ref{item: thm-1} of Theorem \ref{thm:main-all}.

\begin{proof}[Proof of item \ref{item: thm-1} for Theorem \ref{thm:main-all}]
Let  $\mathcal{C}_{\textsf{ALG}}$ be the clustering output by Algorithm \ref{alg: main-alg}, and let $\text{cost}_1(\mathcal{C}_{\textsf{ALG}})$ be the $\ell_1$-norm of the disagreement vector of $\mathcal{C}_{\textsf{ALG}}$.
We further partition the edges in disagreement based on whether they are positive or negative, and which phase in Algorithm \ref{alg: main-alg} that are clustered in: positive edges where at least one endpoint of the edge is pre-clustered (Lemmas \ref{lem: type1b} and  \ref{lem: type-t}), negative edges where both endpoints are pre-clustered (Lemma \ref{lem: expected-negative}), edges where both endpoints are clustered in the Pivot phase (Lemmas \ref{lem: pivot-bar-V0-ell1} and \ref{lem: pivot-v0'-ell1}). Note this is exhaustive  for the disagreements; see Figure \ref{fig:summary-ell1}. Combining the terms from the lemmas, 
we see  
\begin{align*}
   \mathbb{E}[||y_{\mathcal{C}_{\textsf{ALG}}}||_1] \leq O(\nicefrac{1}{\eps^6})\cdot \textsf{OPT}_1.
\end{align*}

\end{proof}

\section{Tight Analysis of Algorithm \ref{alg: main-alg} for $p=\infty$}\label{sec: linf}

\begin{figure}[h]
\begin{minipage}{0.65\textwidth}
\hfill
\includegraphics[height=3.5cm]{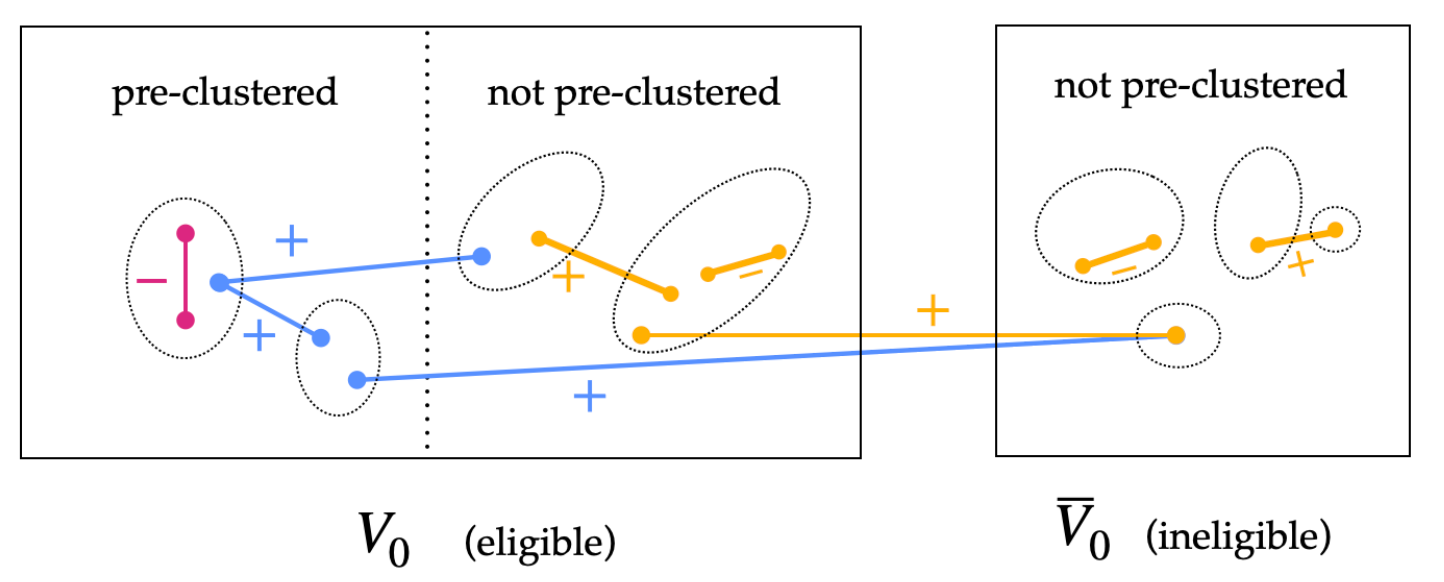}
\end{minipage}
\hfill 
\begin{minipage}{0.28\textwidth}
\footnotesize{\textcolor{myblue}{Lemmas \ref{lem: type1} and \ref{lem: type2}} \\
\textcolor{mypink}{Lemma \ref{lem: neg-whp}}  \\
\textcolor{myyellow}{Lemma \ref{lem: pivot-cost-minmax}}}
\end{minipage}
\caption{An overview of the cost analysis of Algorithm \ref{alg: main-alg} for $p=\infty$. The edges, ovals, etc., are as in Figures \ref{fig:summary-ellp} and \ref{fig:summary-ell1}.}
\label{fig:summary-infinity}
\end{figure}

As in the analysis for the $\ell_1$-norm, for the $\ell_\infty$-norm we bound the disagreements by partitioning edges based on their label and in which phase their endpoints were clustered. However, we are able to partition edges into fewer types than for the finite $\ell_p$-norms (see Figure \ref{fig:summary-infinity}, versus Figures \ref{fig:summary-ellp} and \ref{fig:summary-ell1}), and the analysis is in some sense simpler. For the $\ell_\infty$-norm objective, we know there exists a lower bound of $\Omega(\log n)$ in the online-with-a-sample model (Theorem \ref{thm: lowerbound-inf}). Intuitively, this allows us to essentially sacrifice vertices that have either small (e.g., $O(\log n))$ positive neighborhoods or a small number of close neighbors with respect to $\bar{d}$ (recall this is the estimated correlation metric of Definition \ref{def: est-corr}). On the other hand, for vertices that have large positive neighborhoods and a large number of neighbors with respect to $\bar{d}$,  these sets have good concentration, and thus we can obtain good estimates for $\tilde{d}$ and $|\text{Ball}^{S_b}_{\tilde{d}}(u,r)|$.

As with the analysis of the $\ell_p$-norm ($p \neq 1$) in Section \ref{sec: lp-norm-analysis}, we condition on the \emph{good event} $B^c$ (Definition \ref{def: good-event}) occurring with high probability (Lemma \ref{lem:good-whp}). Here we will use even more of the events defining $B^c$ than in the previous analysis, e.g., we will use that $B^c$ gives that for pairs of nodes $u,v$ with large combined positive neighborhood, 
$\bar{d}_{uv}$ is a good estimate of $d_{uv}$.

\subsection{Cost of Pre-clustering phase} \label{sec: pre-clustering-cost-infinity}

In this section, we bound the cost of disagreements incident to a vertex $u$ that arise when $u$, or a positive neighbor of $u$, is pre-clustered. 
Recall that $s^*(v)$ is the vertex in $S_p$ that clusters $v$, i.e. $v$'s center. Also recall we write $u \succ v$ if $u$ is pre-clustered before $u$ (with respect to the ordering on $S_p$, not the arrival order), or if $u$ is pre-clustered and $v$ is not. Recall the vertices $R_1$ are those isolated by $\tilde{d}$ (Definition \ref{def: est-adj-corr}). Throughout, $C$ will be a constant chosen sufficiently large, for instance $C=100$ is sufficient.

In Lemma  \ref{lem: type1}, we bound the cost of disagreements incurred for vertex $u$ from positive edges $uv$, where $v \succ u$.  
These disagreements are similar to those we
charged in Lemma \ref{lem: type1b-lp} for the finite $\ell_1$-norms, although we do not require $u \in V_0$. Then in Lemma \ref{lem: type2}, we bound the cost of disagreements incurred for vertex $u$ from positive edges $uv$, where $u \succ v$ (cf. Lemma \ref{lem: type1b-lp-backwards}). Then, Lemma \ref{lem: neg-whp} bounds the cost of negative edges whose endpoints were clustered together during the Pre-clustering phase.

\subsubsection{Cost of positive edges}

We first prove a few propositions and lemmas that will be helpful in bounding the costs of disagreements for both Lemmas \ref{lem: type1} and \ref{lem: type2}.

We begin with Proposition \ref{prop: perfect}, which says if $\textsf{OPT}_p = 0$,\footnote{When this holds for some $p$, it holds for all $p$.} then Algorithm \ref{alg: main-alg} finds the perfect clustering. This is necessary for our algorithm to be competitive. Its proof actually follows from the fact that Algorithm \ref{alg: main-alg} has expected cost a bounded factor away from the optimal cost when $p=1$ (i.e., item \ref{item: thm-1} of Theorem \ref{thm:main-all} is sufficient). However, we include below a more direct argument that is entirely independent of this result.

\begin{restatable}{proposition}{perfect}\label{prop: perfect}
    If $\textsf{OPT}_p = 0$, then Algorithm \ref{alg: main-alg} finds the perfect clustering, that is, the unique clustering with 0 disagreements. 
\end{restatable}

\begin{proof}[Proof of Proposition \ref{prop: perfect}]
    Let $G$ be a graph admitting a perfect clustering. Let $Q$ be any maximal positive clique in $G$. We need to show that $Q$ is precisely its own cluster in the clustering produced by Algorithm \ref{alg: main-alg}. 

    \begin{claim} \label{clm: clusters-travel-together}
        If $v \in V_0$, then $N_v^+ \subseteq V_0$. In other words, if any vertex of $Q$ is in $V_0$, then all of $Q$ is contained in $V_0$. 
    \end{claim}

    \begin{proof}[Proof of Claim \ref{clm: clusters-travel-together}]
        Recall that $V_0 = \{v \in V: N_v^+ \cap S_d \neq \emptyset\}$. So if $v \in V_0$, then there exists $w \in N_v^+$ that is in $S_d$. But for any $u \in N_v^+$, we also know that $w \in N_u^+$, since $v$ and $u$ have the same positive neighborhood. Thus $u$ has a positive neighbor (namely $w$) in $S_d$, meaning $u \in V_0$.
    \end{proof}

    \begin{claim} \label{clm: sep-clusters-frac-cut}
        If $v \in Q$ and $u \not \in Q$, then $\tilde{d}_{uv} = 1$. Conversely, if $Q \subseteq V_0$, then  $\tilde{d}_{uv} = 0$ for all $u,v \in Q$. 
    \end{claim}

    \begin{proof}[Proof of Claim \ref{clm: sep-clusters-frac-cut}]
        If $u \not \in V_0$, then $\tilde{d}_{uv} = 1$ by Fact \ref{fct: frac-isolation}. If $u \in V_0$, then $\bar{d}_{uv} = 1$, since $u$ and $v$ have no common positive neighbors. Thus $\tilde{d}_{uv}=1$, since $\tilde{d}_{uv} \geq \bar{d}_{uv}$ always.
        
        For the converse, note that $\bar{d}_{uv} = 0$, since $N_u^+ \cap N_v^+ \cap S_d = (N_u^+ \cup N_v^+) \cap S_d$, and $(N_u^+ \cup N_v^+) \cap S_d \neq \emptyset$. Moreover, by the first part of the claim, no negative edges have distance with respect to $\bar{d}$ strictly less than 1, so $\bar{d} = \tilde{d}$ (i.e., the rounding steps in Definition \ref{def: est-corr} do not apply). In particular, $\bar{d}_{uv} = \tilde{d}_{uv} = 0$ for $u,v \in Q$ when $Q \subseteq V_0$.  
    \end{proof}

    \setcounter{case}{0}
    \begin{case}
        $Q \cap S_p \neq \emptyset$.
    \end{case}
    The following subcases are exhaustive, by Claim \ref{clm: clusters-travel-together}.
    \begin{itemize}
        \item \textbf{Case 1a: $Q \subseteq V_0$.} \\
        Let $v \in C \cap S_p$ be chosen minimally with respect to the ordering on $S_p$. If $v \succ v'$, then $v' \not \in Q$, and then by Claim \ref{clm: sep-clusters-frac-cut}, $\tilde{d}_{uv'} =1$ for every $u \in Q$, so no vertex in $Q$ will be pre-clustered by $v'$. Thus, since $Q \subseteq V_0$ and  $\tilde{d}_{uv} = 0$ for every $u \in Q$ (Claim \ref{clm: sep-clusters-frac-cut}), all of $Q$ will be pre-clustered by $v$. Moreover, if $u \not \in Q$, then $u$ will not be pre-clustered by any $v \in Q$, again by Claim \ref{clm: sep-clusters-frac-cut}. 

        \item \textbf{Case 1b: $Q \cap V_0 = \emptyset$.}\\
        Then the vertices $Q$ are clustered in the run of Modified Pivot on $G[\widebar{V_0}]$, where in this case Modified Pivot is just the standard Pivot algorithm, which makes each maximal positive clique its own cluster. 
    \end{itemize}

    \begin{case}
        $Q \cap S_p = \emptyset$
    \end{case}

    In this case, no vertex in $Q$ will be pre-clustered, because $\tilde{d}_{uv} = 1$ for every $u \not \in Q$ and $v \in Q$ (Claim \ref{clm: sep-clusters-frac-cut}). If $Q \cap V_0 = \emptyset$, the argument is as in Case 1b. If $Q \subseteq V_0$, then $Q$ is clustered in the run of Modified Pivot on $G[V_0 \cap V']$. Modified Pivot in this case takes the set  $E_c = E^+ \cap \{uv \in E : \tilde{d}_{uv} < c \cdot r\}$ of clusterable edges as input. But by Claim \ref{clm: sep-clusters-frac-cut}, $E_c = E^+$, so Modified Pivot is just the standard Pivot algorithm, which will put $Q$ in its own cluster. 
\end{proof}

Given Proposition \ref{prop: perfect}, we may assume in the remainder of this section that $\textsf{OPT}_\infty \geq 1$.

\medskip

Recall in Lemma \ref{lem: whp-distances} we showed with high probability $\bar{d}_{uv}$ is a good estimate of $d_{uv}$, so long as $|N_u^+ \cup N_v^+|$ is large. We do not obtain an analogous statement for $\tilde{d}$. In lieu of this, we show in Proposition \ref{prop: minmax-isolated-lem} that for vertices $u$ affected by the (second) adjustment in Definition \ref{def: est-adj-corr} (that is, the vertices $u \in R_1$ isolated by $\tilde{d}$) we have that $|N_u^+|$ is a bounded factor away from $\textsf{OPT}_\infty$ with high probability.\footnote{Note that even though $|N_u^+|$ and $\textsf{OPT}_\infty$ are deterministic quantities, the statement is probabilistic due to the event $u \in R_1$.} This bound is necessary, as for these vertices, the number of disagreements incident to $u$ is $|N_u^+|$.

 \begin{proposition} \label{prop: minmax-isolated-lem}
     Condition on the good event $B^c$. Fix $u \in R_1$ (the vertices isolated by $\tilde{d}$) such that $|N_u^+| \geq C \cdot \log n / \varepsilon^2$.
     Then
    $ |N_u^+| 
    \leq O \left (\nicefrac{1}{ \varepsilon^4}\cdot \log n\right) \cdot \textsf{OPT}_\infty.$

\end{proposition}
\begin{proof}[Proof of Proposition \ref{prop: minmax-isolated-lem}]
    Since the event $B^c$ holds, 
    $|N_u^+ \cap S_d| \geq \frac{\varepsilon^2}{4} \cdot |N_u^+|.$
    Also, when $u \in R_1$, 
    $|N_u^- \cap S_r \cap \{v \in V: \bar{d}_{uv} \leq 7/10\}| \geq \frac{10}{3} \cdot |N_u^+ \cap S_d|.$
    Combining the above two bounds, we have for $u \in R_1$, 
 
    \begin{align*}
        |N_u^+| &\leq \frac{3}{10} \cdot \frac{4}{\varepsilon^2} \cdot |N_u^- \cap \{v \in V: \bar{d}_{uv} \leq 7/10\}| 
        \leq \frac{3}{10} \cdot \frac{4}{\varepsilon^2} \cdot \sum_{v \in N_u^-} \frac{10}{3} \cdot (1-\bar{d}_{uv})
    \end{align*}
    giving the first inequality in the statement of the proposition. Then by the event $B^c$ (which says $\bar{d}$ is a good estimate of $d$ since $|N_u^+|$ is sufficiently large) and Theorem \ref{thm: corr-metric-cost} (which says that $||D||_\infty \leq 8 \cdot \opt_1$), then
    $ |N_u^+| \leq \nicefrac{4}{\varepsilon^2} \cdot \sum_{v \in N_u^-} (1-\bar{d}_{uv}) 
    \leq \nicefrac{4(C+1)}{3 \varepsilon^4}\cdot \log n \cdot D(u)
    \leq \nicefrac{32(C+1)}{3 \varepsilon^4}\cdot \log n \cdot \textsf{OPT}_\infty.$
\end{proof}

In the following two lemmas, 
Lemmas \ref{lem: R1-asymmetric} and \ref{lem: cut-positive-edges}, we charge the cost of the positive neighborhood of a vertex to the cost of the estimated correlation metric $\bar{d}$, which we in turn know is comparable to $\textsf{OPT}_\infty$ by the good event. 

The proof of Lemma \ref{lem: R1-asymmetric} relies on the fact that since $\bar{d}_{uv}$ is small, the joint sampled neighborhood $|N_u^+ \cap N_v^+ \cap S_d|$ is large, and moreover
$|N_u^+ \cap S_d|$  and $|N_v^+ \cap S_d|$ are similar. Since we condition on the good event, $|N_v^+|$ and $|N_{u}^+|$ are similar too. If both neighborhood sizes $|N_{u}^+|$ and $|N_{v}^+|$ are small, the claim automatically follows (using the fact that $\textsf{OPT}_\infty \geq 1$), and if they are large, then we can upper bound them by the $\ell_\infty$-norm with the help of Proposition \ref{prop: minmax-isolated-lem}. 

\begin{restatable}{lemma}{Roneasym}
    \label{lem: R1-asymmetric}
    Condition on the good event $B^c$. Fix $u,v \in V$. Suppose $|N_u^+| \geq C \cdot \log n / \varepsilon^2$,  $\bar{d}_{u,v} \leq 1/4$, and $\tilde{d}_{u,v} = 1$. 
     Then
    $|N_u^+| \leq O \left (\nicefrac{1}{ \varepsilon^6}\cdot  \log n\right )  \cdot  \textsf{OPT}_\infty$.
\end{restatable}

\begin{proof}[Proof of Lemma \ref{lem: R1-asymmetric}]
    Given the hypotheses of the lemma, the edge $uv$ must have been rounded up in estimating the adjusted correlation metric (Definition \ref{def: est-adj-corr}). So either $u \in R_1$ or $v \in R_1$. If $u \in R_1$, we have by Proposition \ref{prop: minmax-isolated-lem} that
    $|N_u^+| \leq\nicefrac{32 \cdot (C+1)}{3 \varepsilon^4}\cdot \log n \cdot \textsf{OPT}_\infty$. 
        
    Otherwise, $v \in R_1$. Since $\bar{d}_{uv} \leq 1/4$, we have 
    that $|N_u^+ \cap S_d|$ and $|N_v^+ \cap S_d|$ are similar, as in the following fact.
    Recall we define the random variable $X_w$ be 1 if $w \in S_d$ and 0 otherwise.

    \begin{fact}\label{fact:sim-sizes}
        For $u,v$ with $\bar{d}_{uv} \leq 1/4,$ we have that $|N_u^+ \cap S_d| \leq \frac43 \cdot |N_v^+ \cap S_d|$.
    \end{fact}
    \begin{proof}[Proof of Fact \ref{fact:sim-sizes}]
        We rewrite $\bar{d}_{uv}$ as 
        \[\bar{d}_{uv} = \frac{ \sum_{w \in N_u^+ \setminus N_v^+} X_w + \sum_{w \in N_v^+ \setminus N_u^+} X_w}{\sum_{w \in N_u^+ \cup N_v^+} X_w}. \]
        Fixing any realization of $S_d$, we can further rewrite $\bar{d}_{uv} \leq 1/4$ as 
          \begin{align*}
               |(N_u^+ \setminus N_v^+) &\cap S_d |+ |( N_v^+ \setminus N_u^+) \cap S_d|\\ &\leq \frac14 \cdot \big ( |( N_u^+ \setminus N_v^+) \cap S_d| + |(N_v^+ \setminus N_u^+) \cap S_d| + |( N_u^+ \cap N_v^+) \cap S_d | \big ),
            \end{align*}
          which is equivalent to 
          \begin{align*}
          \frac34 \cdot |(N_u^+ \setminus N_v^+) \cap S_d| +  \frac34 \cdot | ( N_v^+ \setminus N_u^+) \cap S_d| &\leq \frac14 \cdot |( N_u^+ \cap N_v^+) \cap S_d| \\
          |(N_u^+ \setminus N_v^+) \cap S_d |+  |( N_v^+ \setminus N_u^+) \cap S_d |&\leq \frac13 \cdot |( N_u^+ \cap N_v^+) \cap S_d |.
          \end{align*}
          Therefore, 
          \begin{align*}
          |N_u^+ \cap S_d| &= |(N_u^+ \cap N_v^+ )\cap S_d| + |(N_u^+ \setminus N_v^+) \cap S_d| \\
          &\leq 
          |(N_u^+ \cap N_v^+) \cap S_d| +\frac13 \cdot | (N_u^+ \cap N_v^+)\cap S_d | - |(N_v^+ \setminus N_u^+) \cap S_d|\\
          &=  \frac43 \cdot |N_u^+ \cap N_v^+ \cap S_d| - |(N_v^+ \setminus N_u^+) \cap S_d|\\
          & \leq \frac43 \cdot |N_v^+ \cap S_d|.
           \end{align*}
    \end{proof}

    Further, since $B^c$ holds, 
$\nicefrac{\varepsilon^2}{4} \cdot |N_u^+| \leq |N_u^+ \cap S_d|.$
Also,  
    $|N_{v}^+ \cap S_d| \leq |N_{v}^+|  .$
     Combining the above inequalities with Fact \ref{fact:sim-sizes}, we have 
    \[ |N_u^+| \leq  \frac{4}{\varepsilon^2} \cdot \frac{4}{3} \cdot |N_{v}^+|.\]
    If $|N_{v}^+| < C \cdot \log n / \varepsilon^2$, then we are done, with 
    $|N_u^+| \leq \nicefrac{16 C}{3  \varepsilon^4} \cdot \log n$. 
    
    Otherwise, $|N_{v}^+| \geq C \cdot \log n / \varepsilon^2$, and we have by Proposition \ref{prop: minmax-isolated-lem} that
    $|N_v^+|  \leq  \nicefrac{32(C+1)}{3 \varepsilon^4}\cdot \log n \cdot \textsf{OPT}_\infty,$ so 
    $|N_u^+| \leq \nicefrac{512(C+1) }{9 \varepsilon^6} \cdot  \log n \cdot  \textsf{OPT}_\infty$.
\end{proof}

Lastly, Lemma \ref{lem: cut-positive-edges} shows that when conditioning on the good event, the number of nodes far from $u$ with respect to $\tilde{d}$ is upper bounded by $\log n \cdot O(\textsf{OPT}_\infty).$ 
The proof follows automatically by Proposition \ref{prop: minmax-isolated-lem} if $u \in R_1$. If $u \not \in R_1$, we can apply Lemma \ref{lem: R1-asymmetric} if there is a $v$ that satisfies the conditions of that lemma. 
Otherwise we are able to use the the good event, and conclude  that $\bar{D}(u)$ is upper bounded by $O\big ( \nicefrac{1}{\eps^2} \cdot \log n \big)\cdot \text{OPT}_\infty.$

\begin{restatable}{lemma}{cutpos}
    \label{lem: cut-positive-edges}
    Condition on the good event $B^c$. Fix $u \in V$. Suppose $|N_u^+| \geq C \cdot \log n / \varepsilon^2$, and let $s > 0$ be a constant.
Then 
    $|\{v \in N_u^+ : \tilde{d}_{uv} \geq s \}| \leq O\left ( \nicefrac{1}{\varepsilon^6} \cdot \log n \right )\cdot \textsf{OPT}_\infty.$
\end{restatable}

\begin{proof}[Proof of Lemma \ref{lem: cut-positive-edges}]
We consider two cases.

\setcounter{case}{0}
\begin{case}
    $u \in R_1$.
\end{case}
We use the bound $|\{v \in N_u^+ : \tilde{d}_{uv} \geq s\}| \leq |N_u^+|$. Since $|N_u^+| \geq C \cdot \log n / \varepsilon^2$, we may apply Proposition \ref{prop: minmax-isolated-lem}, and obtain
    \[|\{v \in N_u^+ : \tilde{d}_{uv} \geq s\}| \leq  |N_u^+| \leq  \nicefrac{(32\cdot (C+1)\cdot \log n  )}{3 \varepsilon^4}\cdot \textsf{OPT}_\infty.\]

\begin{case}\label{case: case2E2}
    $u \not \in R_1$.
\end{case}

We consider two subcases. 

    \begin{itemize}

\item If there exists $v^* \in N_u^+ \cap R_1 \cap \{v \in V: \bar{d}_{uv} \leq 1/4 \}$, then $\bar{d}_{uv^*} \leq 1/4$ but $\tilde{d}_{uv^*} = 1$ (because $v^* \in R_1$). So we may apply Lemma \ref{lem: R1-asymmetric} and obtain that
    \[|\{v \in N_u^+ : \tilde{d}_{uv} \geq s\}| \leq |N_u^+|  \leq \nicefrac{(512\cdot (C+1) \cdot  \log n)}{9 \varepsilon^6}  \cdot  \textsf{OPT}_\infty.\]

    \item Otherwise, $N_u^+ \cap R_1 \cap \{v \in V: \bar{d}_{uv} \leq 1/4 \} = \emptyset$. Then
    \[|\{v \in N_u^+ : \tilde{d}_{uv} \geq s\}| \leq \sum_{v \in N_u^+ \setminus R_1}  \nicefrac{1}{s} \cdot \bar{d}_{uv}  + \sum_{v \in N_u^+ \cap R_1} 4 \cdot \bar{d}_{uv} \leq
\nicefrac{(8(C+1)(4+1/s)\cdot \log n)}{3\varepsilon^2}  \cdot \textsf{OPT}_\infty.\]
    In the first inequality we have used that if $u,v \not \in R_1$ and $uv \in E^+$, then $\bar{d}_{uv} = \tilde{d}_{uv}$, along with the fact that $\tilde{d}_{uv} \geq s$ by assumption. In the second inequality, we have used the fact that $|N_u^+| \geq C \cdot \log n / \varepsilon^2$ and applied  Lemma \ref{lem: whp-distances}. 
    \end{itemize}

\end{proof}

Equipped with all the necessary lemmas, we are ready to prove Lemmas \ref{lem: type1} and \ref{lem: type2}.

\begin{lemma} \label{lem: type1}
    Condition on $B^c$. 
    Fix $u \in V$. 
    Then
    $|\{v \in N_u^+ \mid v \succ u\}|  \leq  O\left (\nicefrac{1}{\eps^6}\cdot \log n \right)\cdot \textsf{OPT}_\infty. $
\end{lemma}

\begin{proof}[Proof of Lemma \ref{lem: type1}]
We may assume $\textsf{OPT}_\infty \geq 1$ per Proposition \ref{prop: perfect}. Take $t := r/(2\delta)$. We have 
    \begin{align*}
   \{v \in N_u^+ \mid v \succ u\}  &= \underbrace{  \{v \in N_u^+ \mid v \succ u, \tilde{d}_{uv} \leq t\} }_{E_1(u)} + \underbrace{  \{v \in N_u^+ \mid  v \succ u, \tilde{d}_{uv} > t\} }_{E_2(u)}
    \end{align*}
If $|N_u^+| < C \cdot \log n / \varepsilon^2$, then we are done, so we assume for the remainder of the proof that $|N_u^+| \geq C \cdot \log n / \varepsilon^2$. In this case, since $B^c$ holds, $|N_u^+ \cap S_d| \geq C \cdot \log n / 4 \geq 1$, so $u \in V_0$.

\paragraph{Bounding $|E_2(u)|$.}
Applying Lemma \ref{lem: cut-positive-edges} with $s = t = \nicefrac{r}{(2\delta)}$, we have 
\begin{align}\label{eqn: e2-linf}
    |E_2(u)| \leq\nicefrac{(512(C+1)\cdot \log n)}{9\varepsilon^6} \cdot \textsf{OPT}_\infty.
\end{align}

\paragraph*{Bounding $|E_1(u)|$.} We case based on $|\ball_{\bar{d}}(u,t)|$.

\medskip 

\noindent  \textbf{Case 1.} $|\ball_{\bar{d}}(u,t)| \leq C' \cdot \log_{1/(1-\varepsilon^2/2)} n$.
\quad
We know that $\bar{d} \leq \tilde{d}$, so $E_1(u) \subseteq \ball_{\bar{d}}(u,t)$. Thus \[|E_1(u)| \leq C' \cdot \log_{1/(1-\varepsilon^2/2)} n \leq C' \cdot \log_{1/(1-\varepsilon^2/2)} n \cdot \opt_\infty,\] 
where the second inequality is from Proposition \ref{prop: perfect}.

\medskip

\noindent 
\textbf{Case 2.}  $|\ball_{\bar{d}}(u,t)| > C' \cdot \log_{1/(1-\varepsilon^2/2)} n.$    
\quad
We use the next two claims to bound $|E_1(u)|$. 

Define  
    \[B(u):= \bigcup_{v \in E_1(u)} \ball_{\tilde{d}}(s^*(v), r).\]

    \begin{claim} \label{clm: upper-and-lower-bds-d}
        For every $b \in B(u)$, 
        $\tilde{d}_{ub} >1/10$ and $1-\tilde{d}_{ub} >3/10.$
    \end{claim}
    \begin{proof}[Proof of Claim \ref{clm: upper-and-lower-bds-d}]
    First we lower bound $\tilde{d}_{ub}$ for $b \in B(u)$. Since $b \in B(u)$, there exists $v \in E_1(u)$ such that 
    $\tilde{d}_{b\hspace{1pt}s^*(v)} \leq r.$
    On the other hand, since $u \in V_0$ and $v$ is clustered before $u$, 
    $\tilde{d}_{u\hspace{1pt}s^*(v)} > c \cdot r.$
    By the approximate triangle inequality (Lemma \ref{lem: tri-inequality}), 
    $\tilde{d}_{ub} \geq \frac{1}{\delta} \cdot \tilde{d}_{u\hspace{1pt}s^*(v)} - \tilde{d}_{b\hspace{1pt}s^*(v)},$
    and rearranging and substituting in $c,r,\delta$, we see 
$\tilde{d}_{ub} \geq c \cdot r / \delta-r > 1/10$.

    Next we upper bound $\tilde{d}_{ub}$ for $b \in B(u)$. Let $v$ and $s^*(v)$ be as before. We have
    $\tilde{d}_{b\hspace{1pt}s^*(v)} \leq r$,
    $\tilde{d}_{v\hspace{1pt}s^*(v)} \leq c \cdot r$, and
    $\tilde{d}_{u v} \leq t.$
   By Lemma \ref{lem: tri-inequality} and substituting in $c,r,\delta$, we have
    $\tilde{d}_{ub} \leq \delta \cdot [\tilde{d}_{b \hspace{1pt} s^*(v)} + \delta(\tilde{d}_{v \hspace{1pt}s^*(v)} + \tilde{d}_{vu})] \leq \delta \cdot r + \delta^2 \cdot c \cdot r + \delta^2 \cdot t < 7/10.$
\end{proof}

    \begin{claim} \label{clm: CuAu}
       At least one of the following holds:
         $|E_1(u)| \leq \frac{8}{\varepsilon^2} \cdot |B(u)|$ or  
         $|E_1(u)| \leq O \left (\nicefrac{1}{ \varepsilon^6} \cdot  \log n  \right ) \cdot  \textsf{OPT}_\infty.$
    \end{claim}

\begin{proof}[Proof of Claim \ref{clm: CuAu}]

     Since $B^c$ holds, there exists  
    $w \in \ball_{\bar{d}}^{S_p}\left(u, t  \right).$
   We consider two cases:

\medskip

    \noindent  \textbf{Case 1}. $\tilde{d}_{u w} = \bar{d}_{u w}$.
\quad In this case, $\tilde{d}_{uw} \leq t \leq c \cdot r$. So since $w \in S_p$ and $u \in V_0$, we know that $u$ is pre-clustered (but not necessarily by $w$). So $s^*(u)$ exists. 
        For every $v \in E_1(u)$, we have that $\tilde{d}_{uv} \leq t$, and also $\tilde{d}_{uw} \leq t$ (by assumption of the case). So by the approximate triangle inequality (Lemma \ref{lem: tri-inequality}), $\tilde{d}_{vw} \leq \delta \cdot (\tilde{d}_{vu} + \tilde{d}_{uw}) \leq \delta \cdot 2t = r$ 
        for every $v \in E_1(u)$, which means that 
        \begin{equation*} \label{eq: type1bd1}
            |E_1(u)| \leq |\ball_{\tilde{d}}(w, r)| .
        \end{equation*}
        If $|\ball_{\tilde{d}}(w, r)| < C \cdot \log n / q(\eps)$, then we are done. So assume $|\ball_{\tilde{d}}(w, r)| \geq C \cdot \log n / q(\eps)$. Then, since $B^c$ holds, we have that  
    $|\ball_{\tilde{d}}(w, r)|\leq 4 \cdot |\ball^{S_b}_{\tilde{d}}(w, r)|/q(\eps).$
        We also have 
 $|\ball^{S_b}_{\tilde{d}}(w, r)| \leq |\ball^{S_b}_{\tilde{d}}(s^*(w), r)|$,
        since $\tilde{d}_{uw} \leq c \cdot r$ and $w \in S_p$. Moreover, since for every $v \in E_1(u)$, $u$ is clustered after $v$ (this by definition implies $v$ pre-clustered, thus that $v \in V_0$), we have that 
    $|\ball^{S_b}_{\tilde{d}}(s^*(u), r)|  \leq |\ball^{S_b}_{\tilde{d}}(s^*(v), r)| $
        for every $v \in E_1(u)$. 

        We know that for any $v \in E_1(u)$,
        $ |\text{Ball}_{\tilde{d}}(s^*(v),r)|\leq |B(u)|.$
        So putting it all together,
        \[|E_1(u)| \leq 4 \cdot |\ball^{S_b}_{\tilde{d}}(s^*(v), r)|/q(\eps)  \leq \nicefrac{8}{\varepsilon^2} \cdot |B(u)|.\]

\medskip 
   \noindent \textbf{Case 2.} $\tilde{d}_{uw} = 1$. \quad
        We have $\bar{d}_{uw} \leq t \leq 1/4$, so we may apply Lemma \ref{lem: R1-asymmetric} and obtain that
    $|E_1(u)| \leq |N_u^+|   \leq \nicefrac{(512(C+1) \cdot  \log n )}{9 \varepsilon^6} \cdot  \textsf{OPT}_\infty.$
\end{proof}

    We are done if the second condition of Claim \ref{clm: CuAu} holds, so we assume that the first condition holds. Then we use Claim \ref{clm: upper-and-lower-bds-d} to see that
    \begin{align*}
        |E_1(u)|& \leq \frac{2}{q(\eps)} \cdot |B(u)| \leq \frac{2}{q(\eps)} \cdot \Big(\sum_{b \in B(u) \cap N_u^+} 10 \cdot \bar{d}_{ub} + \sum_{b \in B(u) \cap N_u^-} \frac{10}{3} \cdot (1-\bar{d}_{ub})\Big)
        \end{align*}
    where we have used that since $\tilde{d}_{ub} \neq 1$ for $b \in B(u)$, we have that $\tilde{d}_{ub} = \bar{d}_{ub}$. As a consequence of conditioning on $B^c$, we can bound $\bar{d}$ by $\textsf{OPT}_\infty$
    (see Lemma \ref{lem: whp-distances}), so
     \begin{align}   
     |E_1(u)| \leq \nicefrac{ (320\cdot (C+1) \cdot \log n)}{3\varepsilon^4}
 \cdot \textsf{OPT}_\infty.   \label{eqn: e1-linf}
 \end{align}

 \medskip

Combining  Equations (\ref{eqn: e2-linf}) and (\ref{eqn: e1-linf}).
 \[\big | \{v \in N_u^+ \mid v \succ u\} \big |\leq  \nicefrac{(1472 \cdot (C+1) \cdot \log n)}{9 \eps^6} \cdot \textsf{OPT}_\infty.\]
\end{proof}

Next, we bound the cost of the positive disagreements incident to $u$ that arise when $u$ is pre-clustered, and either $v \in N_u^+$ is pre-clustered later or not pre-clustered at all, i.e., $u \succ v$. 

\begin{lemma} \label{lem: type2}
    Condition on  $B^c$. Fix $u \in V$. Then
    $\left | \{v \in N_u^+ \mid u \succ v\} \right |  \leq O \left (\nicefrac{1}{\eps^6} \cdot \log n  \right) \cdot \textsf{OPT}_\infty.$
\end{lemma}
We note that the proof of Lemma \ref{lem: type2} is very similar to that of Lemma \ref{lem: type1}.

\begin{proof}[Proof of Lemma \ref{lem: type2}]
 We may assume $\textsf{OPT}_\infty \geq 1$ per Proposition \ref{prop: perfect}. Take $t = r/(2\delta)$. We partition as in Lemma \ref{lem: type1},

    \begin{align*}
     \{v \in N_u^+ \mid u \succ v\}  &= \underbrace{ \{v \in N_u^+ \mid u \succ v, \tilde{d}_{uv} \leq t\} }_{E_1(u)} + \underbrace{  \{v \in N_u^+ \mid  u \succ v, \tilde{d}_{uv} > t\} }_{E_2(u)}
    \end{align*}

If $|N_u^+| < C \cdot \log n / \varepsilon^2$, then we are done, so we assume for the remainder of the proof that $|N_u^+| \geq C \cdot \log n / \varepsilon^2$. Note in this case that, since $B^c$ holds, $|N_u^+ \cap S_d| \geq C \cdot \log n / 4 \geq 1$, so $u \in V_0$.

\paragraph*{Bounding $|E_2(u)|$.} By applying Lemma \ref{lem: cut-positive-edges} with $s = t$, we have that 
\begin{align}
    |E_2(u)| \leq\nicefrac{(512\cdot (C+1)\cdot \log n)}{9\varepsilon^6} \cdot \textsf{OPT}_\infty.\label{eqn: e2-type2-linf}
\end{align}

\paragraph*{Bounding $|E_1(u)|$.} First note that for any $v$ that is clustered after $u$, we have $\tilde{d}_{v \hspace{1pt}s^*(u)} > c \cdot r$. (Note we are using here that by Fact \ref{fct: frac-isolation}, it is not possible that $v \not \in V_0$, since $\tilde{d}_{uv} \leq t < 1$.)  Define 
\[W(u):= \{v \in N_u^+ \mid \tilde{d}_{uv} \leq t, \tilde{d}_{v \hspace{1pt}s^*(u)} > c \cdot r \}.\]
So to bound $|E_1(u)|$, it suffices to bound $|W(u)|$.

\setcounter{case}{0}

\begin{case}
    $|\ball_{\bar{d}}(u,t)| \leq C' \cdot \log_{1/(1-q(\varepsilon))} n$.
\end{case}
We know that $\bar{d} \leq \tilde{d}$, so $W(u) \subseteq \ball_{\bar{d}}(u,t)$. Thus $|W(u)| \leq C' \cdot \log_{1/(1-q(\varepsilon))} n$, concluding the case.

\medskip

\begin{case}
$|\ball_{\bar{d}}(u,t)| > C' \cdot \log_{1/(1-q(\varepsilon))} n$.    
\end{case}

Define
\[B(u):= \ball_{\tilde{d}}(s^*(u), r).\]

\begin{claim} \label{clm: bounds-d-type2}
    For every $b \in B(u)$, $\tilde{d}_{ub} \geq c \cdot r / \delta^2 - t/\delta - r$ and $1-\tilde{d}_{ub} \geq 1- \delta \cdot (c \cdot r+r)$. 
\end{claim}

\begin{proof}[Proof of Claim \ref{clm: bounds-d-type2}]
    Let $b \in B(u)$. Let $w$ be an arbitrary vertex in $W(u)$ (if $W(u) = \emptyset$, then $|E_1(u)| = 0$ and we are done). By the approximate triangle inequality (Lemma \ref{lem: tri-inequality}), we have

     \[\tilde{d}_{ub} \geq \nicefrac{1}{\delta} \cdot \left(\nicefrac{1}{\delta} \cdot \tilde{d}_{w\hspace{1pt}s^*(u)} - \tilde{d}_{u w} \right) - \tilde{d}_{b \hspace{1pt}s^*(u)} \geq c \cdot r / \delta^2 - t/\delta - r\]

     and also 
     \[\tilde{d}_{ub} \leq \delta \cdot (\tilde{d}_{u \hspace{1pt} s^*(u)} + \tilde{d}_{b \hspace{1pt}s^*(u)} ) \leq \delta \cdot (c \cdot r + r) \]

which concludes the proof of the claim. 
    
\end{proof}

\begin{claim} \label{clm: BuAu-type2}
   At least one of the following holds:
    \begin{itemize}
       \item $|W(u)| \leq  O \left (\nicefrac{1}{\eps^2} \right)\cdot |B(u)|$ 
       \item $|W(u)| \leq O \left (\nicefrac{1}{ \varepsilon^6}\cdot \log n \right )  \cdot  \textsf{OPT}_\infty$. 
   \end{itemize}

\end{claim}

\begin{proof}[Proof of Claim \ref{clm: BuAu-type2}]

Since $B^c$ holds, there exists  $z(u) \in \ball_{\bar{d}}^{S_p}\left(u, t  \right).$
 
We consider two cases:

\setcounter{case}{0}
\begin{case}
    $\tilde{d}_{u \hspace{1pt}z(u) } = \bar{d}_{u \hspace{1pt}z(u) }$.
\end{case}

        For every $v \in W(u)$, we have that $\tilde{d}_{uv} \leq t$ (by definition of $W(u))$, and also $\tilde{d}_{u \hspace{1pt}z(u)} \leq t$ (by assumption of the case). So by the approximate triangle inequality (Lemma \ref{lem: tri-inequality}), $$\tilde{d}_{v \hspace{1pt}z(u)} \leq \delta \cdot (\tilde{d}_{vu} + \tilde{d}_{u \hspace{1pt}z(u)}) \leq \delta \cdot 2t = r$$ 
        for every $v \in W(u)$, which means that 
        \begin{equation*} \label{eq: type2bd1}
            |W(u)| \leq |\ball_{\tilde{d}}(z(u), r)|.
        \end{equation*}
         If $|\ball_{\tilde{d}}(z(u), r)| < C \cdot \log n / q(\eps)$, then we are done. So assume $|\ball_{\tilde{d}}(z(u), r)| \geq C \cdot \log n / q(\eps)$. Then, since $B^c$ holds, we have that 
        $|\ball_{\tilde{d}}(z(u), r)| \leq 4 \cdot |\ball^{S_b}_{\tilde{d}}(z(u), r)|/q(\eps)$.
        We have 
  $|\ball^{S_b}_{\tilde{d}}(z(u), r)| \leq |\ball^{S_b}_{\tilde{d}}(s^*(u), r)|,$
        since $\tilde{d}_{u\hspace{1pt}z(u)} \leq t \leq c \cdot r$ and $z(u) \in S_p$. 

        So putting it all together, 

        \[|W(u)| \leq 4 \cdot 
        |\ball^{S_b}_{\tilde{d}}(s^*(u), r)|/q(\eps) \leq   \nicefrac{8}{\eps^2} \cdot |B(u)|.\]

\begin{case}
    $\tilde{d}_{u\hspace{1pt}z(u)} = 1$.
\end{case}

        We have $\bar{d}_{u\hspace{1pt}z(u)} \leq t \leq 1/4$, so we may apply Lemma \ref{lem: R1-asymmetric} and obtain that
    \[|W(u)| \leq |N_u^+|  \leq \nicefrac{(512\cdot (C+1) \cdot \log n)}{9 \varepsilon^6} \cdot  \textsf{OPT}_\infty.\]

\end{proof}

Now we see that the claims imply the sought bound on $|E_1(u)|$. 
We are done if the second bullet of Claim \ref{clm: BuAu-type2} holds, so we assume that the first bullet holds. We then have, using Claim \ref{clm: bounds-d-type2}, that
\begin{align*}
    |E_1(u)| = |W(u)| &\leq \frac{2}{q(\eps)} \cdot  |B(u)|\\ 
    &\leq \frac{2}{q(\eps)} \cdot \Bigg(\sum_{b \in B(u) \cap N_u^-} \frac{1}{1- \delta \cdot (c \cdot r+r)} \cdot (1-\bar{d}_{ub}) + \sum_{b \in B(u) \cap N_u^+ } \frac{1}{c \cdot r / \delta^2 - t/\delta - r} \cdot \bar{d}_{ub} \Bigg)  
    \end{align*}
where we have used that since $\tilde{d}_{ub} \neq 1$ for $b \in B(u)$, we have that $\tilde{d}_{ub} = \bar{d}_{ub}$. So by Lemma \ref{lem: whp-distances} and substituting in $\delta = 10/7$, $c = 2\delta^2 + \delta$, and $ r= \frac{1}{2c\delta^2}$, we have that
\begin{align}
    |E_1(u)| 
\leq \nicefrac{(512(C+1) \cdot \log n)}{3\eps^4} \cdot \textsf{OPT}_\infty. \label{eqn: e1-type2-linf}
\end{align}

Combining the bounds on $|E_1(u)|$ and $|E_2(u)|$ from Equations (\ref{eqn: e1-type2-linf}) and (\ref{eqn: e2-type2-linf}), we see that
\[ \big | \{v \in N_u^+ \mid u \succ v\} \big |  \leq \nicefrac{(2048(C+1) \cdot \log n)}{9\eps^6}\cdot \textsf{OPT}_\infty.\]

\end{proof}

\subsubsection{Cost of negative edges}

Next we bound the cost of the negative disagreements adjacent to $u$ that are incurred during the Pre-clustering phase. Note that the only way a negative disagreement can arise during the Pre-clustering phase is if the same center in $S_p$ clusters its endpoints.

\begin{lemma} \label{lem: neg-whp}
    Fix $u$ pre-clustered. Then
    $\big |  \big \{v \in N_u^- \mid v \text{ clustered with } u \big \}\big | \leq O(\nicefrac{1}{ \eps^4}\cdot \log n )\cdot \textsf{OPT}_\infty.$
\end{lemma}

\begin{proof}[Proof of Lemma \ref{lem: neg-whp}]
    We may assume $\textsf{OPT}_\infty \geq 1$ per Proposition \ref{prop: perfect}. Let $v \in N_u^-$ be such that $u,v$ are clustered together during the Pre-clustering phase. Then we may set $s^* := s^*(u) = s^*(v)$. Thus we have $\tilde{d}_{us^*} \leq c \cdot r$ and $\tilde{d}_{vs^*} \leq c \cdot r$. So by the approximate triangle inequality (Lemma \ref{lem: tri-inequality}), 
    $\tilde{d}_{uv} \leq 2 \delta  c  r.$
    Since $\tilde{d}_{uv} < 1$, $\bar{d}_{uv} = \tilde{d}_{uv}$ and $u \not \in R_1$.  We consider two cases. 
    
\setcounter{case}{0}

\begin{case}
    $|N_u^+| \geq C \cdot \log n / \varepsilon^2$. 
\end{case}
    Then we have by Lemma \ref{lem: whp-distances} that 
    \[\big |  \big \{v \in N_u^- \mid v \text{ clustered with } u \big \}\big |  \leq \sum_{v \in N_u^-} \frac{1}{1-2\delta c r } \cdot (1- \bar{d}_{uv}) \leq 
    \frac{8\cdot (C+1)}{3\varepsilon^2\cdot (1-2 \delta c r)}  \cdot \log n \cdot \textsf{OPT}_\infty.\]

\begin{case}
    $|N_u^+| < C \cdot \log n / \varepsilon^2$.
\end{case}

Note $2\delta \cdot c \cdot r < 7/10$. Then it suffices to bound 
\[|N_u^- \cap \{v \in V \mid \bar{d}_{uv} \leq 7/10 \}|.\]
If $|N_u^- \cap \{v \in V: \bar{d}_{uv} \leq 7/10 \}| < C \cdot \log n / q(\varepsilon)$, then we are done. So we may assume 
\[|N_u^- \cap \{v \in V: \bar{d}_{uv} \leq 7/10 \}| \geq C \cdot \log n / q(\varepsilon).\]
Then, since $B^c$ holds, 
\begin{align*}
    |N_u^- \cap \{v \in V: \bar{d}_{uv} \leq 7/10 \}| &\leq \nicefrac{8}{\varepsilon^2} \cdot |N_u^- \cap S_r \cap \{v \in V: \bar{d}_{uv} \leq 7/10 \}| \\
    &\leq \nicefrac{80}{3\varepsilon^2}  \cdot |N_u^+ \cap S_d| \\
    &\leq \nicefrac{80}{3\varepsilon^2} \cdot |N_u^+| \leq \nicefrac{80}{3\varepsilon^4} \cdot C  \log n \cdot \opt_\infty
\end{align*}
where in the second inequality we have used that $u \not \in R_1$, and in the last we use Proposition \ref{prop: perfect}. This concludes the case and the proof. 

\end{proof}

\subsection{Cost of Pivot phase for $\ell_\infty$}\label{sec: pivot-linfty}

Let $G'=(V', E')$ be the subgraph induced by the unclustered vertices. Let $V_0'$ be defined as for the tight $\ell_1$-norm bound, that is, 
\[ V_0' = V_0 \cap \{v \in V: \tilde{d}_{vu_i} > c \cdot r  \text{ for all } u_i \in S_p\},\]
and recall that $V' = \widebar{V_0} \cup V_0'$.
Also recall that 
$\mathcal{C}_{\text{ALG}}$ is the clustering output by Algorithm \ref{alg: main-alg}. 
\begin{lemma} \label{lem: pivot-cost-minmax}
    Condition on the good event $B^c$. Fix $u \in V'$. The number of disagreements that $\mathcal{C}_{\text{ALG}}$ has in $G[V']$ incident to $u$ is $O(\frac{1}{\varepsilon^6} \cdot \log n )\cdot \textsf{OPT}_\infty$.  
\end{lemma}

\begin{proof}[Proof of Lemma \ref{lem: pivot-cost-minmax}]
    By Proposition \ref{prop: perfect}, we may assume $\textsf{OPT}_\infty \geq 1$. We case on whether $u \in \widebar{V_0}$ or $u \in V_0'$.

    \setcounter{case}{0}

    \begin{case}
    $u \in V_0'$. 
    \end{case}

    Recall that on the subgraph $G[V_0']$, we run Modified Pivot with $E_c = E^+ \cap \{uv \in E \mid \tilde{d}_{uv} < c \cdot r\}$. Let $u^*$ be $u$'s pivot. 
    
    \begin{claim} \label{clm: pivot-minmax-bd}
        The number of disagreements $\mathcal{C}_{\text{ALG}}$ has in $G[V']$ incident to $u \in V_0'$ is at most $2 \cdot |N_u^+| + |N_{u^*}^+|$.
    \end{claim}

    \begin{proof}[Proof of Claim \ref{clm: pivot-minmax-bd}]
    To see this, observe that if $u$ is a pivot (i.e., $u^* = u$), then $u$ has no negative \textit{neighbors} in its cluster, so there are at most $|N_u^+|$ disagreements in $G[V_0']$ incident to $u$ in the worst case. On the other hand, if $u$ is not a pivot (so $u^* \neq u$), then there may be negative disagreements in $G[V_0']$ incident to $u$. These are of the form $uv \in E^-$ where $v$'s pivot is also $u^*$. But then $v \in N_{u^*}^+$. So the number of negative disagreements in $G[V_0']$ incident to $u$ is at most $|N_{u^*}^+|$. 

    Finally, the remaining disagreements $uv$ in $G[V']$ to account for are those with $v \in N_u^+ \cap \widebar{V_0}$, thus giving an additive factor of $|N_u^+|$. 
    \end{proof}

    Now we upper bound $2 \cdot |N_u^+| + |N_{u^*}^+|$ for $u \in V_0'$ (thus also $u^* \in V_0'$). If $|N_u^+|, |N_{u^*}^+| < C \cdot \log n / \varepsilon^2$, then we are done. So suppose $|N_u^+| \geq C \cdot \log n / \varepsilon^2$ (the same argument will apply for $u^*$). Then
    \[|N_u^+| \leq \frac{4}{\eps^2} \cdot |N_u^+ \cap S_p| \leq \frac{4}{\eps^2} \cdot|N_u^+ \cap \{v \in V \mid \tilde{d}_{uv} > c \cdot r\}| \leq   \nicefrac{(2048(C+1)\cdot \log n)}{9 \cdot \varepsilon^6}  \cdot \textsf{OPT}_\infty,\]
where in the first inequality, we have used that $B^c$ holds, in the second that by hypothesis  $u \in V_0'$, and in the third that we may take $s = c \cdot r=0.245$ in Lemma \ref{lem: cut-positive-edges}.

We conclude that the number of disagreements in $G[V']$ incident to $u$ is bounded in the case that $u \in V_0'$ by $\nicefrac{2048(C+1)}{3 \cdot \varepsilon^6}\cdot \log n\cdot \textsf{OPT}_\infty$.

\begin{case}
    $u \in \widebar{V_0}$.
\end{case}

Recall that on the subgraph $G[\widebar{V_0}]$, we run standard Pivot with $E_c = E^+$. Let $u^*$ be $u$'s pivot.  

    \begin{claim}\label{clm: lp-small-nbhd}
        The number of disagreements $\mathcal{C}_{\text{ALG}}$ has in $G[V']$ incident to $u \in \widebar{V_0}$ is at most $2 \cdot |N_u^+| + |N_{u^*}^+|$.
    \end{claim}

The proof of the claim is the same as that of Claim \ref{clm: pivot-minmax-bd} since there, we only used that $E_c \subseteq E^+$. 

Now we upper bound $2 \cdot |N_u^+| + |N_{u^*}^+|$ for $u \in \widebar{V_0}$ (thus also $u^* \in \widebar{V_0}$). Since $B^c$ holds and $|N_u^+ \cap S_d| = 0$ (that is, $u \not \in V_0$), it must be the case that $|N_u^+| \leq C \cdot \log n / \varepsilon^2$, and likewise for $u^*$. This concludes the case and the proof.

\end{proof}

\subsection{Proof of Item \ref{item: thm-infty} of Theorem \ref{thm:main-all}}

We note the disagreements  are all accounted for,
see Figure \ref{fig:summary-infinity}. In particular, disagreements where one endpoint 
was pre-clustered and the other was not are accounted for by Lemmas \ref{lem: type1} and \ref{lem: type2}. Disagreements for the entire Pivot phase are accounted for in Lemma \ref{lem: pivot-cost-minmax}.

In total, we combine the bounds from Lemmas  \ref{lem: type1}, \ref{lem: type2}, \ref{lem: neg-whp}, and \ref{lem: pivot-cost-minmax} to see that 
\[||y_{ \mathcal{C}_{ALG}} ||_\infty \leq  O \left (\nicefrac{1 }{\eps^6}\cdot \log n \right ) \cdot \textsf{OPT}_\infty.\]

\section{Lower Bounds}
\label{sec: LB}
In this section, 
we first show that any strictly online (deterministic or randomized) algorithm is $\Omega(n)$-competitive for the $\ell_\infty$-norm objective. We do this by showing that the lower bound in, e.g., \cite{MathieuSS10, LattanziMVWZ21} giving a competitive ratio of $\Omega(n)$ for the $\ell_1$-norm objective, also easily gives a lower bound for the $\ell_\infty$-norm objective. 

Then we show a lower bound of $\Omega(n^{1/3})$ on the competitive ratio in the RO model for the $\ell_\infty$-norm objective (Theorem \ref{thm: lowerbound-inf-ro}). Thus, while the RO model is more powerful than the strictly online setting, strong lower bounds still exist in this model for $\ell_p$-norm correlation clustering. This motivates our decision to study $\ell_p$-norm correlation clustering in the AOS setting. We further prove a lower bound of $\Omega(\nicefrac{\log n}{\eps})$ on the competitive ratio of any (deterministic or randomized) algorithm in the AOS model for the $\ell_\infty$-norm objective (Theorem \ref{thm: lowerbound-inf}). In both lower bounds, the idea is to use as gadgets the instances from the lower bound in the strictly online setting. 

\medskip

To prove our results, we will use Yao's Min-Max Principle.

\begin{theorem} [Yao's Min-Max Principle]
\label{thm: Yao}
Fix a problem with a set $\mathcal{X}$ of possible inputs and a set $\mathcal{A}$ of deterministic algorithms solving the problem. 
Denote the cost of algorithm $A \in \mathcal{A}$ on input $X \in \mathcal{X}$ as $c(A,X) \geq 0$.
Further, let $\mathcal{D}$ be a distribution over $\mathcal{A}$, and let $A^* \in \mathcal{A}$ be an  algorithm drawn randomly from $\mathcal{D}$ (that is, $A^*$ is a randomized algorithm). Similarly, let $\mathcal{P}$ be a distribution over inputs $\mathcal{X}$ and let $X^* \in \mathcal{X}$ be an input drawn randomly from $\mathcal{P}$. Then 
\[\max_{X \in \mathcal{X}} \mathbb{E}_{A^* \sim \mathcal{D}}[c(A^*,X) ]\geq \min_{A \in \mathcal{A}}\mathbb{E}_{X^* \sim \mathcal{P}}[c(A,X^*)].\]
\end{theorem}
Yao's Min-Max Principle states that the expected cost of any randomized algorithm (thus also the best one) for a worst-case input is at least the expected cost of the best deterministic algorithm for a chosen distribution of inputs.

\subsection{Lower bound for strictly online model} \label{sec: strictly-online-lower}

We begin by recalling the $\Omega(n)$ lower bound for the $\ell_1$-norm objective in the online setting, which was
first given by Mathieu, Sankur, and Schudy \cite{MathieuSS10}, and we note it holds even for randomized algorithms (see Theorem 3.4 in \cite{MathieuSS10}). We show this bound actually holds for the $\ell_\infty$-norm objective too.

For an algorithm $A$ and input $X$, define $\cratio_\infty(A, X)$ to be the ratio of the $\ell_\infty$-norm cost of $A$ on $X$ to the optimal offline cost of $X$; we will take the cost $c$ in Theorem \ref{thm: Yao} to be $\cratio_\infty$. Note that, because we are in the online setting, instance $X$ consists of both an underlying input graph \emph{and} an order of arrival for vertices.
Let $G_1$ be a graph consisting of a positive edge $(v_1, v_2)$, with each endpoint contained in a positive clique of size $\nicefrac{n}{2}$, and the remaining edges between the two cliques negative. Let instance $X_1$ consist of the graph $G_1$ with $v_1, v_2$ arriving first (and the remaining order of arrivals arbitrary). Let $G_2$ be a positive clique of $n$ nodes, and $v_1, v_2$ be two arbitrarily chosen distinguished vertices. Let the instance $X_2$ consist of $G_2$ with $v_1, v_2$ likewise arriving first.
Consider the distribution $\mathcal{P}$ over inputs $G_1$ and $G_2$, where $X_1$ and $X_2$ each have probability $\nicefrac{1}{2}$
of being sampled.

Let $\mathcal{A}$ be the family of all deterministic algorithms for correlation clustering in the strictly online model, and $\mathcal{X}$ be the set of all inputs on $n$ nodes. It is sufficient to partition these based on whether or not they place $v_1,v_2$ together.
If $A \in \mathcal{A}$ puts $v_1$ and $v_2$ in the same cluster, and the remainder of instance $X_1$ arrives online,
the competitive ratio $\cratio_\infty(A,X_1)$ is at least $\nicefrac{n}{2}$, since the $\ell_\infty$-norm cost of $A$ on $G_1$ is at least $\nicefrac{n}{2}$, regardless of the remaining decisions $A$ makes, whereas the optimal (offline) cost for the $\ell_\infty$-norm objective on $G_1$ is 1. 
On the other hand, if $A \in \mathcal{A}$ puts $v_1$ and $v_2$ in different clusters,
and instead the remainder of instance $X_2$ arrives online, the competitive ratio  $\cratio_\infty(A,X_2)$ of $A$  on $G_2$ is  $\infty$, since the optimal solution for the $\ell_\infty$-norm objective is 0.  

 Now we apply Yao's Principle (Theorem \ref{thm: Yao}) to obtain the sought lower bound for \emph{randomized} algorithms, where $\mathcal{D}$ is a distribution over $\mathcal{A}$ and $\mathcal{P}$ is the distribution over $\mathcal{X}$ above:
 \begin{equation} \label{eq: strictly-online-infty}
 \max_{X \in \mathcal{X}} \mathbb{E}_{A^* \sim \mathcal{D}}[\cratio_\infty(A^*,X) ] \geq \min_{A \in \mathcal{A}}\mathbb{E}_{X^* \sim \mathcal{P}}[\cratio_\infty(A,X^*)] \overset{\star}{\geq} \min \Big \{\nicefrac{1}{2} \cdot  \nicefrac{n}{2}, \infty  \Big \}= \nicefrac{n}{4}
\end{equation}
where in ($\star$) we have partitioned $A \in \mathcal{A}$ based on whether $v_1,v_2$ are in the same or different clusters.

\subsection{Lower bound for RO model} \label{sec: RO-lower-bd}
We now prove Theorem \ref{thm: lowerbound-inf-ro}. To do so, we first construct a distribution $\mathcal{P}$ over inputs $\mathcal{X}$ on $n$ nodes. (Since we are now in the RO model, an input is a graph, \emph{not} a graph with an arrival order as above.) 
Each support point of the distribution is a graph consisting of $n^{1-\delta}$ gadgets, $H_1,\ldots, H_{n^{1-\delta}}$, with all negative edges between gadgets. 
For each gadget independently,
$H_i$ is a copy of $G_1$ on $ n^{\delta}$ 
nodes with probability $\nicefrac{1}{2}$, and is a copy of $G_2$ on $ n^{\delta}$ nodes with probability $\nicefrac{1}{2}$.
For each $H_i$, we denote the two distinguished vertices $v_{i,1}$ and $v_{i,2}$. The key idea is to show that with constant probability, there is at least one gadget $H_i$ for which $v_{i,1}$ and $v_{i,2}$ arrive first in $H_i$ with respect to the random ordering.

Let $\mathcal{A}$ be the family of deterministic algorithms for the problem.  Let $\cratio_\infty(A,R,X)$ denote the  ratio of the $\ell_\infty$-norm cost of algorithm $A \in \mathcal{A}$ on graph $X$ with random arrival order $R$ to the optimal offline cost of $X$.

\begin{lemma}\label{lem: helper-ro-lb}
      Given $ \delta \leq \nicefrac{1}{3}$, every \emph{deterministic} algorithm in the RO model is $\Omega(n^{\delta})$-competitive for the $\ell_\infty$-norm objective. 
\end{lemma}
\begin{proof}[Proof of Lemma \ref{lem: helper-ro-lb}] 
With the notation above, we would like to lower bound 
\[
 \min_{A \in \mathcal{A}} \max_{X \in \mathcal{X}} \mathbb{E}_{R} \Big  [\cratio_\infty(A,R,X)
\Big ].\]

Define $E^i$ to be the event that $v_{i,1}$ and $v_{i,2}$ are the first two nodes in $H_i$ to arrive. 
We will show in the following claim that with good probability, there is some gadget $H_i$ where event $E^i$ occurs. Then we will argue one can reduce to the lower bound for the strictly online model on gadget $H_i$. 

\begin{claim} \label{clm: gadget-Ei}
    For $\delta \leq \nicefrac{1}{3}$, at least one event $E^i$ occurs with constant probability. In particular,
    \[\mathbb{P} [\cup_{i \in [n^{1-\delta}]} E^i] \geq 1- \nicefrac{1}{e}.\]
\end{claim}
\begin{proof}[Proof of Claim \ref{clm: gadget-Ei}]
Since the nodes within a single gadget also arrive in random order, 
\begin{align*}
    \mathbb{P}[E^i] &=  \mathbb{P}[v_{i,1}, v_{i,2} \text{ arrive first in }H_i] \geq \nicefrac{1}{n^{2 \delta}}.
\end{align*}
Let $\widebar{E^i}$ denote the complement of the event $E^i$, so $\mathbb{P} [\widebar{E^i}] \leq 1-\nicefrac{1}{n^{2 \delta}}$.
Also note that the events $\{E^i\}_{i \in [n^{1-\delta}]}$ are all independent. Then, 
    \begin{align*}
        \mathbb{P} [\cup_{i \in [n^{1-\delta}]} E^i] = 1-\prod_{i \in [n^{1-\delta}]} \mathbb{P} [\widebar{E^i}]  &=  1-\big ( \mathbb{P} [\widebar{E^i}] \big )^{n^{1-\delta}} \geq  1-\big (1-\nicefrac{1}{ n^{2 \delta}}\big )^{n^{1-\delta}} \geq 1-\nicefrac{1}{e},
    \end{align*}
    where in the last line we use that  $\delta \leq 1/3$. This concludes the proof of the claim.
\end{proof}
 In what follows, conditioned on the event $\cup_{i \in [n^{1-\delta}]} E^i$, let $H^*$ be an arbitrary gadget such that $H^* = H_{i^*}$ and event $E^{i^*}$ occurs. Let $X^* \sim \mathcal{P}$. Then 

\begin{align}
    & \quad \min_{A \in \mathcal{A}} \max_{X \in \mathcal{X}} \mathbb{E}_{R} \left  [\cratio_\infty(A, R, X)\right] \notag \\ 
    &\geq \min_{A \in \mathcal{A}} \mathbb{E}_{X^*}\left[\mathbb{E}_{R} \left  [\cratio_\infty(A, R, X^*)\right] \right]  \label{eq: lower-bd-ro-intermediate}\\
    &\geq \min_{A \in \mathcal{A}} \mathbb{E}_{(X^*, R)}\left[\textsf{CR}_\infty(A, R, X^*) \right] \notag \\
    &\geq \min_{A \in \mathcal{A}} \mathbb{E}_{(X^*, R)}\left[\textsf{CR}_\infty(A, R, X^*) \mid \cup_{i \in [n^{1-\delta}]} E^i \right] \cdot \mathbb{P}\left[\cup_{i \in [n^{1-\delta}]} E^i \right] \notag \\
    &\geq \left(1-\nicefrac{1}{e}\right) \cdot \min_{A \in \mathcal{A}} \mathbb{E}_{(X^*, R)}\left[\textsf{CR}_\infty(A, R, H^*) \mid \cup_{i \in [n^{1-\delta}]} E^i \right] \notag \\
    &\geq \left(1-\nicefrac{1}{e}\right) \cdot \min_{A \in \mathcal{A}} \mathbb{E}_{(X^*, R)}\left[\mathbf{1}_{\{H^* = G_1\}}\textsf{CR}_\infty(A, R, H^*) + \mathbf{1}_{\{H^* = G_2\}}\textsf{CR}_\infty(A, R, H^*)  \mid \cup_{i \in [n^{1-\delta}]} E^i \right] \notag \\
    &\geq \left(1-\nicefrac{1}{e}\right) \cdot \min \left\{ \mathbb{E}_{(X^*, R)}\left[\mathbf{1}_{\{H^* = G_1\}} \cdot \nicefrac{n^\delta}{2} \mid \cup_{i \in [n^{1-\delta}]} E^i \right], \mathbb{E}_{(X^*, R)} \left[ \mathbf{1}_{\{H^* = G_2\}} \cdot \infty  \mid \cup_{i \in [n^{1-\delta}]} E^i \right] \right\} \notag \\
    &\geq \left(1-\nicefrac{1}{e}\right) \cdot \nicefrac{1}{2} \cdot \nicefrac{n^\delta}{2} \notag
\end{align}

where in the second inequality we applied Fubini-Tonelli as $X^*$ and $R$ are independent; in the fourth inequality we used Claim \ref{clm: gadget-Ei}; and in the penultimate inequality we repeated the argument in line (\ref{eq: strictly-online-infty}). 
\end{proof}

\begin{proof}[Proof of Theorem \ref{thm: lowerbound-inf-ro}]
Let $\mathcal{D}$ be any distribution over $\mathcal{A}$. Let $A^* \sim \mathcal{D}$ and $X^* \sim \mathcal{P}$. Define, for any $A \in \mathcal{A}$ and $X \in \mathcal{X}$, cost function $c(A, X) := \mathbb{E}_R\left[\textsf{CR}_\infty(A, R,X) \right]$ . Then the worst-case expected cost of $A^*$ over inputs $\mathcal{X}$ is
\begin{align*}
\max_{X \in \mathcal{X}} \mathbb{E}_{(A^*, R)}\left[\textsf{CR}_\infty(A^*, R, X) \right] = \max_{X \in \mathcal{X}} \mathbb{E}_{A^*} [c(A^*,X)] \geq \min_{A \in \mathcal{A}} \mathbb{E}_{X^*}  [c(A,X^*)]
= \Omega(n^{\delta}),
\end{align*}
where the first equality is by Fubini-Tonelli as $A^*$ and $R$ are independent; the inequality is by Yao's Principle (Theorem \ref{thm: Yao}); and the final bound is by the lower bound on the same expression (\ref{eq: lower-bd-ro-intermediate}) in the proof of Lemma \ref{lem: helper-ro-lb}.
\end{proof}

\subsection{Lower bound for AOS model}

\begin{figure}[t!]
    \centering
    \captionsetup{width=.7\textwidth}
\includegraphics[width=8cm]{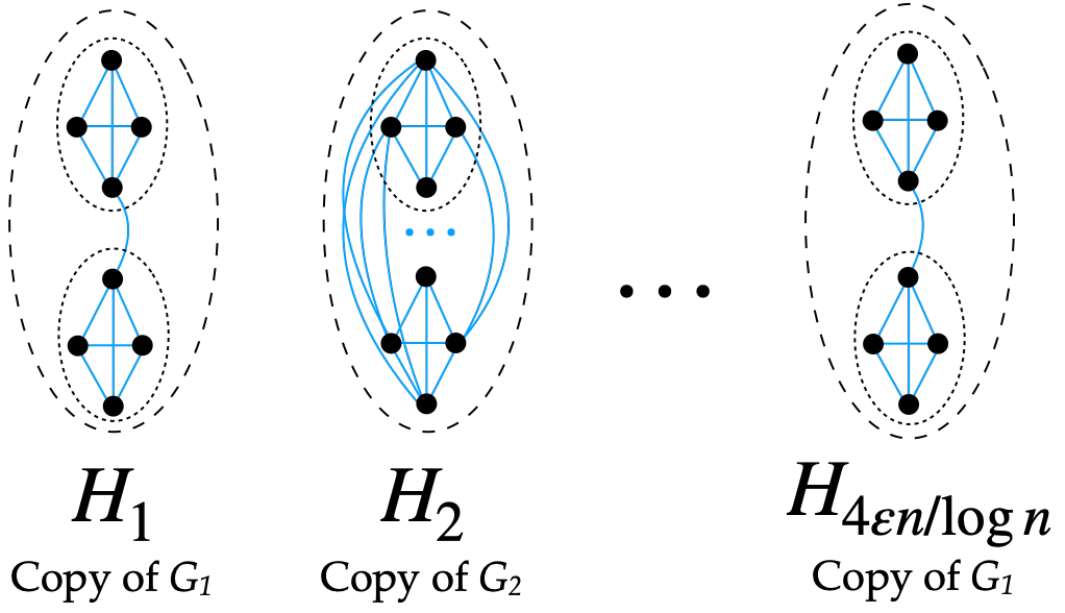}
    \caption{An instance from a distribution $\mathcal{P}$ that gives the lower bound in Lemma \ref{lem: determ-lower-bd-infty}. The larger ovals with the dashed border represent the $4\varepsilon n/ \log n$ gadgets. Each gadget is either a positive clique, or two equally sized positive cliques with exactly one positive edge between them. Positive edges are drawn while negative edges are not.}
    \label{fig:lb-semi-online}
\end{figure}

We now prove Theorem \ref{thm: lowerbound-inf}, specifically, the $\Omega(\nicefrac{\log n}{\eps})$ lower bound on the competitive ratio for the $\ell_\infty$-norm objective in the AOS model. As before, we split a graph on $n$ vertices into several gadgets. The goal is to show that, with constant probability, there is at least one gadget that the sample $S$ does not hit; thus, this gadget is revealed in a strictly online fashion, so the algorithm incurs high cost.

We again construct a distribution $\mathcal{P}$ over inputs $\mathcal{X}$. See Figure \ref{fig:lb-semi-online} for an illustration of the construction.
Consider $\nicefrac{4\varepsilon n}{\log n}$ gadgets $H_i$ for $i \in [\nicefrac{4\varepsilon n}{\log n}]$. 
For each gadget independently, $H_i$ is a copy of $G_1$ on $\nicefrac{\log n}{4\varepsilon}$ vertices with probability $\nicefrac{1}{2}$, and is a copy of $G_2$ (again on $\nicefrac{\log n}{4\varepsilon}$ vertices) with probability $\nicefrac{1}{2}$. The order in which (non-sampled vertices in) gadgets arrive does not matter, but we assume the worst-case order ($v_1$ and $v_2$ arriving first) for each individual gadget as in subsection \ref{sec: strictly-online-lower}. 

Let $\mathcal{A}$ be the family of deterministic algorithms for the problem.\footnote{Note that in order to consider deterministic algorithms, one must consider the random sample $S$ to be part of the input: algorithm $A \in \mathcal{A}$ receives upfront a realization of the random sample $S$ once this has been fixed, and all of its decisions are deterministic from then on.} Let $\cratio_\infty(A,S,X)$ denote the  ratio of the $\ell_\infty$-norm cost of algorithm $A \in \mathcal{A}$ on graph $X$ with random sample $S$ to the optimal offline cost of $X$.

\begin{lemma} \label{lem: determ-lower-bd-infty}
    Given $\nicefrac{1}{n^{1/4}} \leq \varepsilon \leq \nicefrac{3}{4}$, every \emph{deterministic} algorithm in the AOS model is $\Omega(\nicefrac{\log n}{\varepsilon})$-competitive for the $\ell_\infty$-norm objective. 
\end{lemma}
\begin{proof}[Proof of Lemma \ref{lem: determ-lower-bd-infty}]
With the notation above, we would like to lower bound 
\[
 \min_{A \in \mathcal{A}} \max_{X \in \mathcal{X}} \mathbb{E}_{S} \Big  [\cratio_\infty(A,S,X)
\Big ].\]

First we will lower bound the probability that for any fixed instance $X$ in the support of $\mathcal{P}$, at least one gadget of $X$ is not sampled by $S$ (meaning \emph{none} of its vertices are in $S$).

\begin{claim}\label{claim: none-sampled}
    Let $\frac{1}{n^{1/4}} \leq \varepsilon \leq \frac{3}{4}$. For any instance $X$ in the support of $\mathcal{P}$, there is some gadget that is not sampled by $S$ with probability at least $0.99$. 
\end{claim}
\begin{proof}[Proof of Claim \ref{claim: none-sampled}]
    Fix an instance $X$ in the support of $\mathcal{P}$. Recall $X$ contains $\nicefrac{4\varepsilon n}{\log n}$ gadgets $H_i$ of size $\nicefrac{\log n}{4\varepsilon}$ each. The probability that not all gadgets are sampled by $S$ (where we use fact that nodes in $X$ are in $S$ independently and with probability $\varepsilon$) is 
    \begin{align*}
        \mathbb{P}[\text{not all gadgets in } X \text { sampled}] &=1-\prod_{i=1}^{\nicefrac{4\varepsilon n}{\log n}}\mathbb{P}[\text{gadget }H_i \text{ sampled}] \\
        &=  1-\prod_{i=1}^{\nicefrac{4\varepsilon n}{\log n}}\Big (1-(1-\varepsilon)^{|H_i|}\Big )= 1-\Big ( 1-(1-\varepsilon)^{\nicefrac{\log n}{4\varepsilon}}\Big )^{\nicefrac{4\varepsilon n}{\log n}}.
    \end{align*}
    The we continue lower bounding the above using the fact that $1-x \geq e^{-2x}$ for $0 \leq x \leq 3/4$:
    \begin{align*}
        \mathbb{P}[\text{not all gadgets in } X \text { sampled}]&= 1-\Big ( 1-(1-\varepsilon)^{\nicefrac{\log n}{4\varepsilon}}\Big )^{\nicefrac{4\varepsilon n}{\log n}} \geq 1-\Big ( 1-e^{\nicefrac{-\log n}{2}}\Big )^{\nicefrac{4\varepsilon n}{\log n}}\\
        & = 1-\Big ( 1-\nicefrac{1}{\sqrt{n}}\Big )^{\nicefrac{4\varepsilon n}{\log n}} \geq 1-e^{-\nicefrac{4\varepsilon \sqrt{n}}{\log n}}\\
        & \geq 1-e^{-\nicefrac{4 n^{1/4}}{\log n}} \geq 0.99,
        \end{align*}
        where the penultimate  inequality holds using $\varepsilon \geq \nicefrac{1}{n^{1/4}}$. 
\end{proof}
Now we are ready for the lower bound. For shorthand, let $E_{X^*,S}$ be the event the sample $S$ hits every gadget in $X^*$, and let $\overline{E}_{X^*,S}$ be its complement. Let $H^*$ be a gadget that is \emph{not} sampled by $S$ in the case $\overline{E}_{X^*,S}$ holds. For any $A \in \mathcal{A}$, let $\textsf{cost}(A,H^*)$ denote the $\ell_\infty$-norm objective cost of $A$ restricted to the gadget $H^*$. Note that, by definition, $H^*$ arrives \emph{fully online}. Then
\begin{align}
    \min_{A \in \mathcal{A}} \max_{X \in \mathcal{X}} \mathbb{E}_{S} \left  [\cratio_\infty(A,S,X)
\right ] 
&\geq  \min_{A \in \mathcal{A}} \mathbb{E}_{X^*}\left[ \mathbb{E}_{S} \left  [\cratio_\infty(A,S,X^*)
\right ] \right] \label{eq: interm-lower-bd}\\
& =  \min_{A \in \mathcal{A}} \mathbb{E}_{(X^*,S)} \left  [\cratio_\infty(A,S,X^*)
\right ]  \label{eq: fubini}\\
&\geq \min_{A \in \mathcal{A}} \mathbb{E}_{(X^*,S)} \left[ \cratio_\infty(A,S,X^*)
\mid \overline{E}_{X^*,S} \right ]\cdot \mathbb{P}[\overline{E}_{X^*,S}] \notag \\
&\geq \mathbb{P}[\overline{E}_{X^*,S}] \cdot \min_{A \in \mathcal{A}} \mathbb{E}_{(X^*,S)} \left[ \textsf{cost}(A,H^*) 
\mid \overline{E}_{X^*,S} \right ]\cdot  \label{eq: CR-to-cost} \\
&= \mathbb{P}[\overline{E}_{X^*,S}] \cdot \min_{A \in \mathcal{A}} \Big(\frac{1}{2} \cdot \textsf{cost}(A, G_1) + \frac{1}{2} \cdot \textsf{cost}(A, G_2)  \Big) \label{eq: high-cost-split} \\
&\geq \mathbb{P}[\overline{E}_{X^*,S}] \cdot \frac{\log n}{16  \varepsilon} \label{eq: high-cost} \\
& \geq 0.99 \cdot \frac{\log n}{16  \varepsilon} \notag
\end{align}
Line (\ref{eq: fubini}) holds by Fubini-Tonelli, since $X^*$ and $S$ are independent.
In line (\ref{eq: CR-to-cost}) we have used that the optimal solution for any instance $X^* \sim \mathcal{P}$ has cost at most 1. In lines (\ref{eq: high-cost-split}) and (\ref{eq: high-cost}), we repeated the argument in line (\ref{eq: strictly-online-infty}), since $H^*$ arrives fully online. The last line follows from Claim \ref{claim: none-sampled}.
\end{proof}

\begin{proof}[Proof of Theorem \ref{thm: lowerbound-inf}] 
Let $\mathcal{D}$ be any distribution over $\mathcal{A}$. Let $A^* \sim \mathcal{D}$ and $X^* \sim \mathcal{P}$. Define, for any $A \in \mathcal{A}$ and $X \in \mathcal{X}$, cost function $c(A, X) := \mathbb{E}_S\left[\textsf{CR}_\infty(A, S,X) \right]$ . Then the worst-case expected cost of $A^*$ over inputs $\mathcal{X}$ is
\begin{align*}
\max_{X \in \mathcal{X}} \mathbb{E}_{(A^*, S)}\left[\textsf{CR}_\infty(A^*, S, X) \right] = \max_{X \in \mathcal{X}} \mathbb{E}_{A^*} [c(A^*,X)] \geq \min_{A \in \mathcal{A}} \mathbb{E}_{X^*}  [c(A,X^*)]
= \Omega(\nicefrac{\log n}{\varepsilon}),
\end{align*}
where the first equality is by Fubini-Tonelli as $A^*$ and $S$ are independent; the inequality is by Yao's Principle (Theorem \ref{thm: Yao}); and the final bound is by the lower bound on the same expression (\ref{eq: interm-lower-bd}) in the proof of Lemma \ref{lem: determ-lower-bd-infty}.

The lower bound of $\Omega(\nicefrac{1}{\varepsilon})$ for the $\ell_1$-norm in the statement of the theorem is already proven in Theorem 1 of \cite{LattanziMVWZ21}. 

\end{proof}

\end{document}